\documentclass[letterpaper]{article} 
\usepackage{aaai23}  
\usepackage{times}  
\usepackage{helvet}  
\usepackage{courier}  
\usepackage[hyphens]{url}  
\usepackage{graphicx} 
\usepackage{natbib}  
\usepackage{caption} 
\usepackage{algorithm}

\usepackage{newfloat}
\usepackage{listings}
\usepackage{natbib}
\DeclareCaptionStyle{ruled}{labelfont=normalfont,labelsep=colon,strut=off} 
\floatstyle{ruled}
\newfloat{listing}{tb}{lst}{}
\floatname{listing}{Listing}
\usepackage{subfig}

\usepackage{booktabs}       
\usepackage{amsfonts}       
\usepackage{amsmath}	    
\usepackage{dsfont}         
\usepackage{multirow} 
\usepackage{numprint}

\usepackage{algpseudocode}  

\newcommand{\ignore}[1]{}

\newcommand{\norm}[1]{\left\lVert#1\right\rVert}
\newcommand{\sprod}[2]{\langle{#1}, {#2}\rangle}
\newcommand{\setR}{\mathbb{R}}
\newtheorem{theorem}{Theorem}

\DeclareMathOperator{\cns}{cns}
\DeclareMathOperator{\cnm}{cnm}
\DeclareMathOperator{\cas}{cas}
\DeclareMathOperator{\car}{car}
\DeclareMathOperator{\cds}{cds}
\DeclareMathOperator{\cdr}{cdr}
\DeclareMathOperator{\std}{std}
\DeclareMathOperator{\cactive}{active}
\DeclareMathOperator{\dead}{dead}
\DeclareMathOperator{\dyr}{dyr}
\DeclareMathOperator{\dys}{dys}
\DeclareMathOperator{\pr}{pr}

\title{Why Capsule Neural Networks Do Not Scale:\\ Challenging the Dynamic Parse-Tree Assumption}
\author{
    Matthias Mitterreiter,\textsuperscript{\rm 1,\rm 4}
    Marcel Koch,\textsuperscript{\rm 2}
    Joachim Giesen,\textsuperscript{\rm 1}
    S\"oren Laue\textsuperscript{\rm 3}
}
\affiliations{
    \textsuperscript{\rm 1}Friedrich-Schiller-University Jena, Germany\\
    \textsuperscript{\rm 2}Ernst Abbe University of Applied Sciences Jena, Germany\\ 
    \textsuperscript{\rm 3}Technical University Kaiserslautern, Germany\\
    \textsuperscript{\rm 4}Data Assessment Solutions GmbH, Hannover, Germany\\
    matthias.mitterreiter@uni-jena.de, marcel.koch@eah-jena.de, joachim.giesen@uni-jena.de, laue@cs.uni-kl.de
}

\begin{document}

\maketitle

\begin{abstract}
Capsule neural networks replace simple, scalar-valued neurons with vector-valued capsules. They are motivated by the pattern recognition system in the human brain, where complex objects are decomposed into a hierarchy of simpler object parts.
Such a hierarchy is referred to as a parse-tree. Conceptually, capsule neural networks have been defined to realize such parse-trees. The capsule neural network (\mbox{CapsNet}), by Sabour, Frosst, and Hinton, is the first actual implementation of the conceptual idea of capsule neural networks.
\mbox{CapsNets} achieved state-of-the-art performance on simple image recognition tasks with fewer parameters and greater robustness to affine transformations than comparable approaches.
This sparked extensive follow-up research.
However, despite major efforts, no work was able to scale the  \mbox{CapsNet} architecture to more reasonable-sized datasets.
Here, we provide a reason for this failure and argue that it is most likely not possible to scale  \mbox{CapsNets} beyond toy examples.
In particular, we show that the concept of a parse-tree, the main idea behind capsule neuronal networks, is not present in  \mbox{CapsNets}.
We also show theoretically and experimentally that CapsNets suffer from a vanishing gradient problem that results in the starvation of many capsules during training.

  \vspace{1ex}
\end{abstract}

\section{Introduction}


The concept of capsules~\cite{icann/HintonKW11} describes a hypothetical system that parses a complex image scene into a hierarchy of visual entities that stand in part-whole relationship to each other~\cite{nips/HintonGT99}. A capsule is conceptually defined as a highly informative, compact representation of a visual entity or object within an image. The idea of capsules is motivated by the pattern recognition system in the visual cortex of the human brain~\cite{nips/SabourFH17}. There is some psychological evidence that the human object recognition system assigns hierarchical structural descriptions to complex objects by decomposing them into parts~\cite{cogsci/Hinton79}. The theory of recognition by components~\cite{psychorev/biederman1987} proposes that a relatively small set of simple 3D shapes, called geons, can be assembled in various arrangements to form virtually any complex object, which can then be recognized by decomposition into its respective parts~\cite{psychorev/biederman1987}. 

A capsule may represent a visual entity by encapsulating its properties, also known as instantiation parameters, such as position, size, orientation, deformation, texture, or hue. A multi-level assembly of such capsules represents a complex image scene, where lower-level capsules model less abstract objects or object parts, and higher-level capsules model complex and composite objects. Lower-level capsules are connected to higher-level capsules if the corresponding entities are in a part-whole relationship. For a composite object, the hierarchy of capsules defines a syntactic structure like a parse-tree defines the syntactic structure of a sentence. Therefore, the hierarchy of capsules is also referred to as parse-tree. If an object or object part is present in an image, its respective capsule will be present within the parse-tree.

Ideally, the parse-tree is invariant under affine transformations as well as changes of viewpoint. That is, a slightly modified viewpoint on a visual entity should not change a capsule's presence within the parse-tree. Such parse-trees would be highly efficient distributed representations of image scenes~\cite{nips/SabourFH17, nips/HintonGT99}. Also, explainable machine learning can profit from interpretable capsules that stand for dedicated visual entities, and the discrete nature of trees may connect deep learning with a symbolic approach to AI. Furthermore, capsules can be related to inverse graphics, and there is hope that they can lead to debuggable, parameter efficient, and interpretable models with a broad range of applications for all kinds of image-related tasks like image classification or segmentation.

However, capsules are only conceptually defined, and the difficulty is finding an implementation with all the highly-desirable properties from above.
The capsule neural network (CapsNet) by \citet{nips/SabourFH17} aims at such an implementation of the conceptual capsule idea. It was specifically designed to surpass convolutional neural networks (ConvNets)~\cite{neco/LeCunBDHHHJ89} as the latter were found to suffer from several limitations, including a lack of robustness to affine transformations and change of viewpoint, the susceptibility to adversarial attacks, exponential inefficiencies, and a general lack of interpretability in the network's decision-making process. Considering these limitations, the parse-tree sounds particularly appealing with all its advantages.

\subsubsection{Contributions.}
Here, our aim is a thorough investigation of the question, whether the CapsNets implementation as proposed by~\citet{nips/SabourFH17} realizes all the conceptual ideas that make capsule networks so appealing. We summarize this in two key assumptions. The \textbf{first key assumption} is that the CapsNet learns to associate a capsule with a dedicated visual entity within an input image~\citep{nips/SabourFH17}. The \textbf{second key assumption} is that the CapsNet's capsules can be organized hierarchically in a parse-tree that encodes part-whole relationships. We test both assumptions experimentally and have to reject them. We show that the \mbox{CapsNet} does not exhibit any sign of an emerging parse-tree. Thus, the \mbox{CapsNet} implementation cannot provide the theoretical benefits of capsule networks like invariance under affine transformations and change of viewpoint. Furthermore, we provide a theoretical analysis, exposing a vanishing gradient problem, that supports our experimental findings.

\section{Related Work}\label{sec:related_work}
Early references to the hierarchy of parts appear already in~\citep{cogsci/Hinton79}. The idea of parsing images into parse-trees was proposed by~\citet{nips/HintonGT99} and the concept of capsules was established in~\citep{icann/HintonKW11}. An important addition by the \mbox{CapsNet}~\citep{nips/SabourFH17} was the routing-by-agreement (RBA) algorithm that creates capsule parse-trees from images. With its introduction, the \mbox{CapsNet} demonstrated state-of-the-art classification accuracy on~MNIST~\cite{dataset/lecun10} with fewer parameters and stronger robustness to affine transformations than the \mbox{ConvNet} baseline, which sparked a flood of follow-up research. This includes different routing mechanisms, such as EM-Routing~\citep{iclr/HintonSF18}, Self-Routing~\citep{nips/HahnPK19}, Variational Bayes Routing~\cite{aaai/RibeiroLK20}, Receptor Skeleton~\cite{icml/ChenYQC021} and  attention-based routing~\citep{nips/AhmedT19, iclr/TsaiSGS20, nature/Mazzia21, aaai/Gu21}. \citet{iclr/Wang018} reframed the routing algorithm in~\cite{nips/SabourFH17} as an optimization problem, and \citet{corr/Rawlinson18} introduced an unsupervised learning scheme for \mbox{CapsNets}. Other work replaced the capsule vector representations by matrices~\cite{iclr/HintonSF18} or tensors~\cite{cvpr/RajasegaranJJJS19}, or added classic \mbox{ConvNet} features to the general routing mechanisms, such as dropout~\cite{spl/XiangZTZX18} or skip-connections~\cite{cvpr/RajasegaranJJJS19}. The GLOM architecture, which was proposed by \cite{corr/Hinton21}, suggests a routing-free approach for creating parse-trees from images, but has not been implemented yet. Furthermore, other publications focus on learning better first layer capsules~(PrimeCaps), such as the Stacked Capsule Autoencoders~\cite{nips/KosiorekSTH19}  and Flow Capsules~\cite{icml/SabourTYHF21}.

However, after a while it turned out that the \mbox{CapsNet} falls short of the anticipated benefits and promises of the capsule idea. To this date, \mbox{CapsNets} do not scale beyond small-scale datasets. Works that empirically report scaling issues include~\citep{corr/XiBingJin17, acml/PaikKK19}. Although the \mbox{CapsNet} was introduced in the realm of computer vision, the best performing capsule implementation~\cite{nips/AhmedT19} achieves only $60.07\%$ \mbox{top-1} image classification accuracy on ImageNet~\cite{data/imagenet}, far behind state-of-the-art transformer-based approaches~\cite{icml/Wortsman22} and  \mbox{ConvNets}~\cite{cvpr/Pham21} with $90.88\%$ and  $90.02\%$ accuracy respectively. The original \mbox{CapsNet} itself has not been demonstrated to work on ImageNet.

Further negative results regarding \mbox{CapsNets} emerged, questioning the promised benefits and technical progress altogether. \citet{acml/PaikKK19} observed that increasing the depth of various \mbox{CapsNet} variants did not improve accuracy, and routing algorithms, the core components of capsule implementations, do not provide any benefit regarding accuracy in image classification. \citet{corr/Michels19}, and \citet{iclr/Gu2021} showed that \mbox{CapsNets} can be as easily fooled as \mbox{ConvNets} when it comes to adversarial attacks. \citet{cvpr/GuT021} showed that the individual parts of the \mbox{CapsNet} have contradictory effects on the performance on different tasks and conclude that with the right baseline, \mbox{CapsNets} are not generally superior to \mbox{ConvNets}. Finally, \citet{cvpr/GuT20} showed that removing the dynamic routing component improves affine robustness, and \citet{corr/Rawlinson18} show that capsules do not specialize without supervision.

Here, we explain these shortcomings, which can be attributed to a lack of an emerging parse-tree.

\section{The Capsule Neural Network}
\label{sec:capsule_neural_network}

\begin{figure}[!ht]
	\vspace{2ex}
	
	\includegraphics[width=0.48\textwidth]{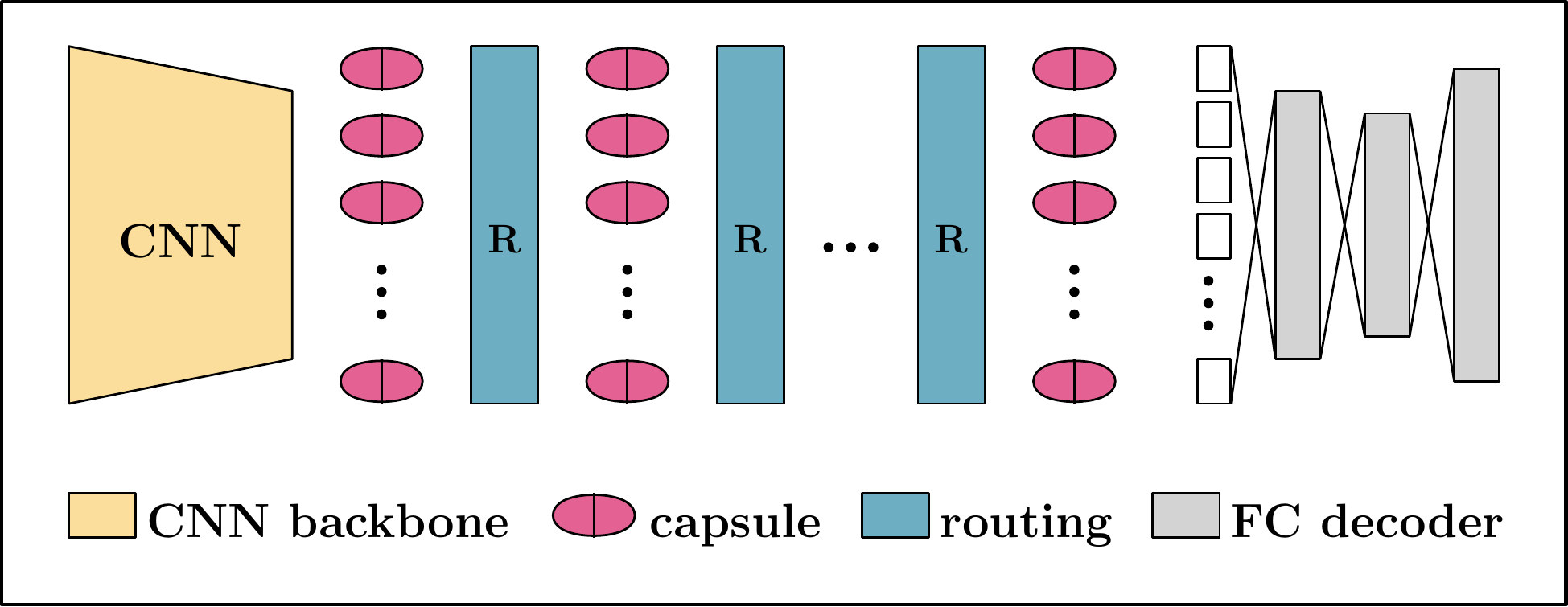}
	\caption{A generic CapsNet architecture.}
	\label{fig:cn:architecture}
\end{figure}

The CapsNet implements capsules as parameter vectors.
An illustration of the CapsNet architecture, which consists of a multi-layer hierarchy of capsules, is shown in Figure~\ref{fig:cn:architecture}. In the following, we introduce basic notations and definitions, the generic CapsNet architecture, and a loss function for training \mbox{CapsNets}. Furthermore, we discuss how \mbox{CapsNets} implement the crucial concept of a parse-tree.

\subsection{Notation}

\textbf{Capsules.}\quad The matrix $U^l\in \mathbb{R}^{n^l\times d^l}$ holds $n^l$ normalized capsules of dimension $d^l$ for layer $l\in \{1, 2, \ldots, \ell\}$. The $i$-th capsule in $U^l$ is the vector $u^l_{(i,:)}\in\mathbb{R}^{d^l}$, and $u^l_{(i,j)}\in\mathbb{R}$ is the $j$-th entry of capsule $i$ on layer $l$.\\
\textbf{Transformation matrices.}\quad The tensor $W^l\in\mathbb{R}^{n^{l+1}\times n^l\times d^{l+1}\times d^l}$ holds transformation matrices $W^l_{(j,i,:,:)}\in\mathbb{R}^{d^{l+1}\times d^l}$. The transformation matrix $W^l_{(j,i,:,:)}$  maps the $i$-th capsule of layer $l$ to its unnormalized contribution to the $j$-the capsule of layer $l+1$.\\
\textbf{Coupling coefficients.\quad }The matrix $C^l\in \mathbb{R}^{n^l\times n^{l+1}}$ holds coupling coefficients for the connections of capsules from layer $l$ to layer $l+1$. The entry $c^l_{(i,j)}\in [0,1]$ specifies the coupling strength between capsule $i$ on layer $l$ and capsule $j$ on layer $l+1$. The coupling coefficients satisfy $\sum_{j=1}^{n^{l+1}}  c^l_{(i,j)} =1$ for all $i\in\{1, 2, \ldots ,n^l\}$.\\
\textbf{Squashing function.}\quad The squashing function normalizes the length of a capsule vector $u\in\mathbb{R}^d$ into the range $[0,1)$. Here, we use a slightly modified squashing function~\citep{nature/Mazzia21}, 
\begin{equation}
g(u) = \left( 1 - \frac{1}{\exp (\| u\|_2)}\right) \frac{u}{\| u\|_2} 
\label{eq:squash}
\end{equation}
that behaves similarly to the original squashing function proposed by~\citet{nips/SabourFH17}, but is more sensitive to small changes near zero \cite{corr/XiBingJin17}, which is required to stack multiple layers of capsules. 

\subsection{Architecture}

First, the backbone network extracts features from an input image into a feature matrix in $\mathbb{R}^{n^1\times d^1}$. The feature matrix is then normalized by applying the squashing function to each row, which constitutes the first capsule layer in $U^1\in \mathbb{R}^{n^1\times d^1}$. The capsules in $U^1$ are also called PrimeCaps. Starting from the PrimeCaps, consecutive layers of capsules are computed as follows: First, the linear contribution of capsule $i$ on layer $l$ to capsule $j$  on layer $l+1$ is computed as
\begin{equation}
\hat u^{l+1}_{(i,j,:)} = W^l_{(j,i,:,:)} u^l_{(i,:)}, 
\label{eq:votes}
\end{equation}
where the entries in the matrix $\hat U^{l+1}_{(i)}$, which holds the vectors $\hat u^{l+1}_{(i,j,:)}$, are called \textbf{votes} from the $i$-th capsule on layer $l$. An upper layer capsule $u^{l+1}_{(j,:)}$ is the squashed, weighted sum over all votes from lower layer capsules $u^l_{(i,:)}$, that is,
\[
u^{l+1}_{(j,:)} = g\left(\sum_{i=1}^{n^l} c^l_{(i,j)} \cdot \hat{u}^{l+1}_{(i,j,:)}\right),
\]
where the the coupling coefficients $c^l_{(i,j)}$ are dynamically computed, that is, individually for every input image, by the Routing-by-agreement Algorithm (RBA), see Algorithm~\ref{alg:capsnet_routing}.

\begin{algorithm}[!ht]
	\caption{Routing-by-agreement (RBA)}\label{alg:capsnet_routing}
	\textbf{Input}: votes $\hat{u}$, number of iterations $r$, routing priors $b$  \\
	\textbf{Output}: coupling coefficients $c$
	\begin{algorithmic}[1]
		\For{$r$ iterations}
		\State $c_{(i,j)} \gets \frac{\exp(b_{(i,j)})}{\sum_{k}\exp(b_{(i,k)})}$
		\State $v_{(j,:)} \gets g\left(\sum_{i} c_{(i,j)} \hat{u}_{(i,j,:)}\right)$
		\State $b_{(i,j)} \gets b_{(i,j)} + \sprod{\hat{u}_{(i,j,:)}}{v_{(j,:)}}$
		\EndFor
	\end{algorithmic}
\end{algorithm}

\begin{figure}[!ht]
	\begin{center}
		\begin{tabular}{ccc}
			\includegraphics[width=0.29\linewidth]{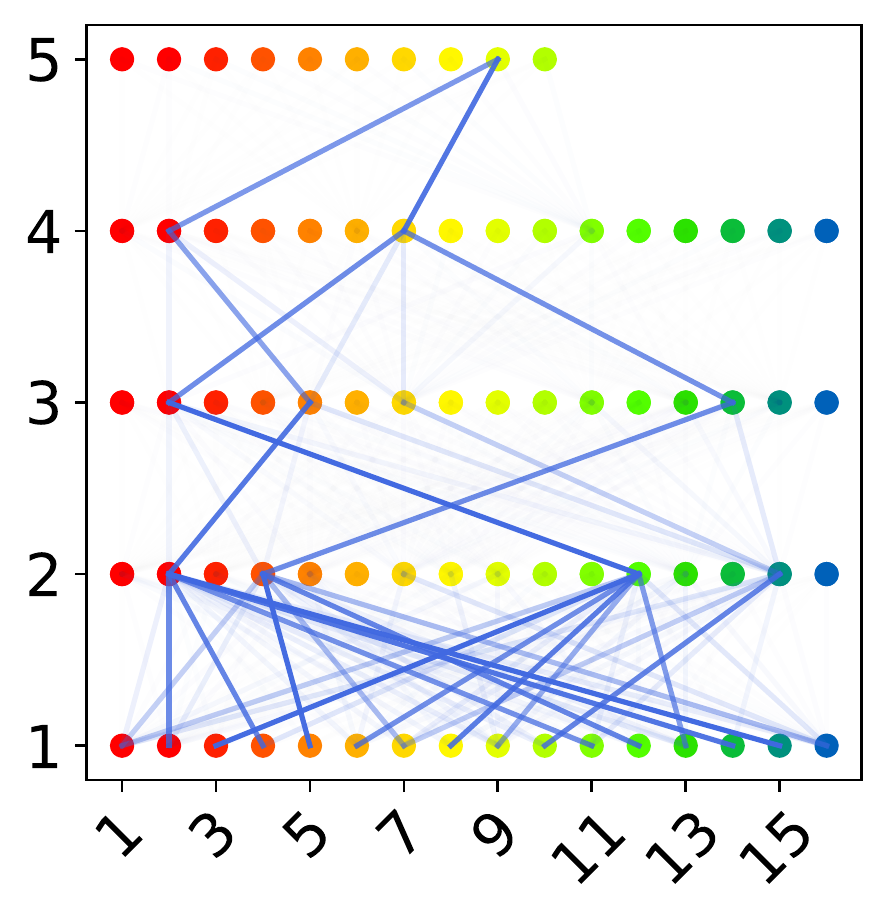} &
			\includegraphics[width=0.29\linewidth]{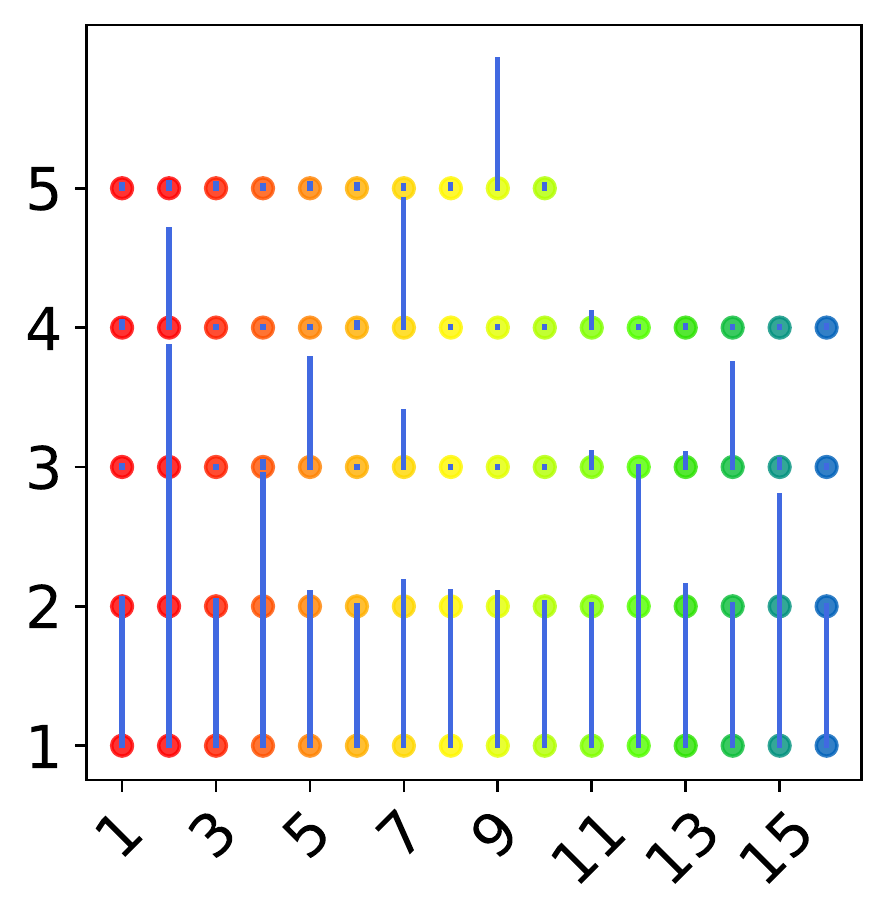} &
			\includegraphics[width=0.29\linewidth]{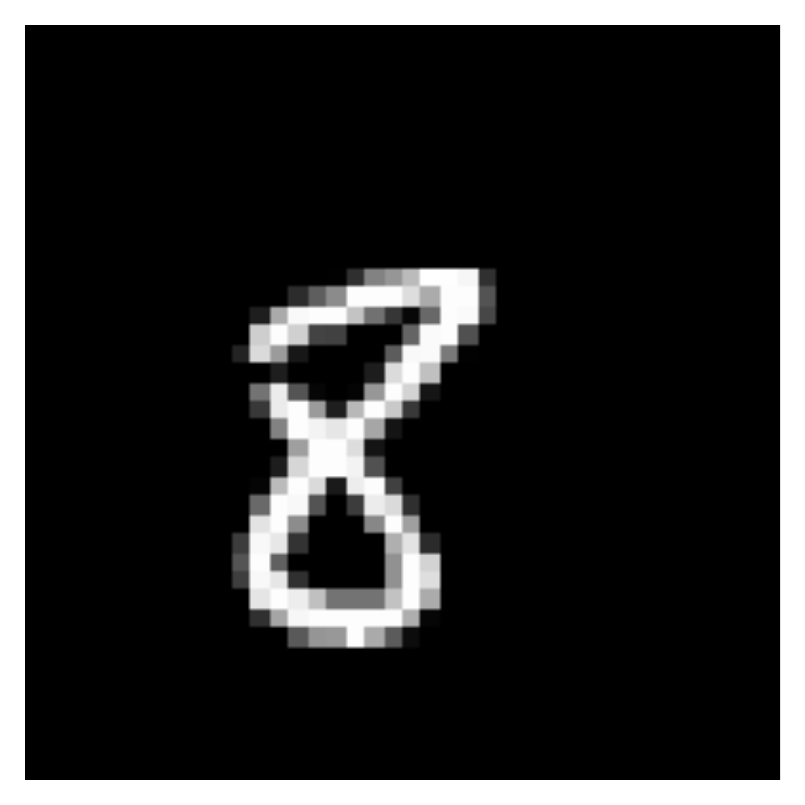} \\
			\includegraphics[width=0.29\linewidth]{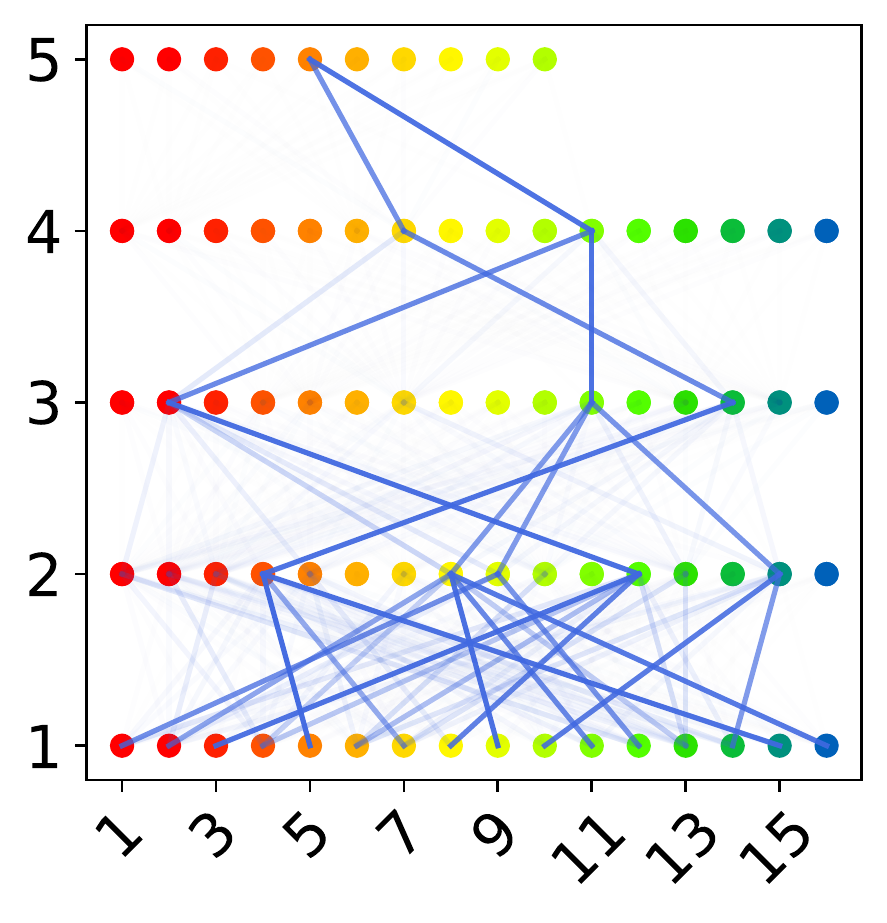} &
			\includegraphics[width=0.29\linewidth]{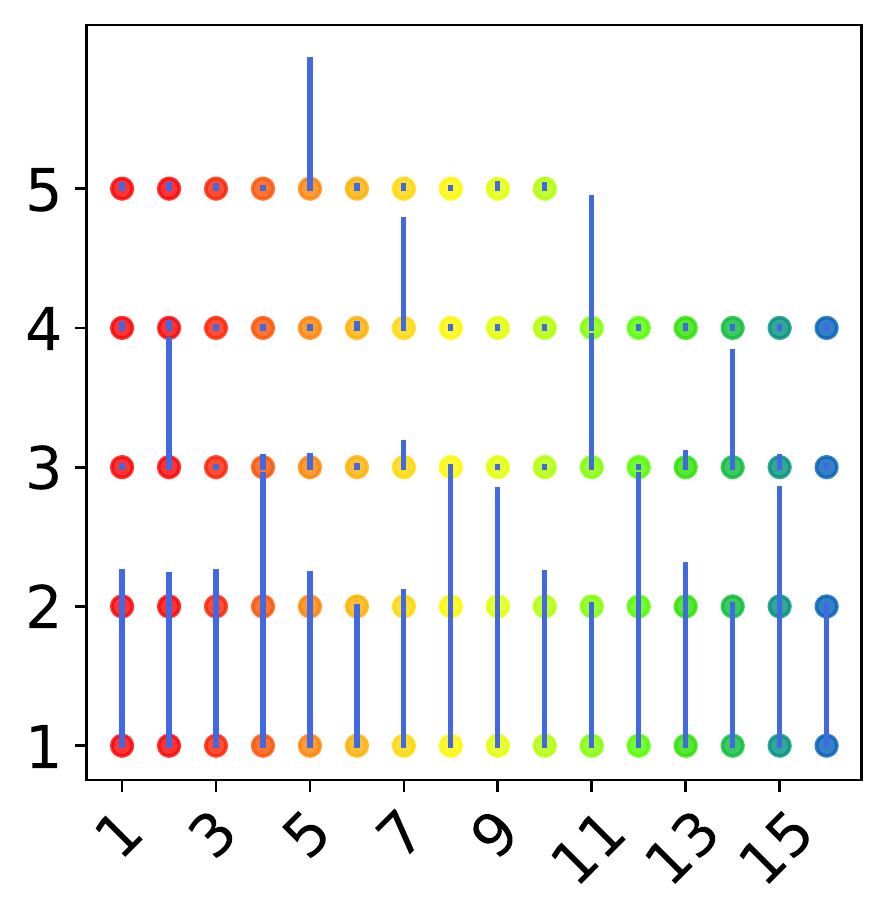} &
			\includegraphics[width=0.29\linewidth]{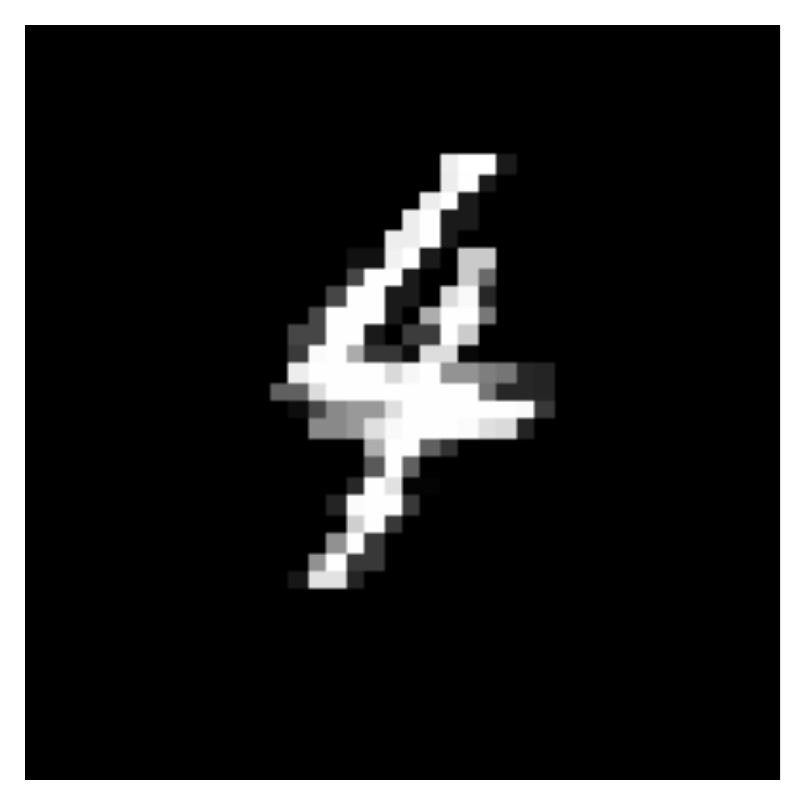} 
		\end{tabular}
	\end{center}
	\caption{Fuzzy parse-trees for images from the AffNIST dataset for a model with five capsule layers, 16 capsules on each intermediate layer, and ten on the last layer.
		The figure shows the coupling coefficients as connections between capsules (left), the capsule norms/activations (middle), and the input image (right). The blue tone of the edges is darker for greater coupling coefficients. }
	\label{fig:affnist:main_model:class_samples}
\end{figure}

The number of output capsules on the last layer $\ell$ is set to match the number of classes in the respective dataset.
Finally, the fully-connected decoder network reconstructs the input image from the capsules on layer $\ell$.

\subsection{Training}

The parameters in the backbone network, in the reconstruction network, as well as the transformation tensors $W^l$ and the RBA routing priors $b_{(i,j)}^ l$ are all learned by minimizing a weighted sum of a supervised classification loss $L_m$ and an unsupervised reconstruction loss $L_r$, that is, $L = L_m + \alpha\cdot L_r$, with $\alpha > 0$. The classification loss function 
\begin{align}
L_m = \sum_{j=1}^{n^\ell} &t_j \cdot \max \{ 0, m^+ - \|u^\ell_{(j,:)}\|_2 \}^2 \nonumber \\ &+\lambda\cdot (1-t_j)\cdot \max\{0, \|u^\ell_{(j,:)}\|_2  - m^-\}^2 \label{eq:margin_loss}
\end{align}
is only applied to the output capsules. Here $m^+, m^- > 0$ and $\lambda > 0$ are regularization parameters, and $t_j$ is $1$ if an object of the $j$-th class is present in the input image, and~$0$ otherwise.
Output capsules that correspond to classes not present in the input image are masked by zeros.
The reconstruction loss function is applied to the output of the reconstruction network and sums the 
distances between the reconstruction and the pixel intensities in the input image.

\begin{figure*}[!ht]%
	\centering
	\subfloat[mean couplings]
	{\includegraphics[width=0.19\linewidth]{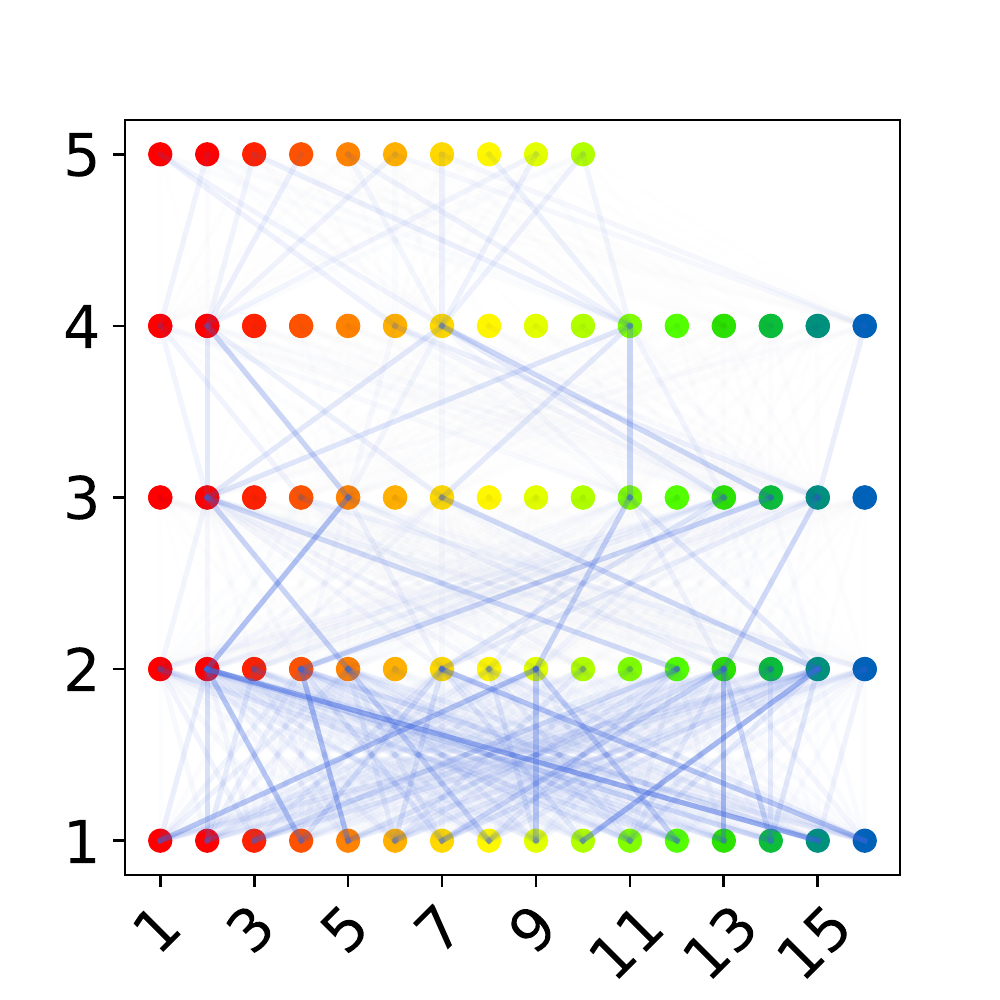}\label{fig:affnist:main_model:parse_tree_stats:mean_coupling}}
	\subfloat[std.\ couplings]
	{\includegraphics[width=0.19\linewidth]{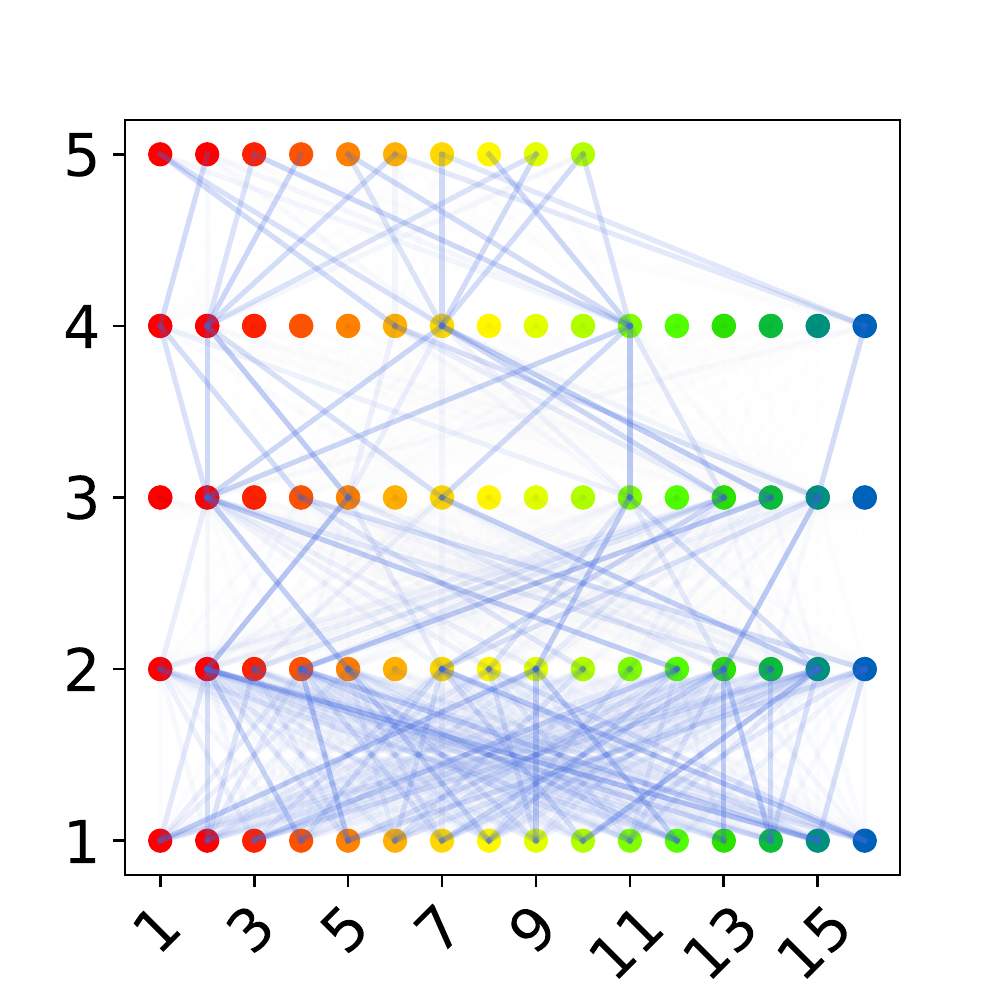}\label{fig:affnist:main_model:parse_tree_stats:std_coupling}}
	\subfloat[mean activations]
	{\includegraphics[width=0.19\linewidth]{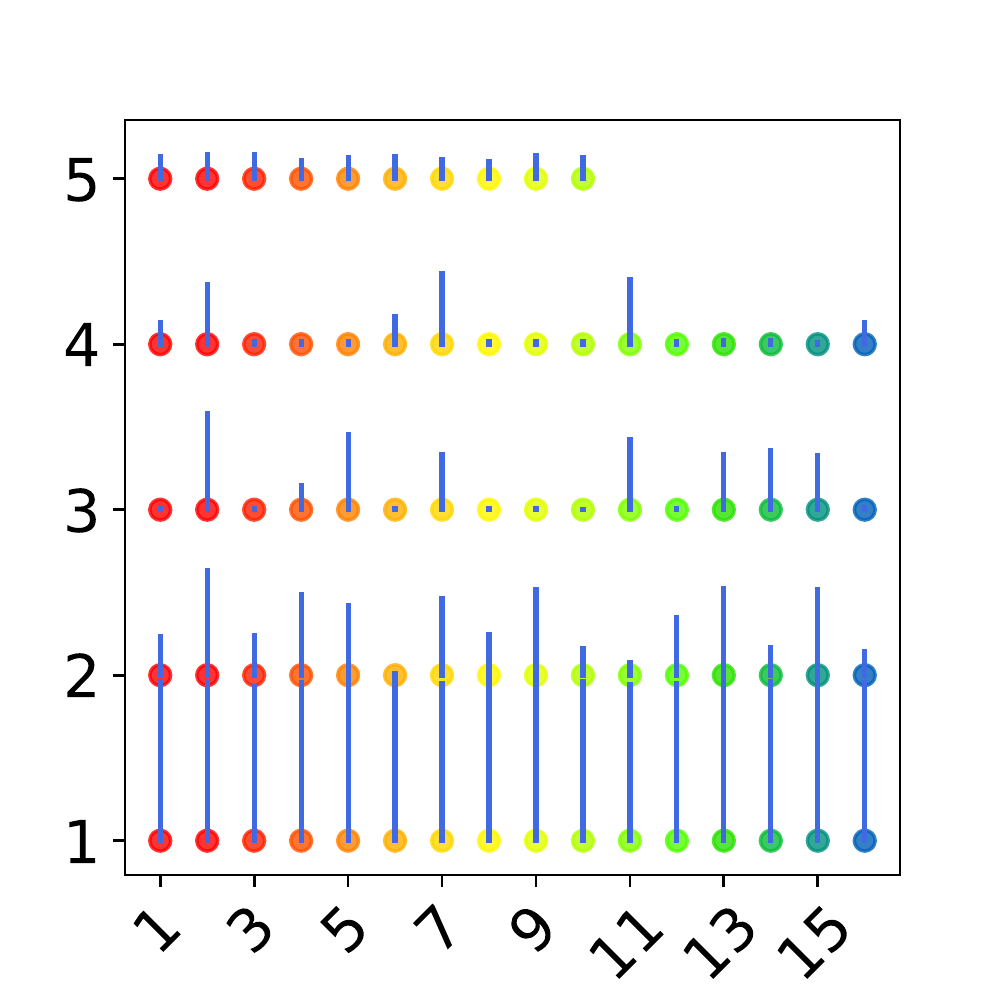}\label{fig:affnist:main_model:parse_tree_stats:mean_capsule}}
	\subfloat[std.\ activations]
	{\includegraphics[width=0.19\linewidth]{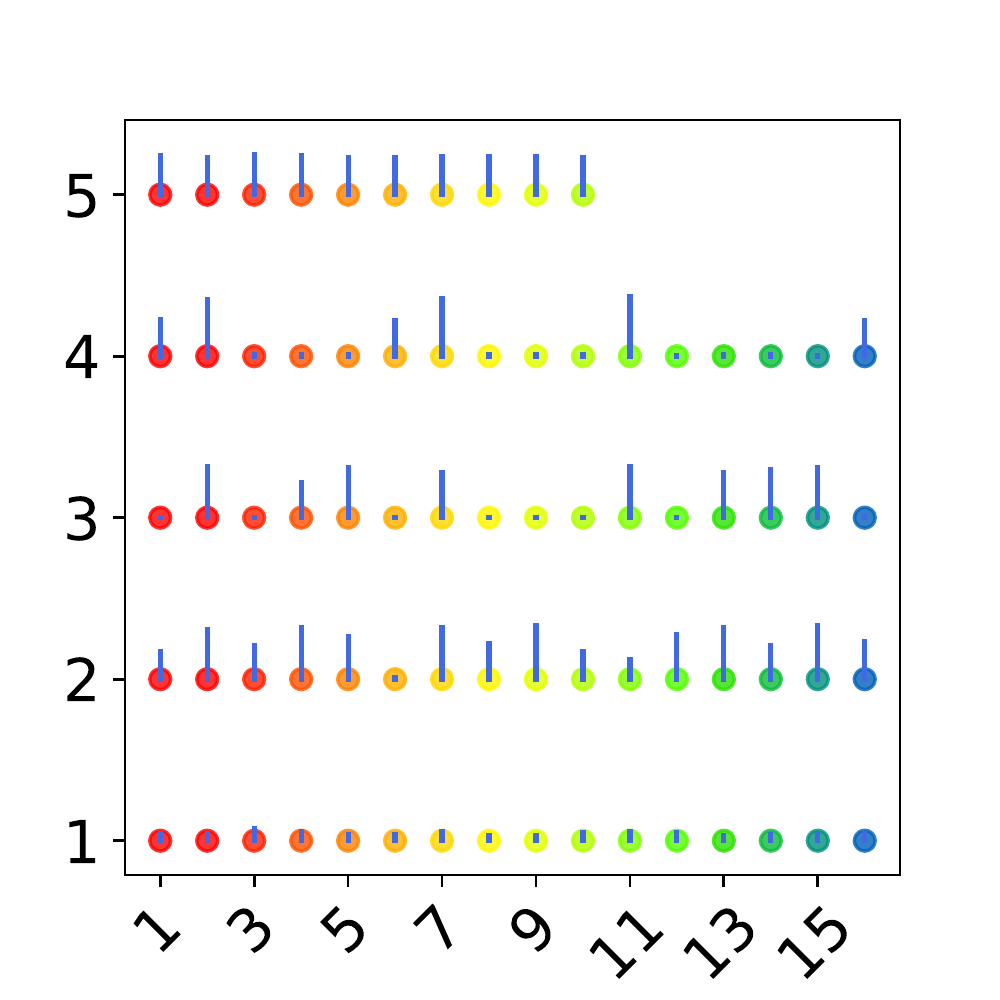}\label{fig:affnist:main_model:parse_tree_stats:std_capsule}}
	\subfloat[dead capsules]
	{\includegraphics[width=0.19\linewidth]{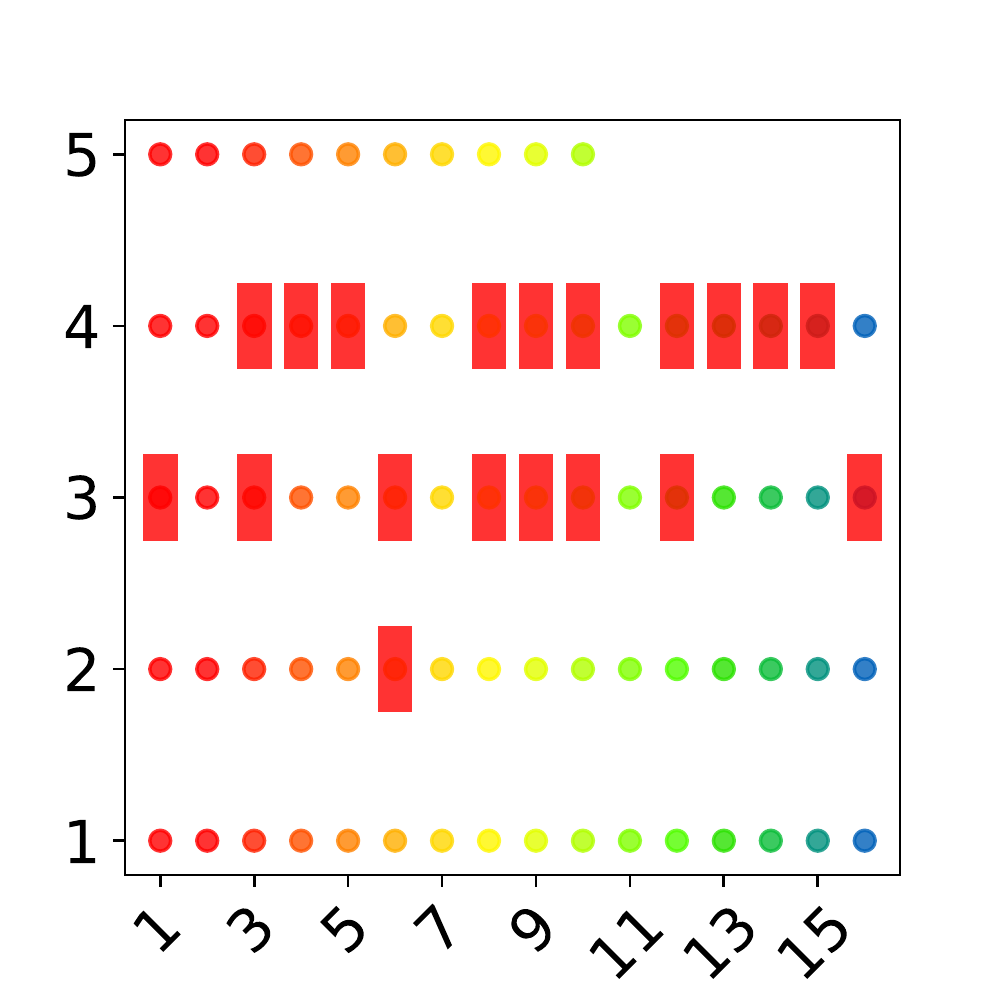}\label{fig:affnist:main_model:parse_tree_stats:dead_capsule}}
	\caption{
		Parse-tree statistics for the complete AffNIST validation dataset for a five-layer CapsNet model with 16/10 capsules. The mean~(a) 
		and the standard deviation~(b) 
of the coupling coefficient matrices for each layer are visualized as connections between capsules. Higher coupling coefficients have a darker blue tone. The capsule norms' mean~(c) 
and standard deviation~(d) 
are visualized by bars. Dead capsules~(e)  
are highlighted with a red bar.
	}%
	\label{fig:affnist:main_model:parse_tree_stats}
\end{figure*}

\subsection{Parse-Trees}

The parse-tree is the most important concept that allows us to understand and interpret capsules and their connections. The CapsNet defines a parse-tree, where capsules represent nodes, and coupling coefficients represent fuzzy edges.

The capsules magnitude, that is, the norm of the parameter vector, which is always in $[0,1)$ after applying the squashing function, represents the probability that the corresponding entity is present in an input image. The capsules direction represents instantiation parameters of the entity like its position, size, or orientation. Changing the viewpoint on an entity does not affect its presence, but only its instantiation parameters. Therefore, the respective capsule's magnitude should be unaffected, whereas its direction can change.

In the CapsNet, the dynamic part-whole relationships are implemented by coupling coefficients between capsules on consecutive layers. The coupling coefficients in the routing layers are computed dynamically by the RBA Algorithm~\ref{alg:capsnet_routing}. Taking the row-wise \emph{softmax} ensures that the coupling coefficients in $c^l_{(i,:)}$ are positive and sum up to one. Therefore, we can view the coupling coefficients as fuzzy edges that connect capsules $u^l_{(i,:)}$ and $u^{l+1}_{(j,:)}$ with probability $c^l_{(i,j)}$. The multi-layer hierarchy of capsule nodes, connected by fuzzy edges, defines the parse-tree analogously to a probabilistic context-free grammar. Examples are shown in Figure~\ref{fig:affnist:main_model:class_samples}.

\section{Challenging the Parse-Tree Assumption}
As mentioned in the introduction, there are two key assumptions regarding the parse-tree: 
(1) The nodes of the parse-tree, the activated capsules, are viewpoint-invariant representations of visual entities present in the input image.
(2) Lower-level capsules represent object parts,  higher-level capsules represent composite objects, and part-whole relationships are represented by the edges of the parse tree, that is, by the coupling coefficients. 
In the following, we are going to challenge both assumptions.

If Assumption~(2) holds, then the parse-tree computed by a CapsNet is a part-whole representation of the image scene, and the routing dynamics defined by the coupling coefficients is specific to the input image.
We conduct experiments challenging this assumption in Section~\ref{sec:routing}.

If Assumption~(1) holds, then affine transformations of an image only change the direction of a parameter vector of a capsule, but not its magnitude. Hence, we take an image, transform it affinely, and analyze the resulting capsule activations. These experiments can be found in Section~\ref{sec:viewpoint}. Furthermore, we examine the capsule activation in general in Section~\ref{sec:starvation}.

\subsubsection{Experimental Setup}
We use the AffNIST benchmark~\cite{nips/SabourFH17} to assess a model's robustness to affine transformations, and we use the \mbox{CIFAR10} dataset~\cite{report/Krizhevsky09} to test a model's performance on complex image scenes. 
We conduct extensive experiments using a total of 121 different model architectures of various scales, featuring different numbers of routing layers and different numbers and capsule dimensions. Shallow models resemble the original CapsNet implementation while deeper models allow for a more semantically expressive parse-tree.

We refer to the appendix for detailed architecture and dataset descriptions, training procedures, and full results for all models used in our experiments.

\subsection{Routing Dynamics}
\label{sec:routing}
We measure the diversity of parse-trees, that is, the routing dynamics, by assessing the diversity of routing targets for a single capsule $u$. For $k$ input images, let $C\in  \mathbb{R}^{k \times n}$ hold all the coupling coefficients that connect $u$ to capsules on the next layer, which contains $n$ target capsules. We use the standard deviation $\std(c_{(:, i)})$ with respect to all input images as a measure for the routing diversity of $u$ to the $i$-th capsule on the next layer. A routing $\pr_n$ is called \textbf{perfect} if it always routes to exactly one capsule on the next layer and routes to all $n$ capsules on the next layer equally likely. The standard deviation of a perfect routing computes to $\std(pr_n) = \sqrt{\left(1 - \frac{1}{n}\right) \frac{1}{n}}$. We use a perfect routing to define the \textbf{dynamic routing coefficient (dyr)} for capsule $u$ as
\[
\dyr(u) = \frac{1}{n} \sum_{i=1}^{n} \frac{\std(c_{(:, i)})}{\std(\pr_n)}
\]
The expected number of target capsules \textbf{(dys)} for $u$ is $\dys(u) = n \cdot \dyr(u)$. For a whole layer, we define the coefficients \textbf{dyr} and \textbf{dys} as the mean over all capsules of this layer.

\subsubsection{Results}
In the following, we report the routing statistics for a CapsNet architecture with four routing layers, 16 capsules per layer in the first four layers, and ten capsules in the last layer. We set the capsule dimension to eight and train multiple models on the AffNIST dataset until a target accuracy of $99.2\%$ is reached. 
The routing statistics for the models are summarized in Table~\ref{tab:affnist:main_model:dynamics} and the corresponding coupling coefficients of a single model are visualized in Figure~\ref{fig:affnist:main_model:parse_tree_stats}. 
The $\dys$ values below two for Layers~2 and~3 indicate low routing dynamics.
A route is mostly predetermined once a capsule is activated; hence, the routing is almost static.
Only the last layer exhibits higher routing dynamics, which can be attributed to the supervisory effect of the classification loss $L_m$.
Since the routing is almost static, we conclude that the parse-trees do not encode the information necessary for a distributed representation of diverse image scenes, violating Assumption~(2).
As can be seen from Tables~\ref{tab:affnist:model_d1:best} to~\ref{tab:affnist:model_d6:best} in the supplement, the results look similar for all models trained on AffNIST. The more complex data set CIFAR10 exhibits even worse routing dynamics; see Table~\ref{tab:cifar:main_model:activation_dynamics} and Figure~\ref{fig:cifar:main_model:training:parse_tree} in the appendix.

\begin{table}[!ht]
	\centering
	\begin{tabular}{ccc|cc}
		\toprule
		\multirow{2}{*}{Layer} &
		\multirow{2}{*}{Capsules Alive} &
		\multicolumn{2}{c}{Routing Dynamics} \\
		&  & Rate ($\dyr$) & Mean ($\dys$) \\
		\midrule
		1 & 16.00 $\pm$ 0.00 & 0.30 $\pm$ 0.00 & 4.50 $\pm$ 0.17 \\
		2 & 14.90 $\pm$ 0.70 & 0.25 $\pm$ 0.01 & 1.72 $\pm$ 0.11 \\
		3 & 7.00 $\pm$ 0.63  & 0.30 $\pm$ 0.02 & 1.79 $\pm$ 0.16 \\
		4 & 5.90 $\pm$ 0.70  & 0.38 $\pm$ 0.04 & 3.78 $\pm$ 0.38 \\
                output & 10.00 $\pm$ 0.00\\
		\bottomrule
	\end{tabular}
	\caption{The routing statistics for a CapsNet with four routing layers, 16/10 capsules per layer, and a capsule dimension of eight. We separately train and evaluate ten models on AffNIST the same way and report the mean and standard deviation.}
	\label{tab:affnist:main_model:dynamics}
\end{table}

\subsection{Viewpoint Invariance}
\label{sec:viewpoint}
We investigate to which degree capsule activations are invariant to affine transformations of the input images.
Let $x^{(1)}$ and $x^{(2)}$ show the same visual entity though differently instantiated.
For one specific capsule, let $u^{(1)}, u^{(2)}\in\setR^d$ be the corresponding capsule responses for the two images.  
For a viewpoint invariant parse-tree, it holds that $\norm{u^{(1)}}_2 = \norm{u^{(2)}}_2$, since $\norm{u}_2$ measures the probability that the visual entity is present in the image.
We repeat this process for a set of $k$ input images and collect the corresponding capsule activations in the two vectors $v^{(1)}$ and $v^{(2)} \in\setR^k$. We compute the empirical cross-correlation between these two vectors as $\frac{(v^{(1)})^\top v^{(2)}}{\norm{v^{(1)}}_2\cdot \norm{v^{(2)}}_2}$. For one layer, we compute the average of this value over all capsules of this layer. For a viewpoint-invariant parse-tree, this value should be one.

\begin{figure}[!ht]
	\centering
	\subfloat[rotation]{
	\includegraphics[width=0.41\linewidth]{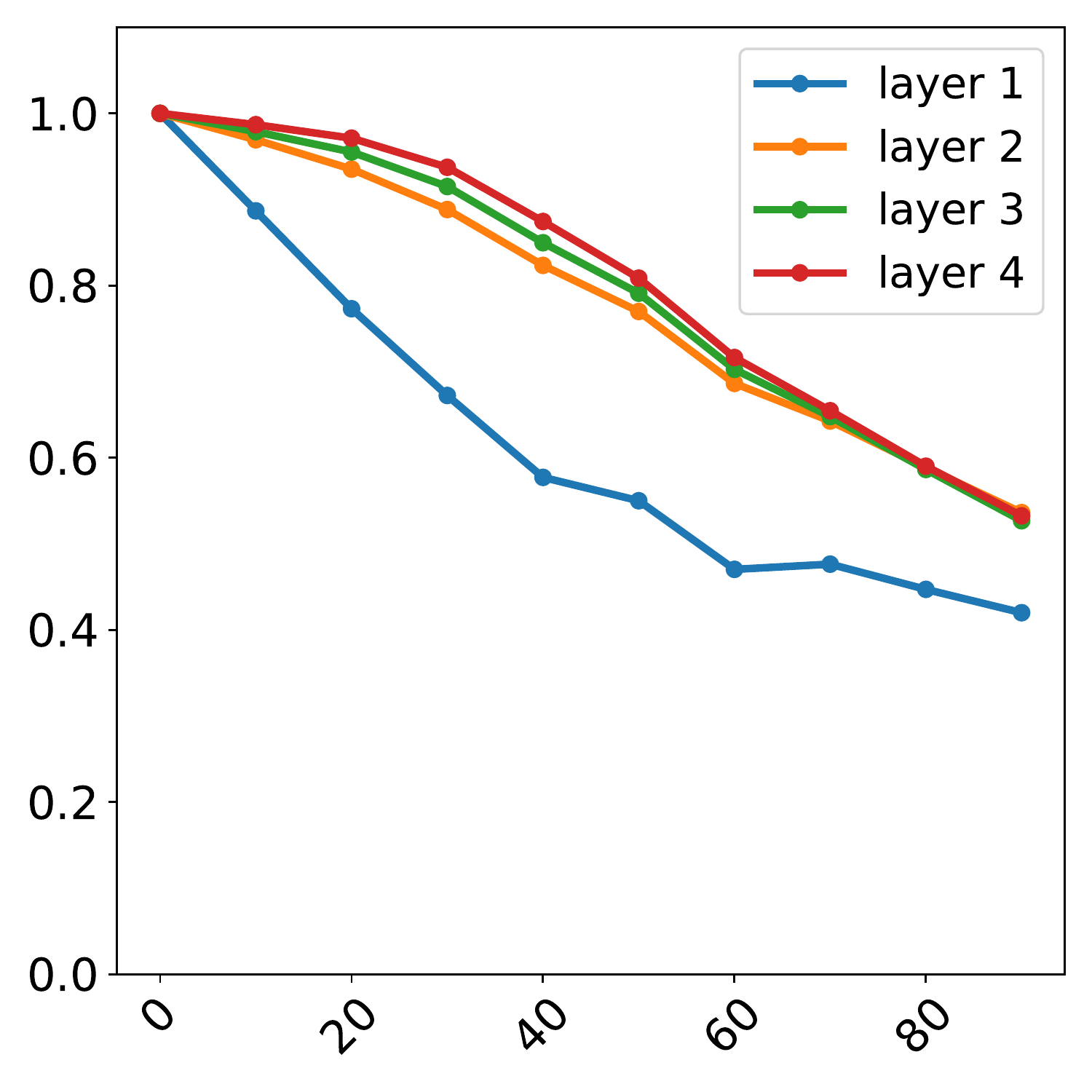}}
	\subfloat[scaling]{
	\includegraphics[width=0.41\linewidth]{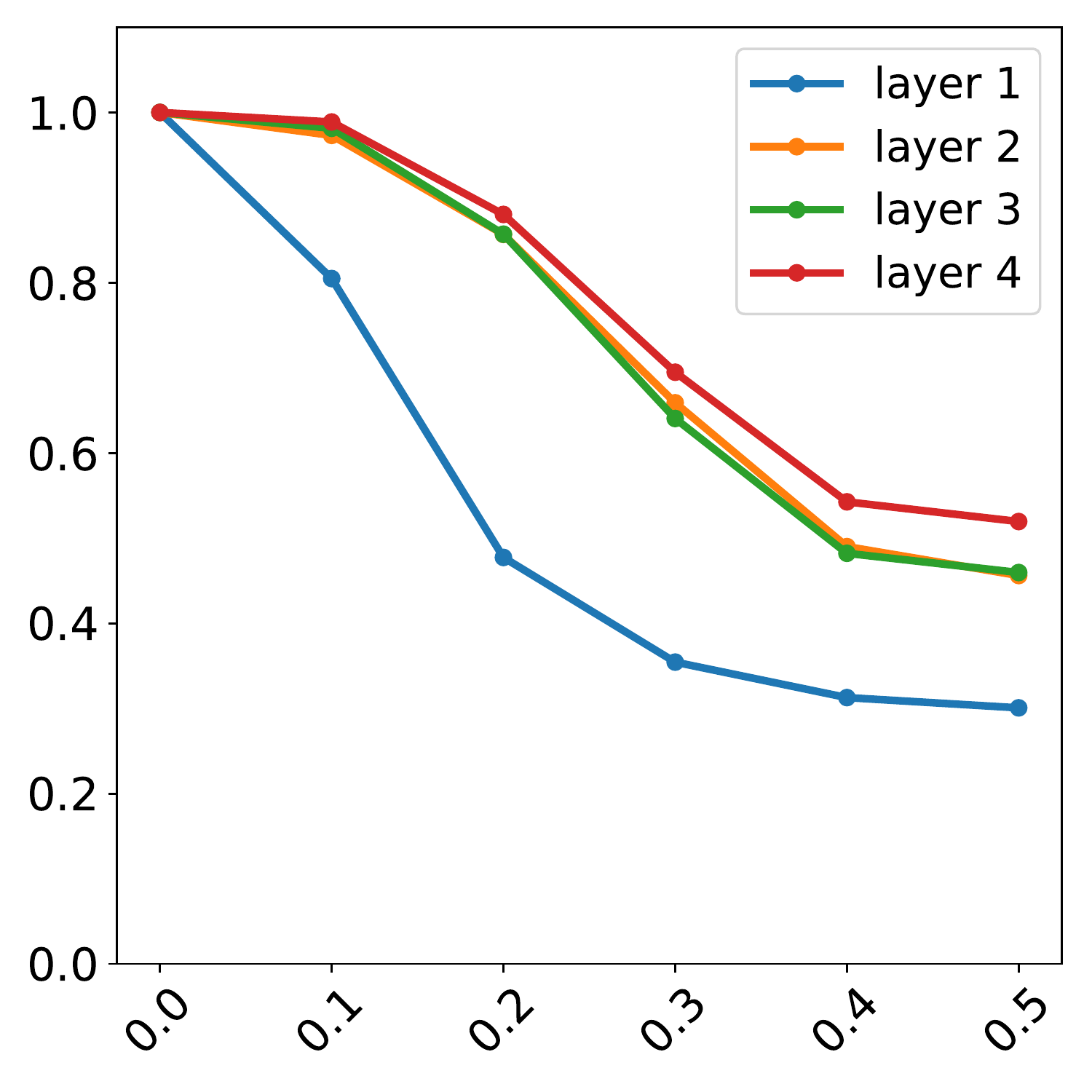}}\\
	\subfloat[horizontal translation]{
	\includegraphics[width=0.41\linewidth]{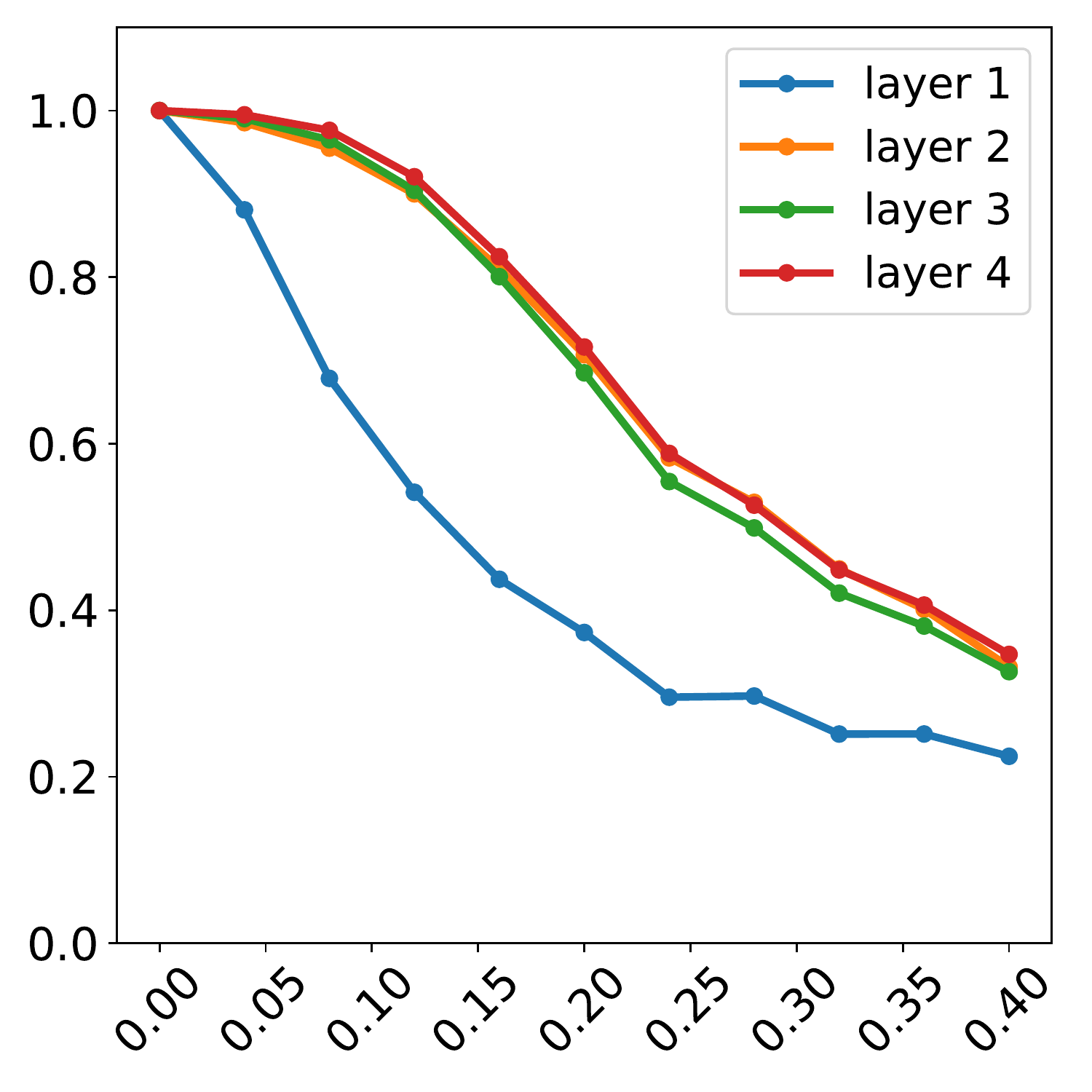}}
	\subfloat[vertical translation]{
	\includegraphics[width=0.41\linewidth]{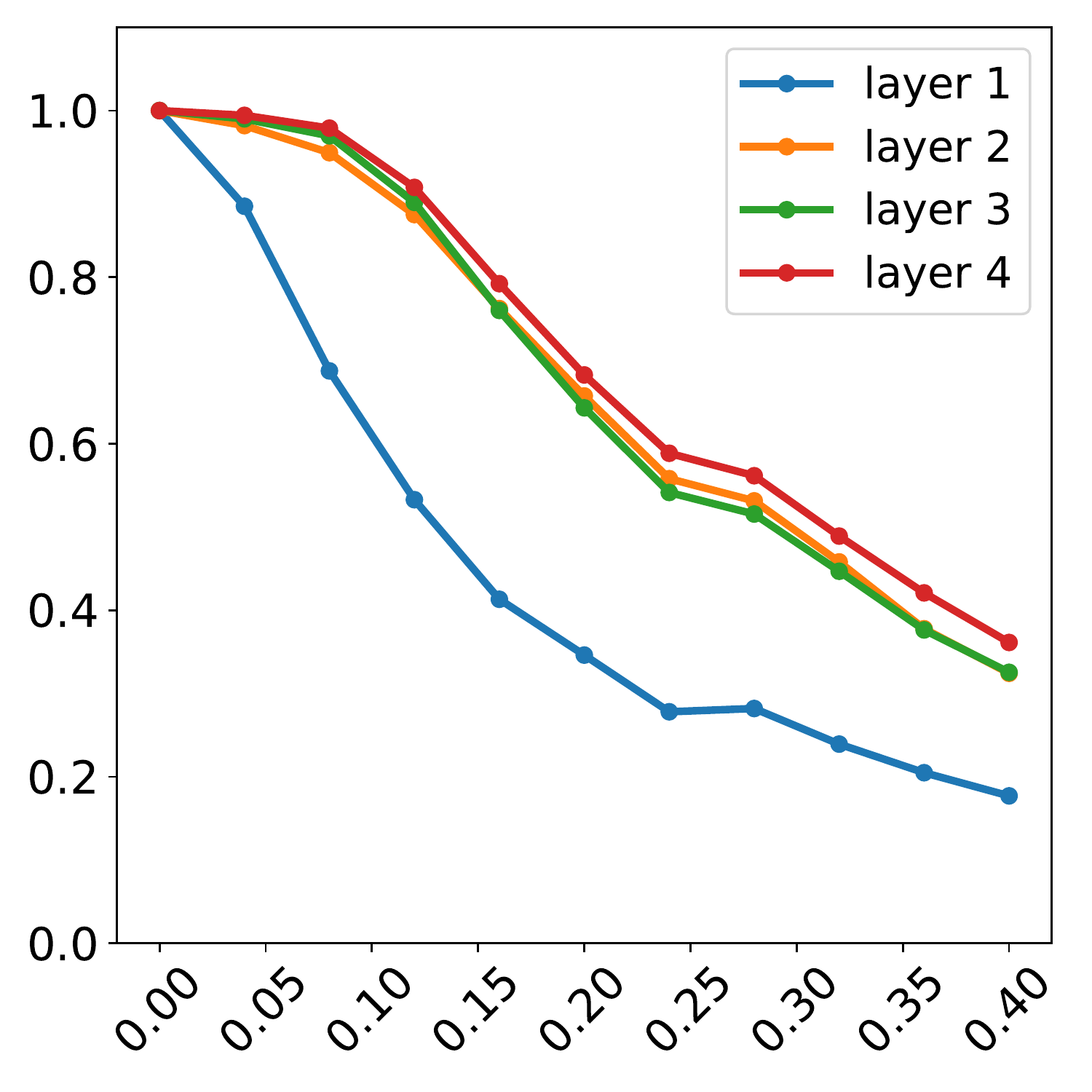}}
	\caption{The capsule activation correlations for each layer with respect to increasing degree of affine transformations.}
	\label{fig:affnist:main_model:view_point_invariance}
\end{figure}

\begin{figure}[!ht]
	\begin{center}
		\begin{tabular}{ccc}
			\includegraphics[width=0.29\linewidth]{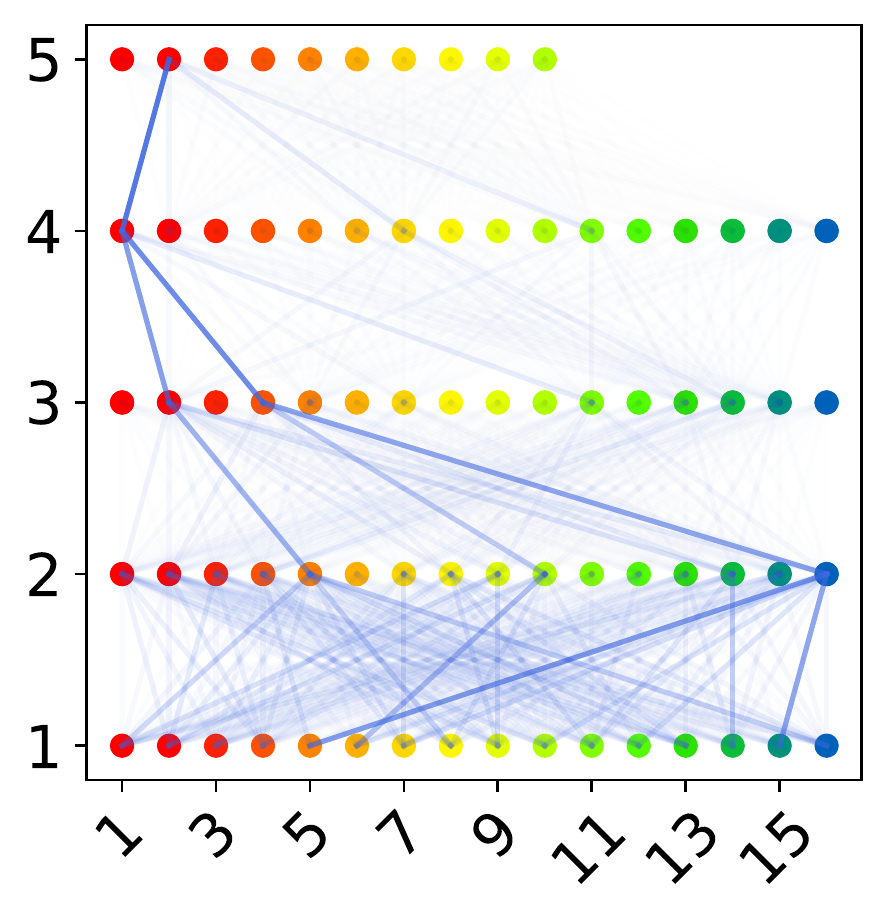} &
			\includegraphics[width=0.29\linewidth]{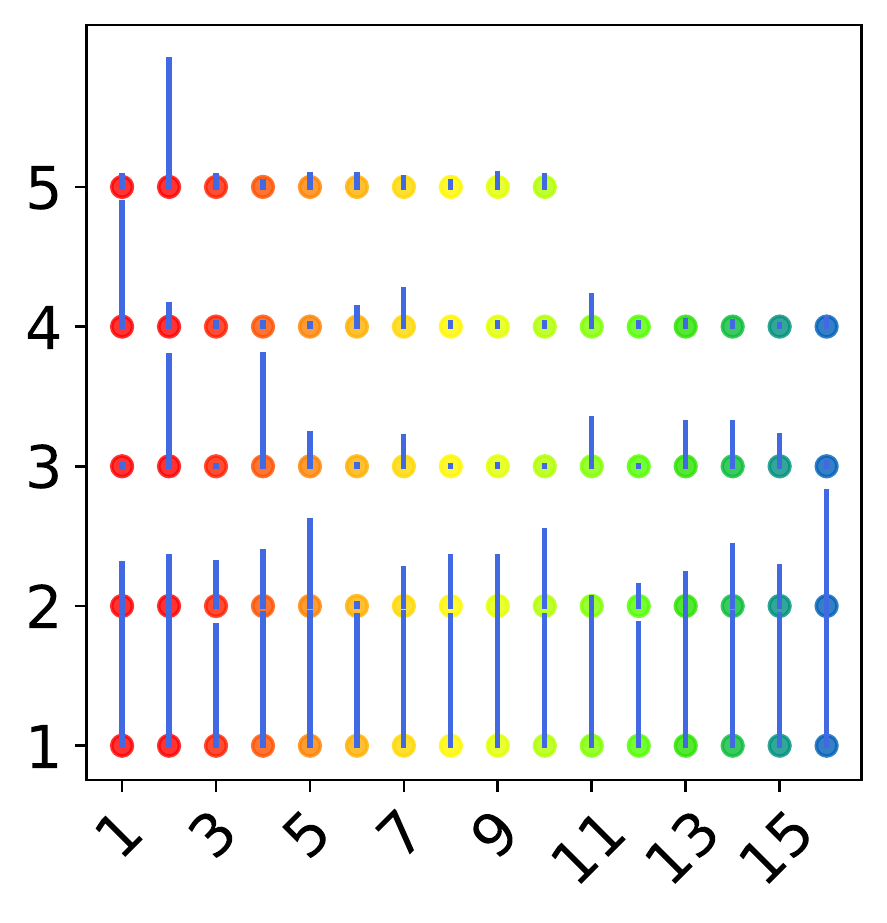} &
			\includegraphics[width=0.29\linewidth]{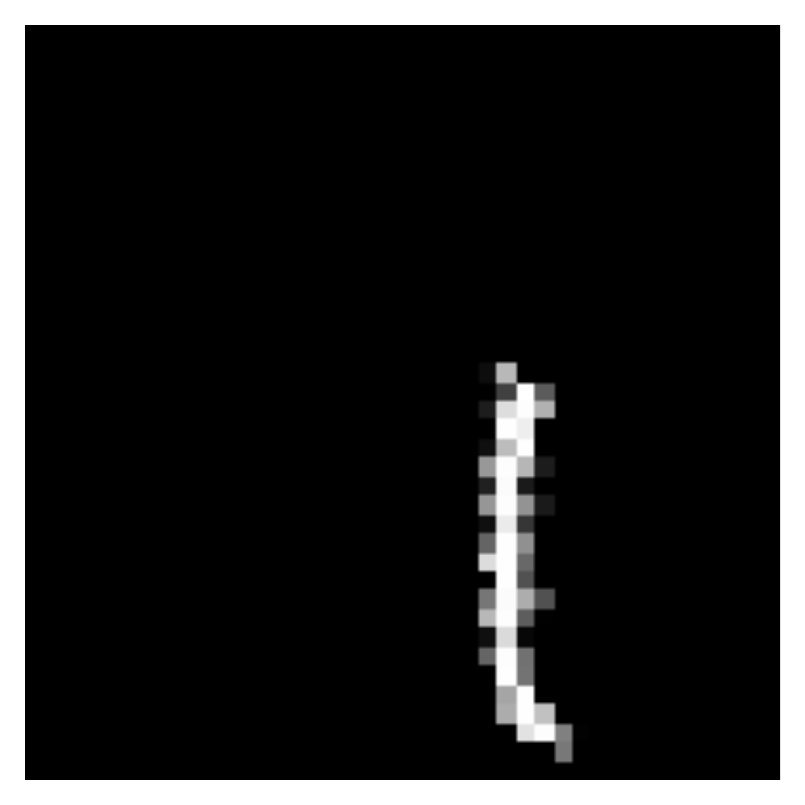} \\
			\includegraphics[width=0.29\linewidth]{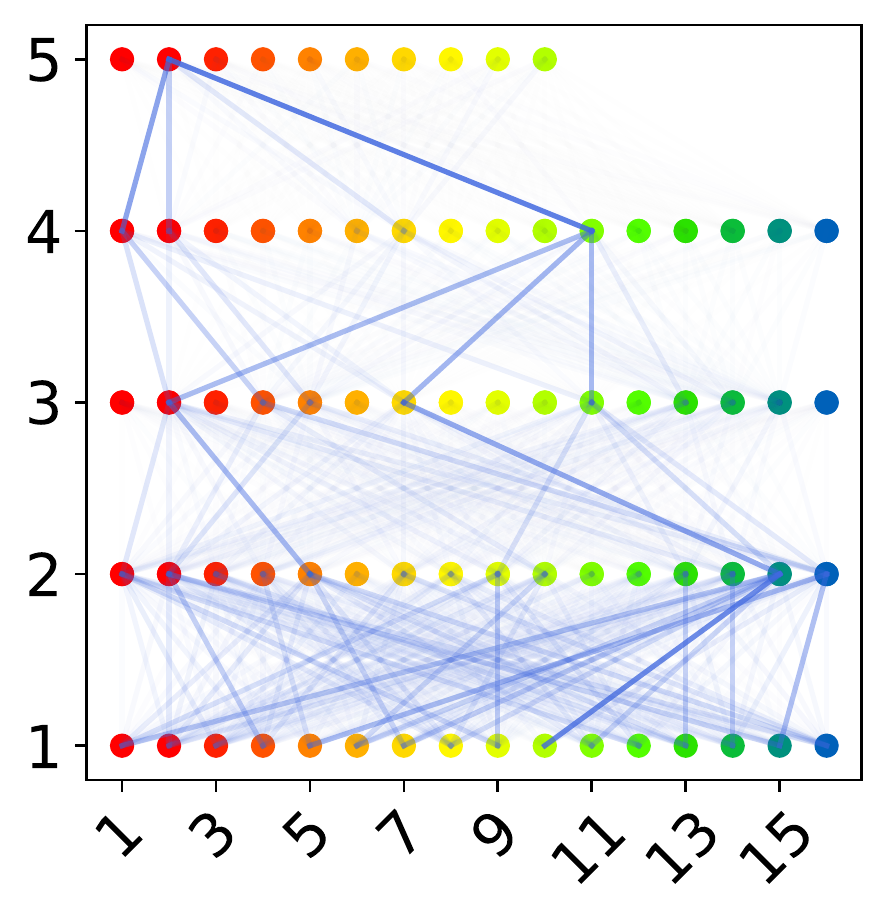} &
			\includegraphics[width=0.29\linewidth]{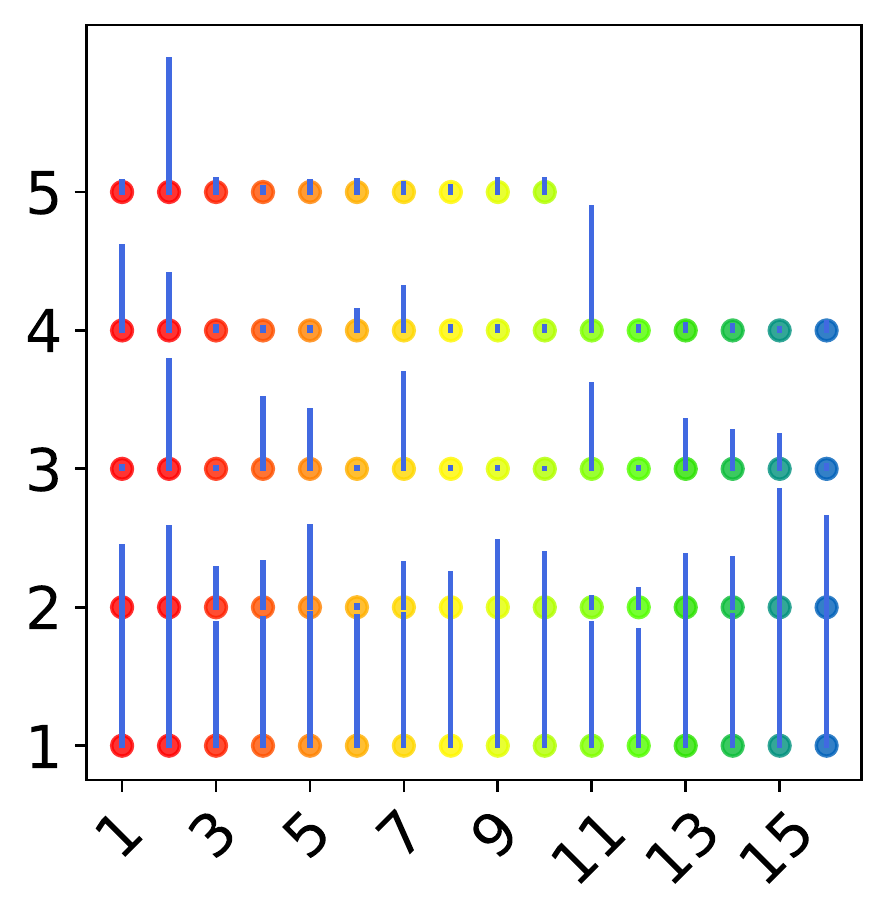} &
			\includegraphics[width=0.29\linewidth]{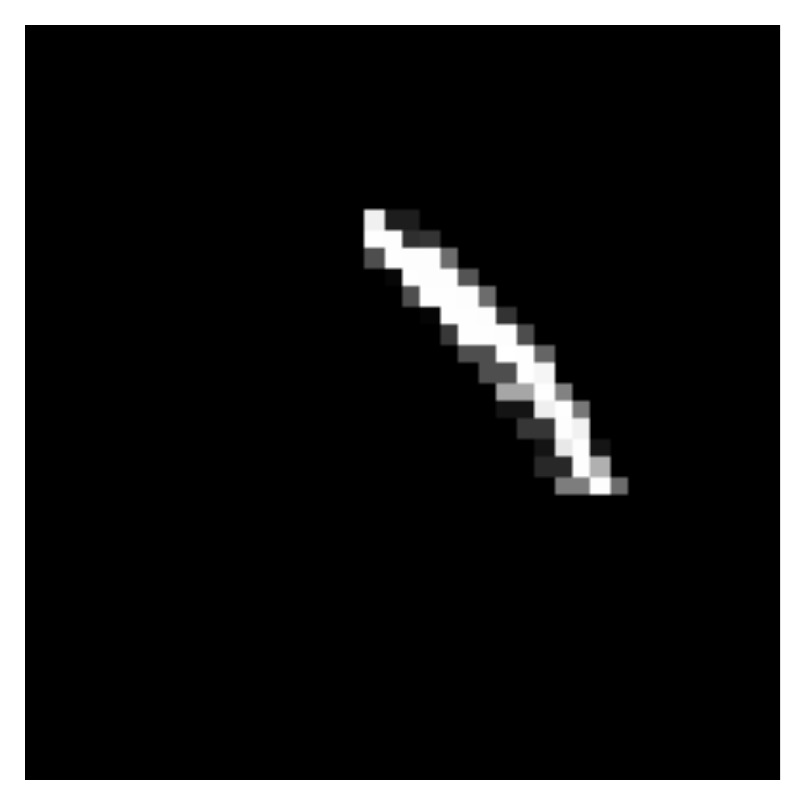} \\
		\end{tabular}
	\end{center}
	\caption{Similar input image, different parse tree.}
	\label{fig:affnist:main_model:sim_image_diff_parse_tree}
\end{figure}

\subsubsection{Results}
We observe that the capsule activation correlation decreases for increasingly stronger affine transformations, see Figure~\ref{fig:affnist:main_model:view_point_invariance}.
This observation holds for all intermediate capsule layers, and all tested affine transformations. See also Figure~\ref{fig:affnist:main_model:sim_image_diff_parse_tree} for a qualitative example showing two different parse-trees that are expected to be identical.
We conclude that the parse-tree is not invariant under affine transformations of the input image, violating Assumption~(1).
Furthermore, since already the activations of the PrimeCaps do not exhibit viewpoint-invariance, we believe that capsules need a different backbone that gives rise to better PrimeCaps.

\subsection{Capsule Activation}
\label{sec:starvation}
\begin{table*}[!ht]
	\centering
	\begin{tabular}{ccc|cc|cc}
		\toprule
		\multirow{2}{*}{Capsule Layer} &
		\multicolumn{2}{c}{Capsule Norms} &
		\multicolumn{2}{c}{Capsule Activation} &
		\multicolumn{2}{c}{Capsule Deaths} \\
		& Mean ($\cnm$) & Sum ($\cns$) & Rate ($\car$) & Sum ($\cas$) & Rate ($\cdr$) & Sum ($\cds$) \\
		\midrule
		1 & 0.95 $\pm$ 0.00 & 15.25 $\pm$ 0.06 & 1.00 $\pm$ 0.00 & 16.00 $\pm$ 0.00 & 0.00 $\pm$ 0.00 & 0.00 $\pm$ 0.00 \\
		2 & 0.32 $\pm$ 0.01 & 5.12 $\pm$ 0.18 & 0.70 $\pm$ 0.03 & 11.17 $\pm$ 0.48 & 0.07 $\pm$ 0.04 & 1.10 $\pm$ 0.70 \\
		3 & 0.18 $\pm$ 0.01 & 2.83 $\pm$ 0.08 & 0.35 $\pm$ 0.03 & 5.57 $\pm$ 0.42 & 0.56 $\pm$ 0.04 & 9.00 $\pm$ 0.63 \\
		4 & 0.12 $\pm$ 0.00 & 1.90 $\pm$ 0.07 & 0.22 $\pm$ 0.02 & 3.51 $\pm$ 0.27 & 0.63 $\pm$ 0.04 & 10.10 $\pm$ 0.70 \\
		5 & 0.15 $\pm$ 0.01 & 1.48 $\pm$ 0.05 & 0.30 $\pm$ 0.03 & 3.05 $\pm$ 0.00 & 0.00 $\pm$ 0.00 & 0.00 $\pm$ 0.00 \\
		\bottomrule
	\end{tabular}
	\caption{Capsule activation statistics for a CapsNet with five capsule layers, 16/10 capsules per layer, and a capsule dimension of eight. 
		We separately train and evaluate ten models on AffNIST the same way and report the mean and standard deviation.}
	\label{tab:affnist:main_model:activation}
\end{table*}
In this section, we analyze the capsule activation and thus the parse-tree nodes.
For a layer with $n$ capsules of dimension $d$, let $U \in \mathbb{R}^{k\times n \times d}$ hold the $n$ capsule responses of dimension $d$ for $k$ input images. We denote the entries of $U$ by the lower letter $u$ with corresponding lower indices.
We define the \textbf{capsule norm sum (cns)} as the sum of capsule norms in the respective layer averaged over all input images. For comparing layers with different numbers of capsules, we define the \textbf{capsule norm mean (cnm)}, which is the $\cns$ adjusted for the number of capsules that are present in the layer:
\[
\cns(U) = \frac{1}{k} \sum_{i=1}^{k}\sum_{j=1}^{n} \norm{u_{(i,j,:)}}_2, \qquad \cnm(U) = \frac{\cns(U)}{n}
\]

We say a capsule $j$ is active for input image $i$, if its norm exceeds a certain threshold $\varepsilon$, that is, 
\[
\mathds{1}_{\cactive}(u_{(i,j,:)}) = 
\begin{cases}
1 & \norm{u_{(i,j,:)}}_2 \geq \varepsilon \\
0 & \text{otherwise.}
\end{cases}
\]
Furthermore, we define the \textbf{sum of active capsules (cas)} as the mean sum of active capsules per layer and the \textbf{rate of activate capsules} as the $\cas$ adjusted for the number of capsules:
\[
\cas(U) = \frac{1}{k}\sum_{i=1}^{k}\sum_{j=1}^{n} \mathds{1}_{\cactive}(u_{(i,j,:)}), \quad \car(U) = \frac{\cas(U)}{n}
\]

We say that a capsule $j$  is \textbf{dead} if the mean $\mu$ and the standard deviation $\sigma$ of its norm over the $k$ input images are below certain thresholds, that is,
\[
\mathds{1}_{\dead}(u_{(:,j,:)}) = 
\begin{cases}
1 & \mu(u_{(:,j,:)}) \leq \varepsilon_1 \text{ and } \sigma(u_{(:,j,:)}) \leq \varepsilon_2 \\
0 & \text{otherwise.}
\end{cases}
\]
We compute the \textbf{sum of dead capsules (cds)} and the \textbf{rate of dead capsules (cdr)} as
\begin{align*}
\cds(U) = \sum_{j=1}^{n} \mathds{1}_{\dead}\left(u_{(:,j,:)}\right), \quad \cdr(U) = \frac{\cds(U)}{n}
\end{align*}

\subsubsection{Results}
Table~\ref{tab:affnist:main_model:activation} summarizes the capsule activation statistics for the AffNIST experiment.
As expected, the agreement of lower layer capsules, enforced by the RBA algorithm, results in a declining number of active capsules in the upper layers, as is witnessed by decreasing $\cas$ values.
As a result, the overall sum of norms per layer drops, as can be seen in the $\cns$ values.
Surprisingly, there is no sign of sparse activation within the PrimeCaps.
All PrimeCaps are consistently active, as seen from the $\car$ values.
This implies that PrimeCaps do not represent parts that are present in one image and not present in another, questioning the underlying assumption of distributed representation learning.
It is another indication that the backbone does not deliver the representations required for PrimeCaps.

Furthermore, we observe that the number of dead capsules $\cdr$ increases with the depth of the model. For instance, $63\%$ of the capsules on Layer 4 in the AffNIST experiment are dead.
Figure~\ref{fig:affnist:main_model:parse_tree_stats:dead_capsule} highlights the dead capsules.
In other experiments, this value increases up to $84\%$, see Table~\ref{tab:cifar10:model_d5:best} in the appendix.
This  has the following implications:
First, the depth of a CapsNet is limited as the number of dead capsules rises with the number of layers. 
Second, the parse-tree cannot carry separate semantic information for each class if the number of active capsules is less than the number of classes. 
Third, dead capsules limit the capacity of a CapsNet as their respective parameters are not in use.
This explains why baseline models trained with \textbf{uniform routing} perform better than models trained with RBA. Uniform routing, which uses all parameters, achieves better classification accuracies; see Tables~\ref{tab:affnist:grid_run:best_overall_models} and~\ref{tab:cifar10:grid_run:best_overall_models} in the appendix.
In uniform routing, all entries in the coupling coefficient matrix are set to the same fixed value. Our results stand in line with prior work~\citep{acml/PaikKK19,cvpr/GuT20,cvpr/GuT021} that also observed a negative impact of routing on model performance.

\subsubsection{Theoretical Analysis}
In order to theoretically explain the dynamics of the activation of the capsules during training, we analyze the gradient of the loss function. We have the following theorem:

\begin{theorem} Let $L_m$ be the margin loss function. 
The gradient of a single capsule $u^l_{(j,:)}$ of the upper-most layer $l$ evaluates to:
\begin{align*}
 \frac{\partial L_m}{\partial u^l_{(j,:)}} = &\left(-t_j \max(0, m^+ - \norm{u^l_{(j,:)}}_2)\right. \\
&+ \left.\lambda (1-t_j) \max(0, \norm{u^l_{(j,:)}}_2 - m^-)\right) \cdot \frac{2u^l_{(j,:)}}{\norm{u^l_{(j,:)}}_2}
\end{align*}
\end{theorem}
The theorem follows directly from the definition of the classification loss function, Equation~\eqref{eq:margin_loss}. \nocite{LaueMG18, LaueMG20} The gradient is independent of the magnitude $\norm{u^l_{(j,:)}}$ of the capsule activation. Hence, as long as the loss function is not zero, the gradient is large enough to force the capsules to either become active or inactive, depending on the label of the data point. Hence, all capsules on the upper-most layer will be active for the corresponding data points. This is in stark contrast to capsules that are not on the upper-most layer. Here, it can happen that a capsule becomes dead during training. We observe, that once a capsule is dead, it never becomes active again, resulting in a starvation of capsules. The following theorem asserts this behavior.

\begin{theorem} Let $L_m$ be the margin loss function. The gradient of a single capsule $u^l_{(i,:)}$, that does not belong to the upper-most layer evaluates to:
\[ \frac{\partial L_m}{\partial u^l_{(i,:)}} = \sum_j \frac{\partial L_m}{\hat u^{l+1}_{(i,j,:)}}\cdot W^l_{(j,i,:,:)}
\]
The gradients of the corresponding weight matrices $W^l_{(j,i,:,:)}$ evaluate to:
\[ \frac{\partial L_m}{\partial W^l_{(j,i,:,:)}} = \frac{\partial L_m}{\partial \hat u^{l+1}_{(i,j,:)}}\cdot u^l_{(i,:)}
\]
\label{thm:vanishing}
\end{theorem}
The theorem follows from Equation~\eqref{eq:votes}. It states that the gradient of the weight matrix scales with the activation of the corresponding capsule, and the gradient of the capsule scales with the magnitude of the weight matrix. Hence, once both are small, they will not change sufficiently. In the limit, i.e., of magnitude zero, they will never change. In other terms, once a capsule becomes dead, it never becomes active again. 
Figure~\ref{fig:affnist:main_model:training:caps_norms_max} in the appendix clearly show this behavior for the gradient of the capsule activation and Figures~\ref{fig:affnist:main_model:training:grad_norms_0}-\ref{fig:affnist:main_model:training:grad_norms_3} in the appendix show this behavior for the gradient of the weight matrices. Dead capsules do not participate in the routing and are not part of any parse-tree. Also, note that the supervised loss forces the upper-most layers to be active. However, capsules can become dead on the intermediate layers where no supervised loss is directly present.

\section{Comparing RBA with Self-Attention}
We compare routing-by-agreement with the multi-head self-attention (MHSA) mechanism used in transformers~\citep{nips/VaswaniSPUJGKP17} and more task-related vision transformers~\citep{iclr/DosovitskiyB0WZ21}. 
Like RBA, the MHSA mechanism operates on vectors and uses the softmax function to compute the normalized attention matrix, similar to the coupling coefficient matrix in RBA.
However, contrary to RBA, which computes the softmax row-wise, MHSA enforces the softmax column-wise and, as a result, does not suffer from the previously discussed vanishing gradient problem.
However, MHSA is not considered routing and does not intend to implement a parse-tree.
Considering space and time complexity, we observe that RBA is extremely expensive compared to MHSA; see Table~\ref{tab:rba_mha_complexity}.
This fact may contribute to the substantial interest in models relying on MHSA rather than RBA.
\begin{table}[!ht]
	\centering
	\begin{tabular}{l|ll}
		& Space & Time \\
		\midrule
		MHSA & $O(d^2)$  & $O(n^2 \cdot d + n \cdot d^2)$ \\
		RBA & $O(n^2 \cdot d^2)$ & $O(n^2 \cdot d^2)$ \\
	\end{tabular}
	\caption{Comparing space and time complexity of routing-by-agreement and multi-head self-attention for a routing layer with $n$ input and output vectors of dimension~$d$.}
	\label{tab:rba_mha_complexity}
\end{table}

\section{Broader Impact and Limitations}
In this work, we focus on the original CapsNet with RBA routing since it is the predominant implementation of the capsule idea.
An exhaustive investigation, including all capsule variants and follow-up models, is difficult because the absence of a formal definition of capsules makes the topic hard to cover.
Different approaches from the vast literature are technically diverse.
As a result,  whether a follow-up work implements the concept of capsules is not easy to judge.
However, our claims are general enough to cover many implementations.
The softmax-based routing approach is part of many capsule implementations, see for instance \cite{spl/XiangZTZX18, aaai/ZhouJSSC19, nature/Mazzia21}, and we expect that they face similar issues.

\section{Conclusion}
The core concept of capsules is the part-whole hierarchy of an image represented by a parse-tree.
While this concept has appealing properties like robustness under affine transformations, interpretability, and parameter efficiency, so far, implementations of the capsule concept have not taken over yet. Instead, some of their properties were questioned in recent work. Here, we have shown that the core idea of a parse-tree does not emerge in the CapsNet implementation. Furthermore, starvation of capsules caused by a vanishing gradient limits their capacity and depth.
Our observations explain recently reported shortcomings, especially that \mbox{CapsNets} do not scale beyond small datasets. Hence, the \mbox{CapsNet} is not a sufficient implementation of the capsule idea.

\section{Acknowledgments}
This work was supported by the German Science Foundation (DFG) grant (GI-711/5-1) within the priority program (SPP 1736) Algorithms for Big Data and by the Carl Zeiss Foundation within the project "Interactive Inference".

\bibliography{references}

\onecolumn
\newpage
\appendix
\section*{\LARGE Why Capsule Neural Networks Do Not Scale:\\Challenging  the Dynamic Parse-Tree Assumption\\
\vspace{0.4cm}
\Large Appendix}
\vspace{1cm}

This appendix provides additional materials which did not fit into the main paper. It is organized as follows:
\vspace{\baselineskip}

\begin{center}
\begin{tabular}{ll}
A & Detailed descriptions of models, training procedures, and data sets. \\[3pt]
B & Additional results from experiments testing viewpoint invariance\\
  & (Section~\ref{sec:viewpoint} in the main paper) for different models.\\[3pt]
C & Additional results for different models on the AffNIST data set.\\[3pt]
D & Exhaustive experiments concerning different aspects on the AffNIST dataset.\\[3pt]
E & Detailed evaluation of a single CIFAR10 model.\\[3pt]
F & Exhaustive experiments concerning different aspects on the CIFAR10 data set.
\end{tabular}
\end{center}

\vspace{\baselineskip}

\section{Model Architectures and Training Procedures}\label{app:model_architectures}

\subsection{CapsNets for AffNIST}
\label{app:capsnet_for_affnist}

The AffNIST dataset~\citep{nips/SabourFH17} is derived from the classic MNIST data set~\cite{dataset/lecun10} as follows:
First, all MNIST images are zero-padded to dimensions 40$\times$40 and affinely transformed by random rotations up to $20$ degrees, random shearings up to $40$ degrees, random scaling from $0.8$ to $1.2$, and random translations up to eight pixels in each direction.
We use this dataset to assess a model's robustness to affine transformations of the input data.
For this, we train the model on the original MNIST train set, where the images are randomly placed on a 40$\times$40 empty background.
For a fair comparison of different models, training is stopped once a model reaches a target accuracy of $99.20\%$ on the original MNIST validation set.
The trained models are then evaluated on the AffNIST validation set.
The difference between the MNIST validation set accuracy and the AffNIST validation set accuracy measures the model's robustness to affine transformations.

The general model architecture for all models trained on AffNIST largely follows the description in Section~\ref{sec:capsule_neural_network} of the main paper.
We borrow the backbone proposed by~\citet{nature/Mazzia21}, which utilizes four standard convolutional layers Conv(32,7,1), Conv(64,3,1), Conv(64,3,2), and Conv($n^{1} \cdot d^{1}$,3,2), followed by a fifth depthwise convolutional layer Conv($n^{1} \cdot d^{1}$,7,1,$n^{1} \cdot d^{1}$). The reconstruction network consists of three fully connected layers FC(512), FC(1024), and FC(40$\cdot$40), all using the ReLU activation function.
The hyper-parameters for the loss all are set to $m^+ = 0.9$, $m^- = 1 - m^+ =0.1$, $\lambda = 0.5$ and $\alpha = 0.392$.
We found the best number of iterations for the routing algorithm to be ten.
We used the Adam optimizer with an initial learning rate of $10^{-3}$, an exponential learning rate decay of $0.97$, and a weight decay regularizer with a value of $10^{-6}$.
We used a batch size of $512$.

\subsection{CapsNets for CIFAR10}

The CapsNet architecture for the CIFAR10 classification task largely follows the architecture used for the AffNIST benchmark with slight modifications in the backbone to adjust for the smaller input image size. 
The backbone utilizes four standard convolutional layers Conv(32,7,1), Conv(64,3,1), Conv(128,3,2), and Conv($n^{1} \cdot d^{1}$,3,2), followed by a fifth deepthwise convolutional layer Conv($n^{1} \cdot d^{1}$,5,1,$n^{1} \cdot d^{1}$).

We trained all CIFAR10 models for 100 epochs similarly to the AffNIST models, but with an initial learning rate of  $10^{-4}$. For computing the performance metrics, we selected the best model regarding validation set accuracy.

\subsection{The Original CapsNet for MNIST Digit Classification}
\label{app:capsnet_for_mnist}
%
%
%
%
Here we describe the original CapsNet architecture for the MNIST classification task by~\citet{nips/SabourFH17}.
The backbone function consists of two consecutive convolutional layers, Conv(256,9,1) and Conv(256,9,2), followed by a single routing layer as generally described in Section~\ref{sec:capsule_neural_network} of the main paper.
There are 1152 PrimeCaps of dimension eight and ten output capsules of dimension 16.
The reconstruction network consists of three fully connected layers, namely FC(512), FC(1024), and FC(28$\cdot$28), all using the ReLU activation function.
The hyper-parameters for the loss are set to $m^+ = 0.9$, $m^- = 1 - m^+ =0.1$, $\lambda = 0.5$ and $\alpha = 0.0005$. The number of iterations for the routing algorithm is set to $3$.

\subsection{Notes on Backbone Architectures and PrimeCaps}

The role of the backbone is to extract meaningful features from the input images. The features constitute the PrimeCaps.
Therefore, the backbone also controls the number and dimension of the PrimeCaps.
The concept of capsules requires that a capsule is activated if its related conceptual entity is present within the input image~\cite{nips/SabourFH17}.
For this reason, each capsule must have a full receptive field over the input image.

The original CapsNet backbone, used on the MNIST classification task, consists of two consecutive convolutional layers.
However, for larger input dimensions, e.g., AffNIST (40$\times$40) or ImageNet (224$\times$224), using only two convolutional layers is not practical since this would either result in a vast number of PrimeCaps or large capsule dimensions that make the models computationally infeasible.
Table~\ref{tab:rba:primecaps_dimensions} lists the number of parameters in the routing layer for the original CapsNet model for different input image dimensions, depending on the dimension of the output capsules. As can be seen, the number of parameters increases quickly with the input dimension, and as a result, computational and space requirements also rise quickly; see Table~\ref{tab:rba_mha_complexity} in the main paper.

Therefore, we borrow the backbone proposed by~\citet{nature/Mazzia21} because it allows us to control both the number and dimension of the PrimeCaps, and thus scales well to larger input dimensions since the number of parameters only grows moderately with the input dimension.

However, we also evaluated the original CapsNet backbone in our investigations of viewpoint invariance. See Appendix~\ref{app:viewpoint_invariance} for more details.

\begin{table}[ht]
	\centering
	\begin{tabular}{ccc|cccc}
		\toprule
		\multicolumn{3}{c}{Dataset} &
		\multicolumn{4}{c}{Dimension of a DigitCaps} \\
		Name & Classes & Input Dimensions & 16  & 32  & 64  & 128 \\
		\midrule
		MNIST   & 10  &  (28x28x1) &   1.5 & 3.0 & 5.9 & 11.8 \\
		AFFNIST & 10  &  (40x40x1) &    5.9 & 11.8 & 23.6 & 47.2 \\
		CIFAR10 & 10  &  (32x32x3) &    2.6 & 5.3 & 10.5 & 21.0 \\
		TinyImageNet & 200  &  (64x64x3) &  472.3 & 944.2 & 1887.9 & 3775.3 \\
		ImageNet 224 & 1000  &  (224x224x3) & 44324.0 & 88626.3 & 177231.0 & 354440.3 \\
		\bottomrule
	\end{tabular}
	\caption{The number of parameters in the routing layer for the original CapsNet architecture for different input image sizes and output capsule dimensions. The number of output capsules is set to match the number of data set classes. The number of parameters is given in \textbf{millions}.}
	\label{tab:rba:primecaps_dimensions}
\end{table}

\clearpage
\section{Viewpoint Invariance: Additional Results}\label{app:viewpoint_invariance}

\begin{figure*}[!ht]%
	\centering
	\subfloat[rotation]
	{\includegraphics[width=0.19\linewidth]{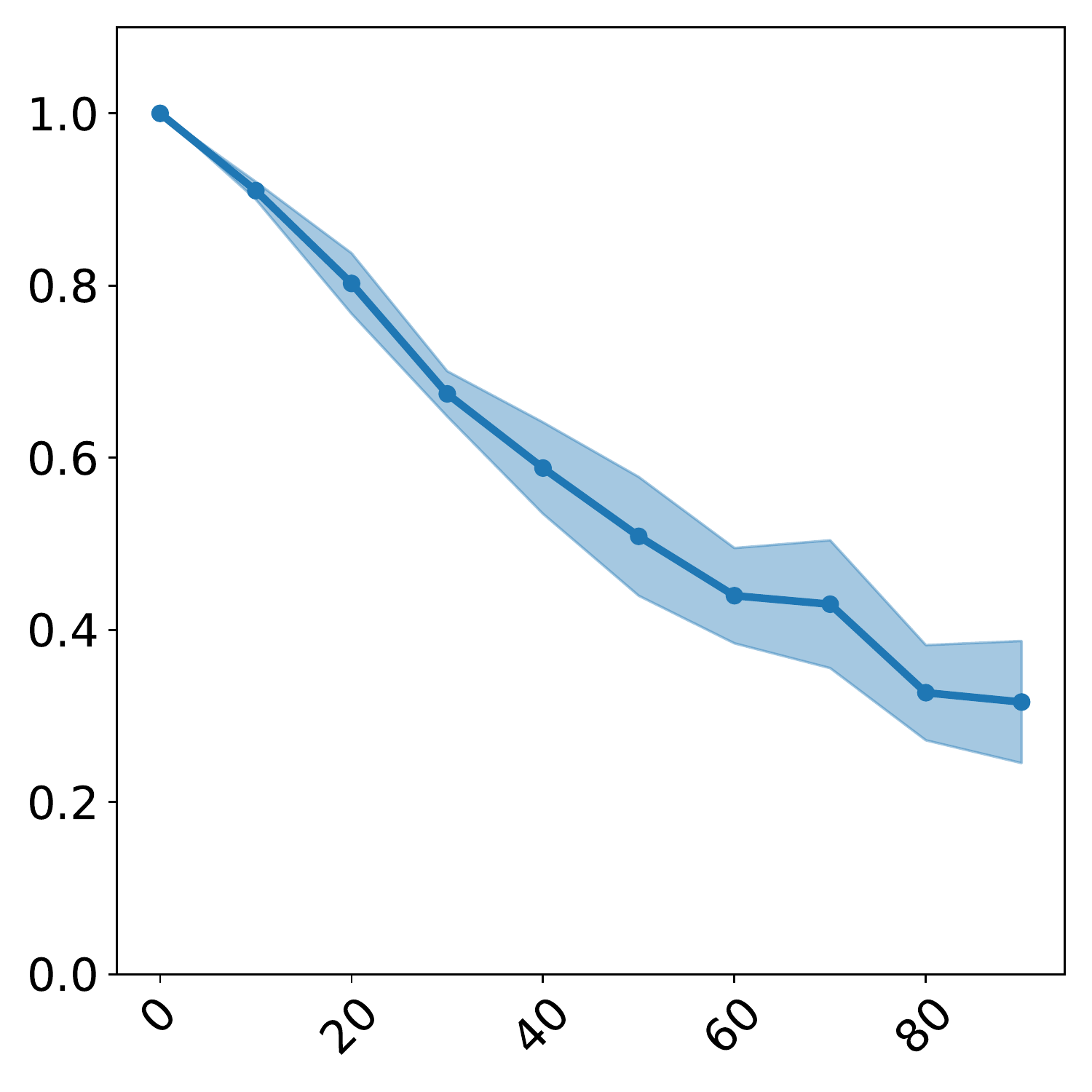}}
	\subfloat[shearing]
	{\includegraphics[width=0.19\linewidth]{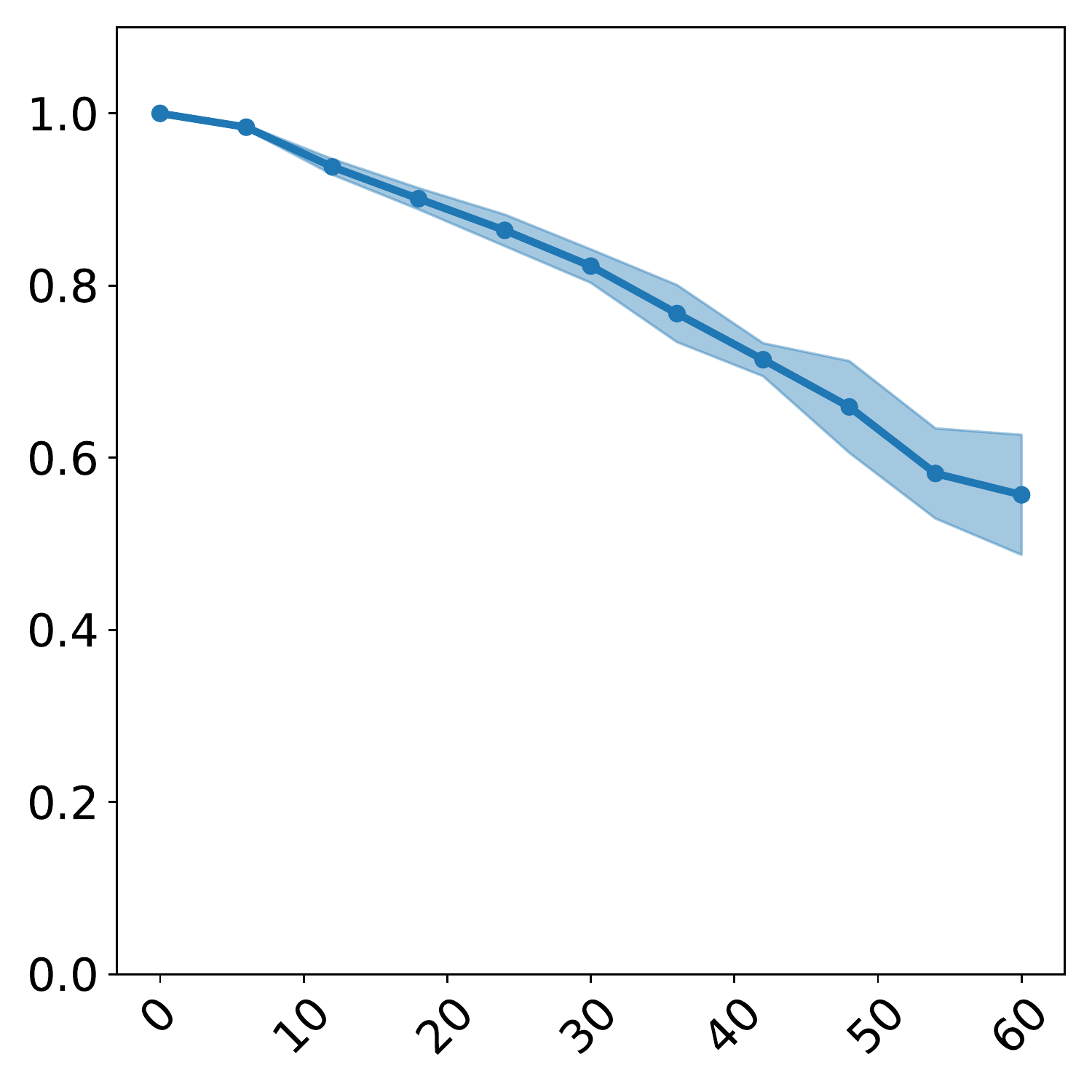}}
	\subfloat[scaling]
	{\includegraphics[width=0.19\linewidth]{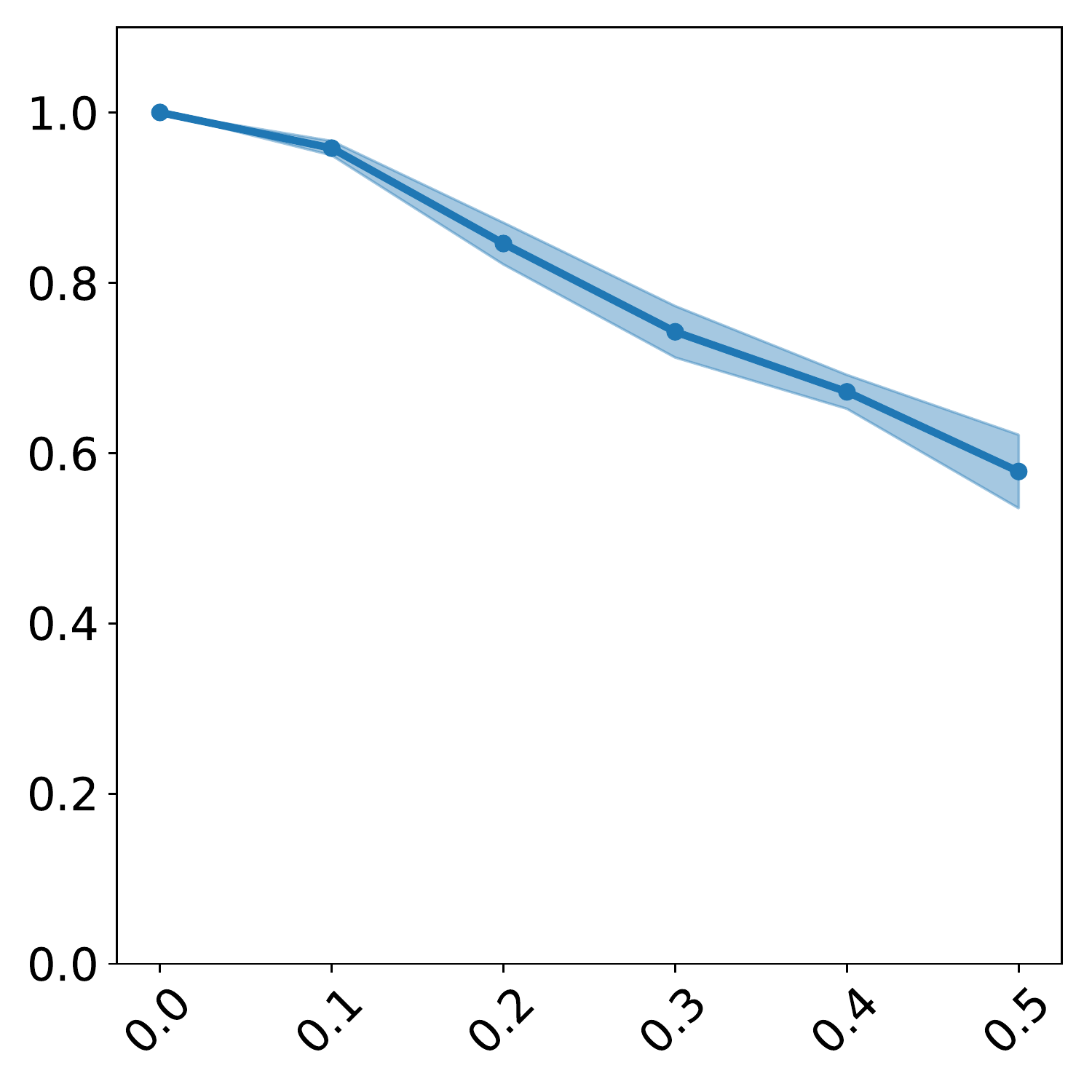}}
	\subfloat[horizontal translation]
	{\includegraphics[width=0.19\linewidth]{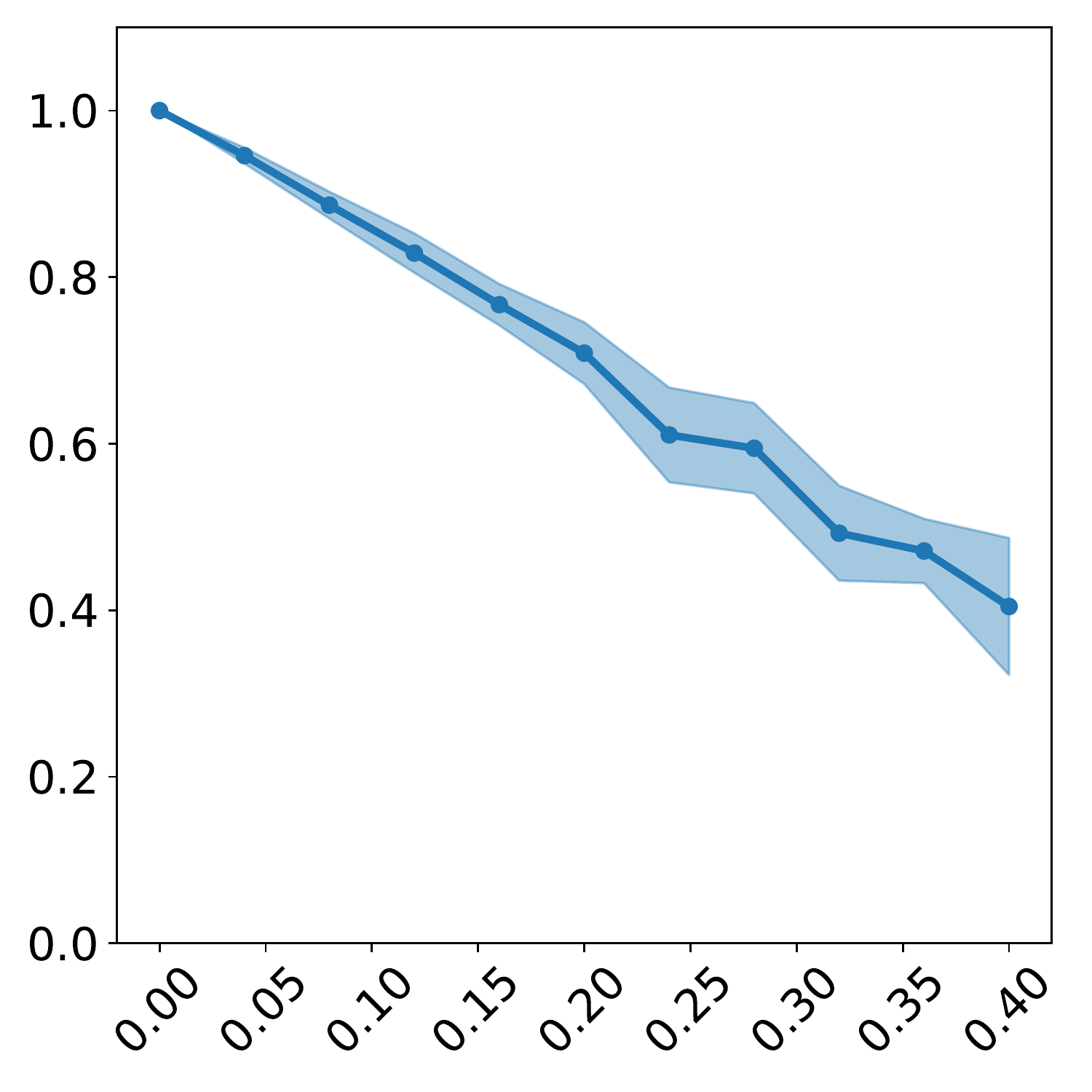}}
	\subfloat[vertical translation]
	{\includegraphics[width=0.19\linewidth]{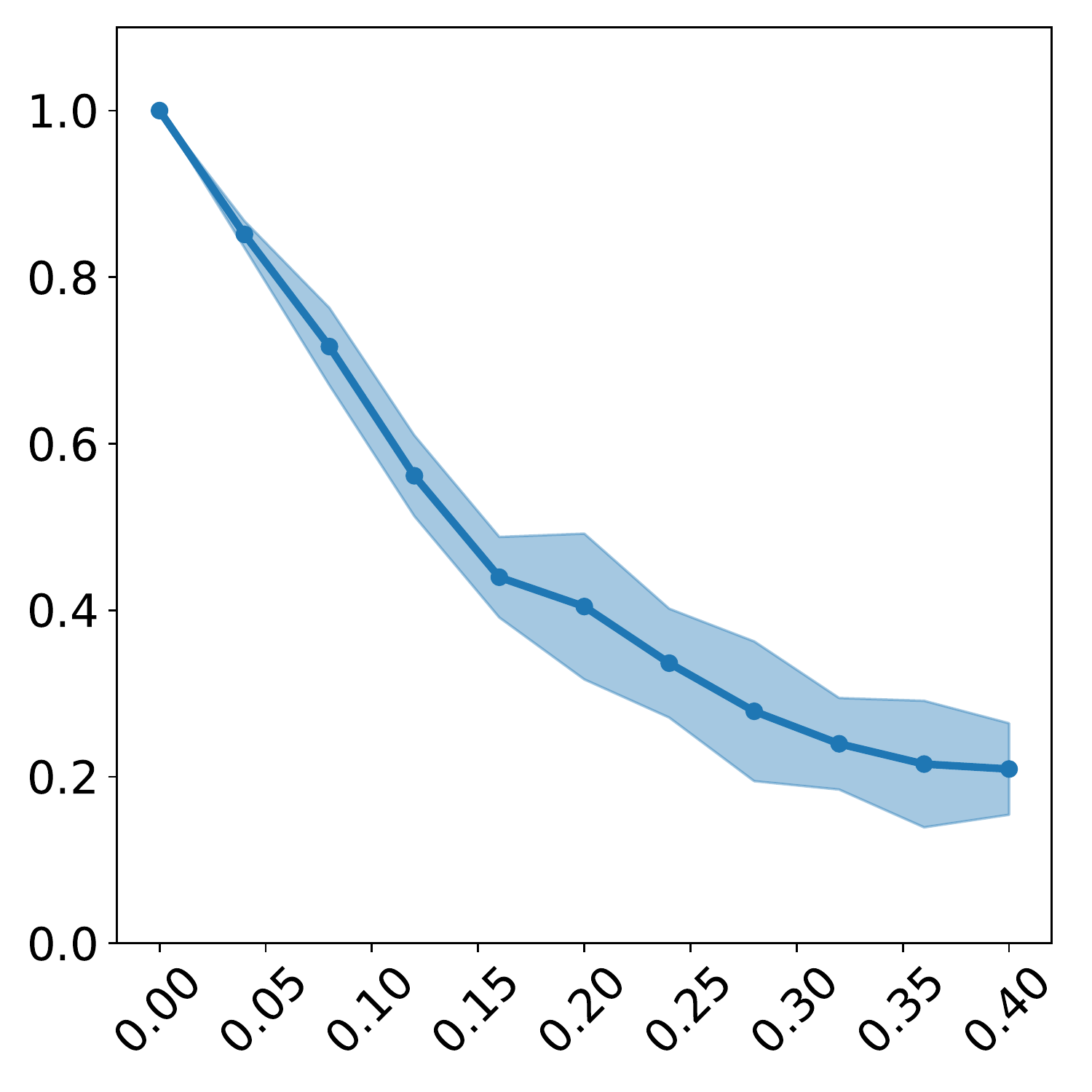}}
	\caption{Capsule activation correlation for the original CapsNet (Appendix \ref{app:capsnet_for_mnist}). The correlation decreases with stronger affine transformation. Left to right: rotation, shearing, scaling, translation in x-direction, translation in y-direction. Shown are the means and the standard deviations from PrimeCaps activation over ten trained models.}
	\label{fig:affnist:corss_corr:hinton_orig_model}
\end{figure*}

\begin{figure*}[!ht]%
	\centering
	\subfloat[rotation]
	{\includegraphics[width=0.19\linewidth]{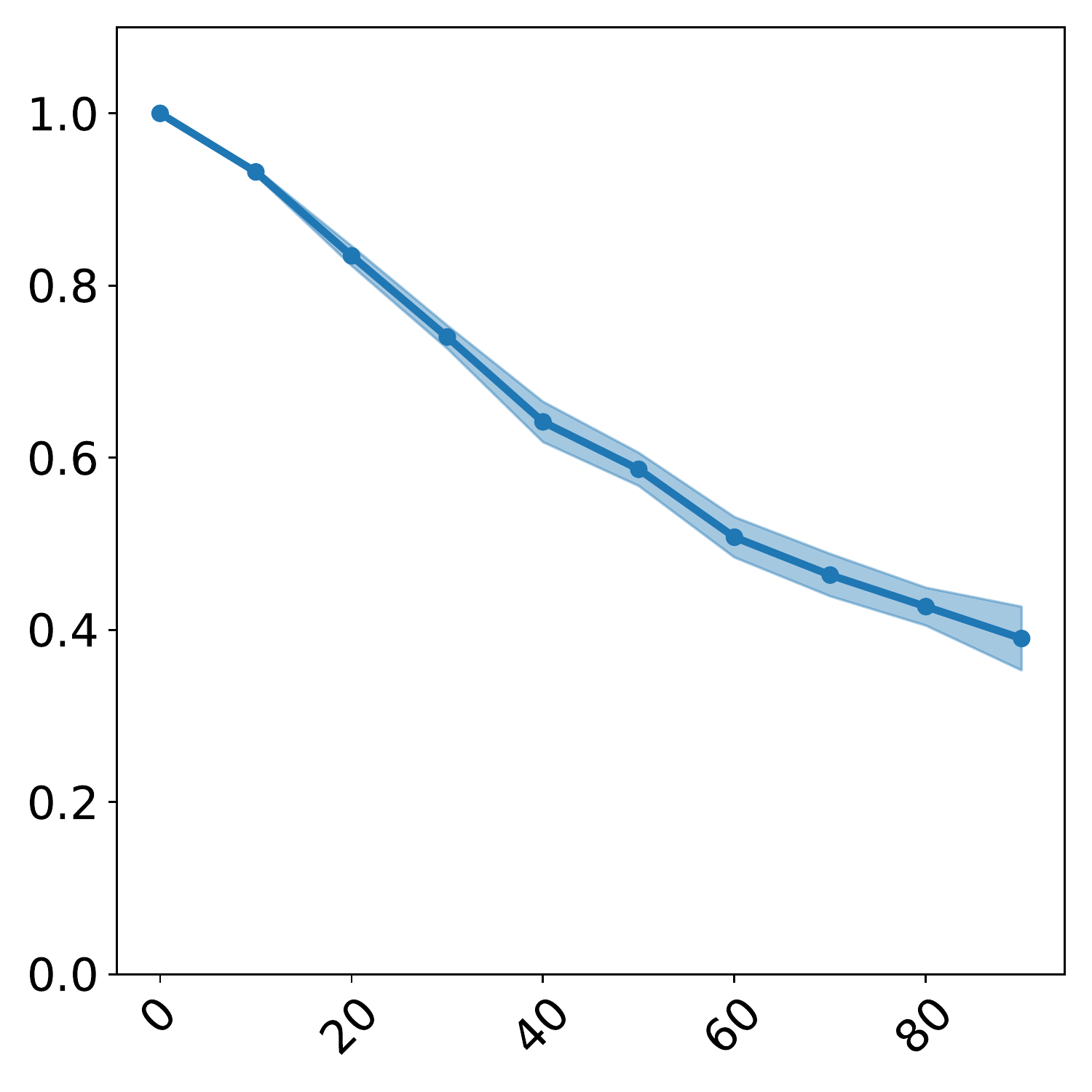}}
	\subfloat[shearing]
	{\includegraphics[width=0.19\linewidth]{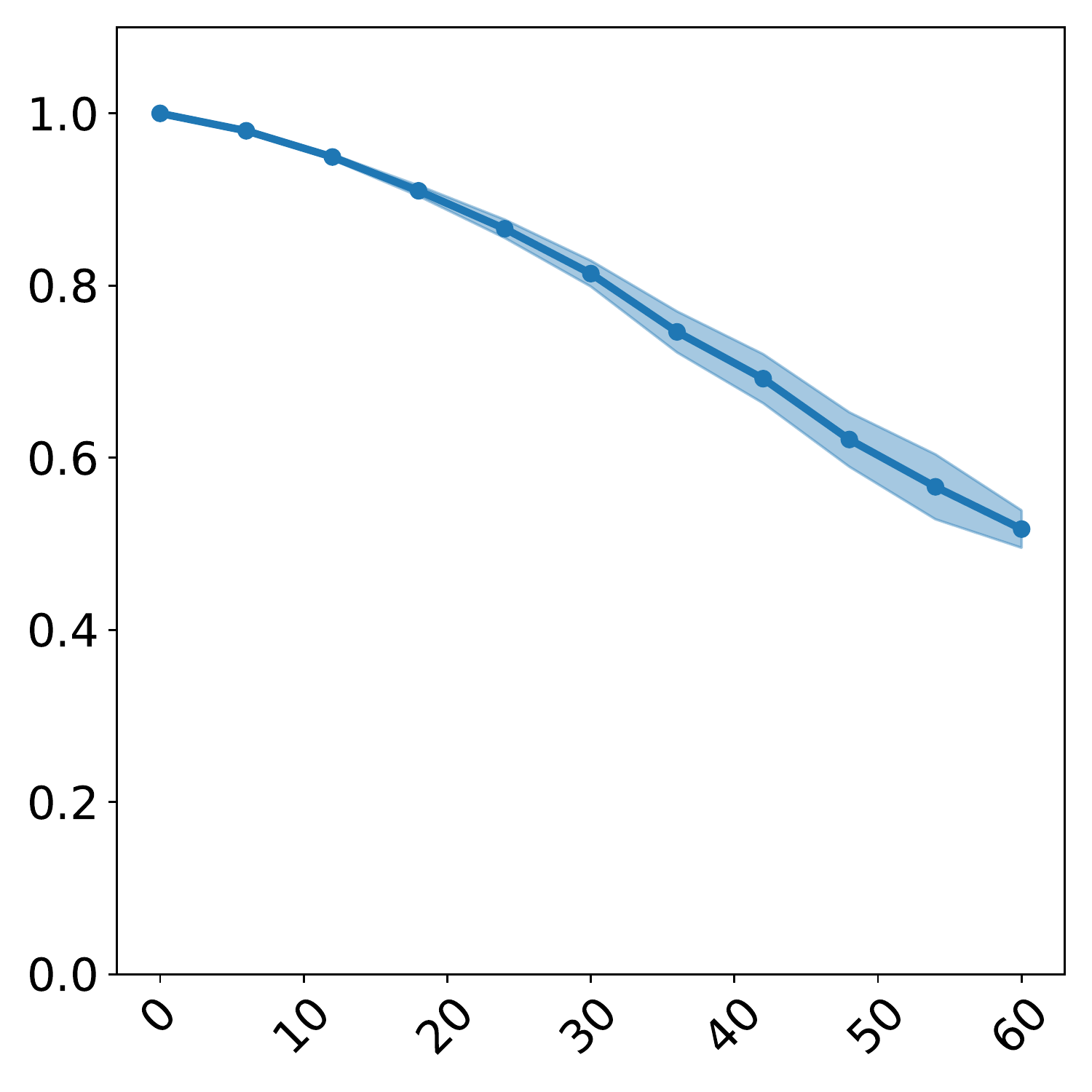}}
	\subfloat[scaling]
	{\includegraphics[width=0.19\linewidth]{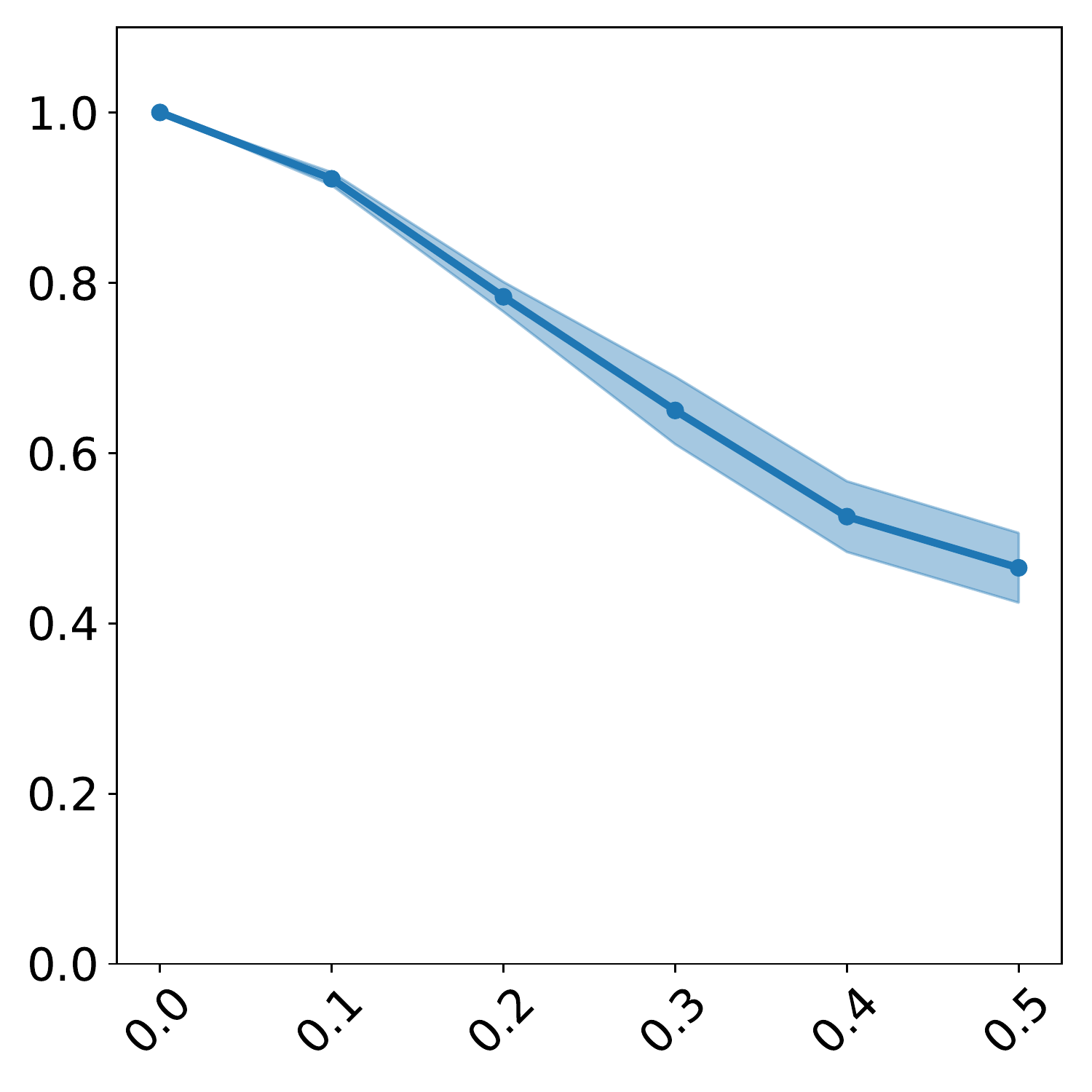}}
	\subfloat[horizontal translation]
	{\includegraphics[width=0.19\linewidth]{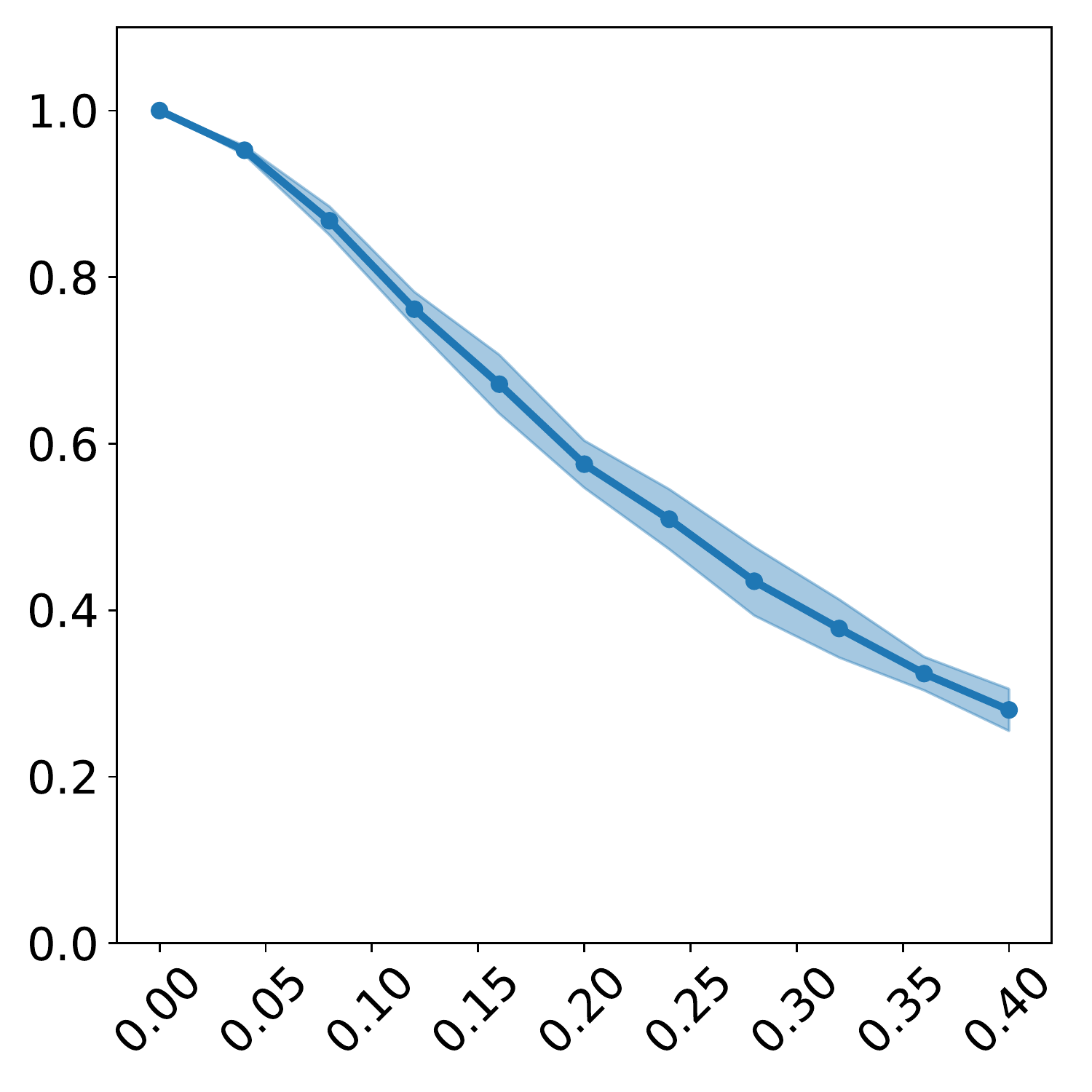}}
	\subfloat[vertical translation]
	{\includegraphics[width=0.19\linewidth]{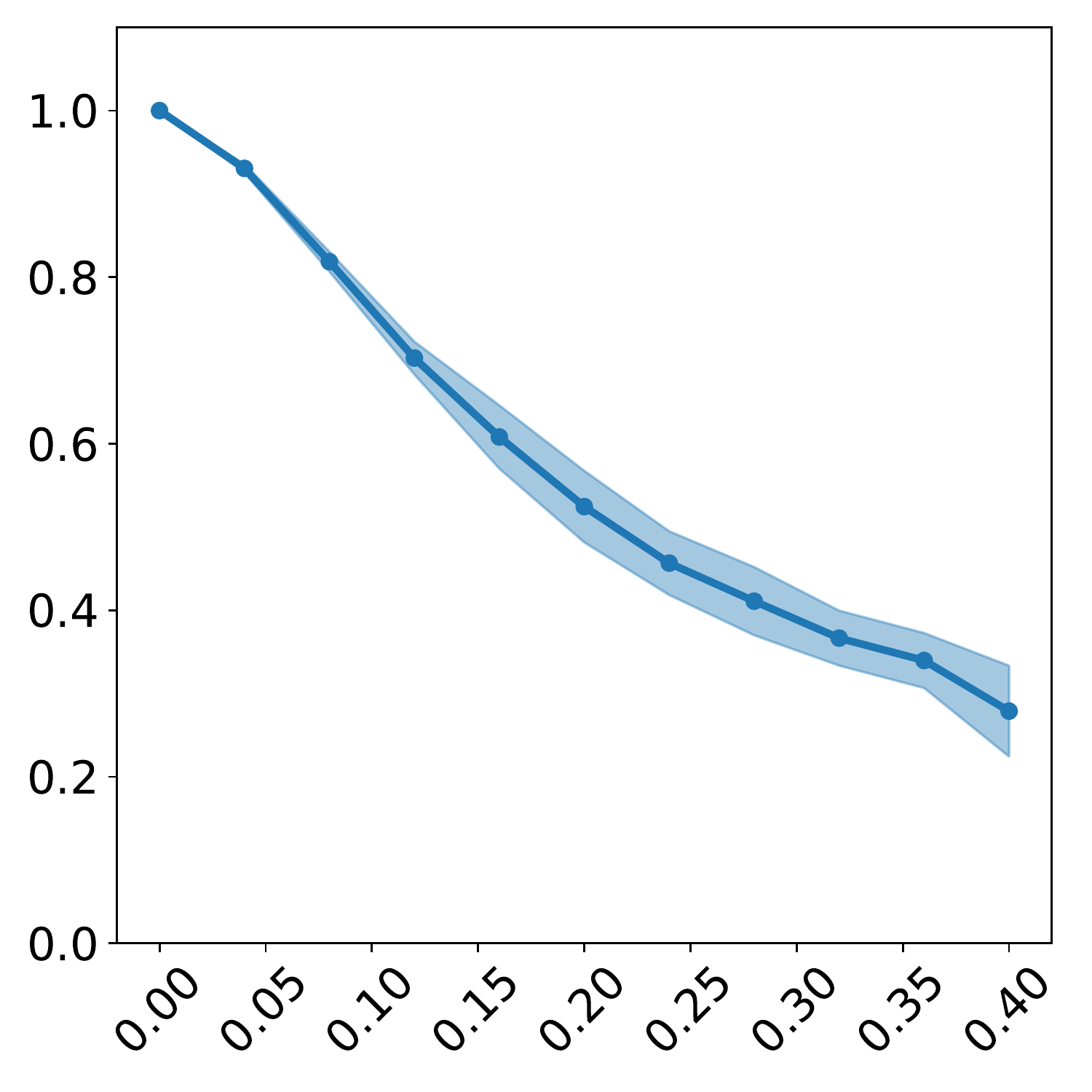}}
	\caption{Capsule activation correlation for a CapsNet trained on AffNIST with a single routing layer. The correlation decreases with stronger affine transformation. Left to right: rotation, shearing, scaling, translation in x-direction, translation in y-direction. Shown are the means and the standard deviations from PrimeCaps activation over ten trained models.}
	\label{fig:affnist:corss_corr:custom_16_8_1_model}
\end{figure*}

\begin{figure*}[!ht]%
	\centering
	\subfloat[rotation]
	{\includegraphics[width=0.19\linewidth]{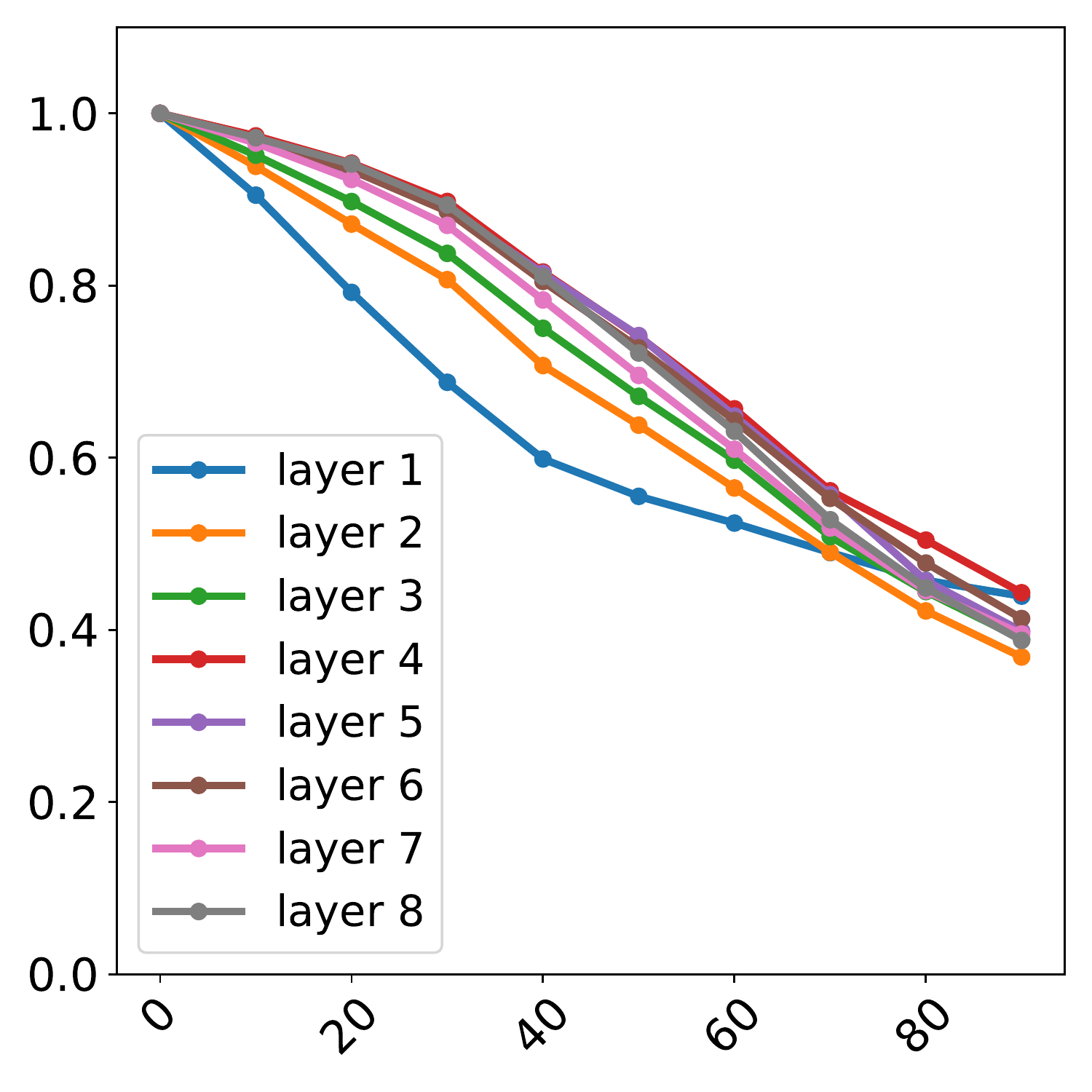}}
	\subfloat[shearing]
	{\includegraphics[width=0.19\linewidth]{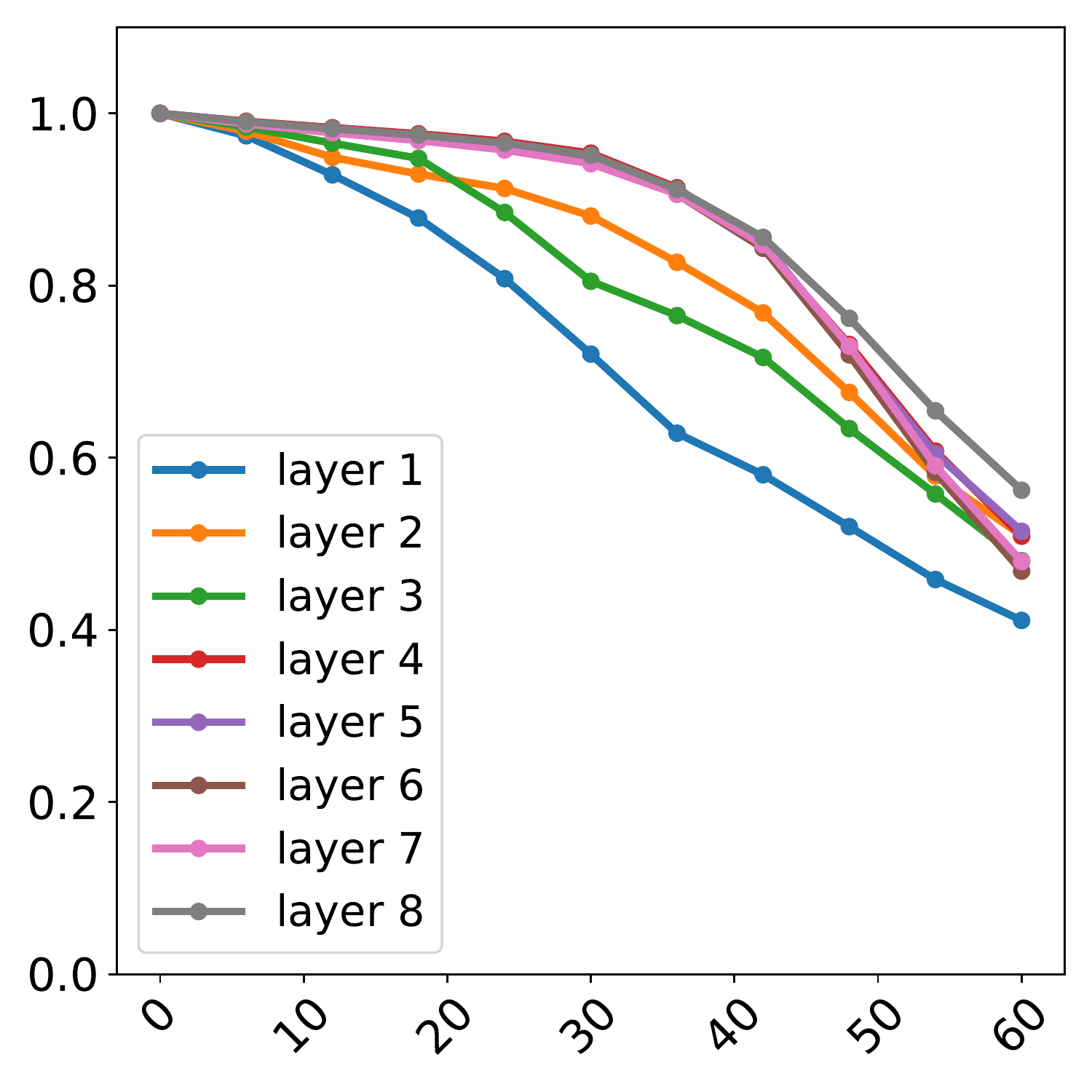}}
	\subfloat[scaling]
	{\includegraphics[width=0.19\linewidth]{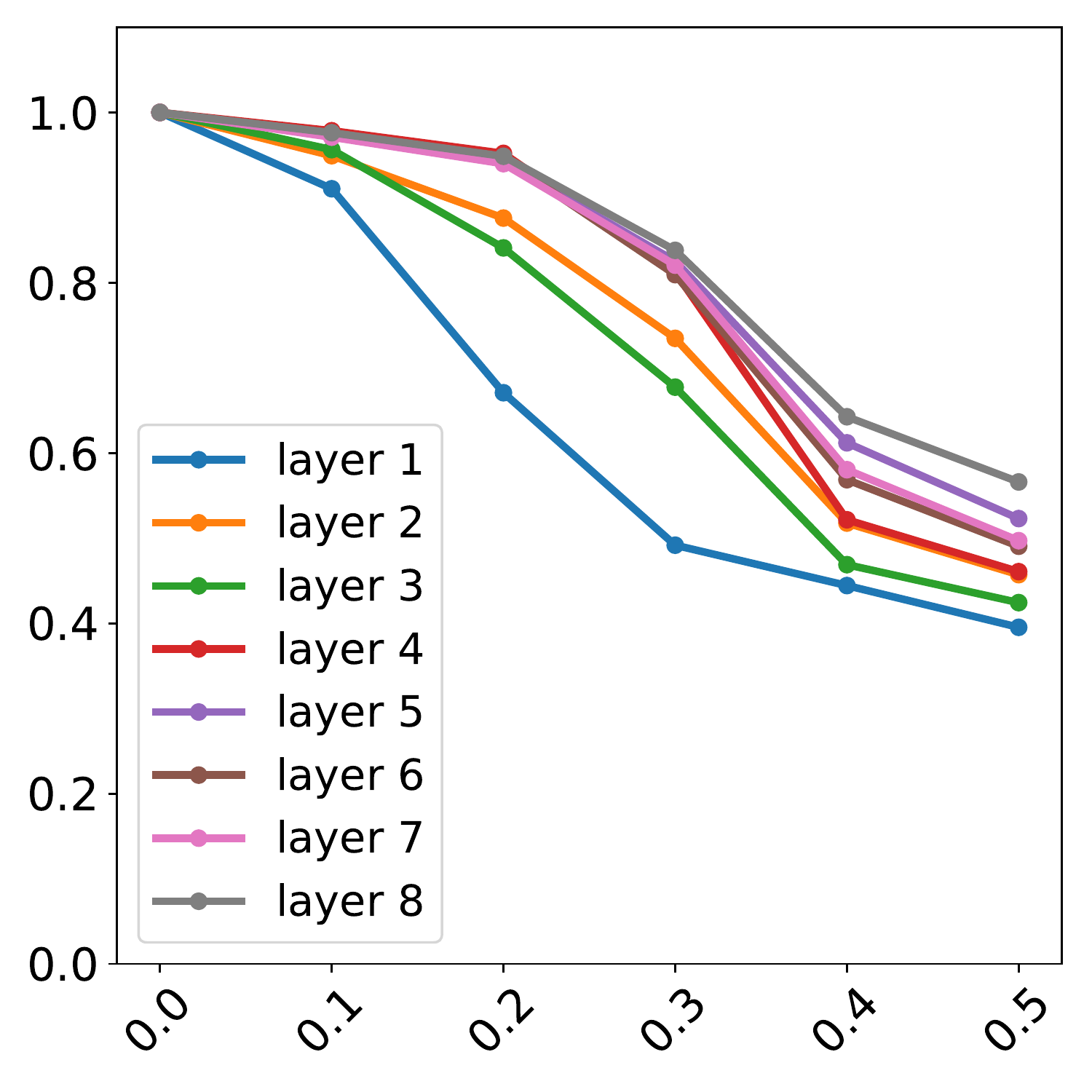}}
	\subfloat[horizontal translation]
	{\includegraphics[width=0.19\linewidth]{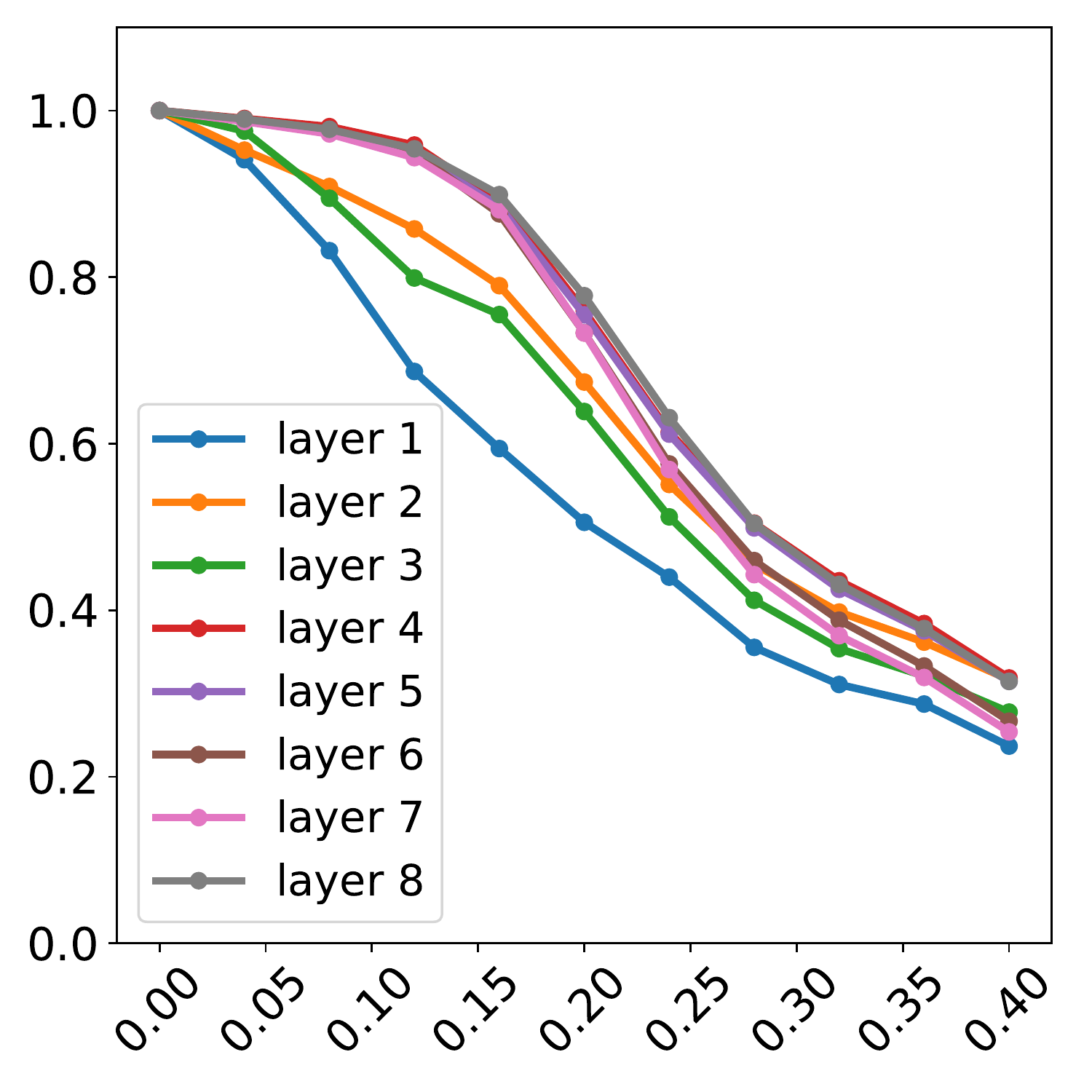}}
	\subfloat[vertical translation]
	{\includegraphics[width=0.19\linewidth]{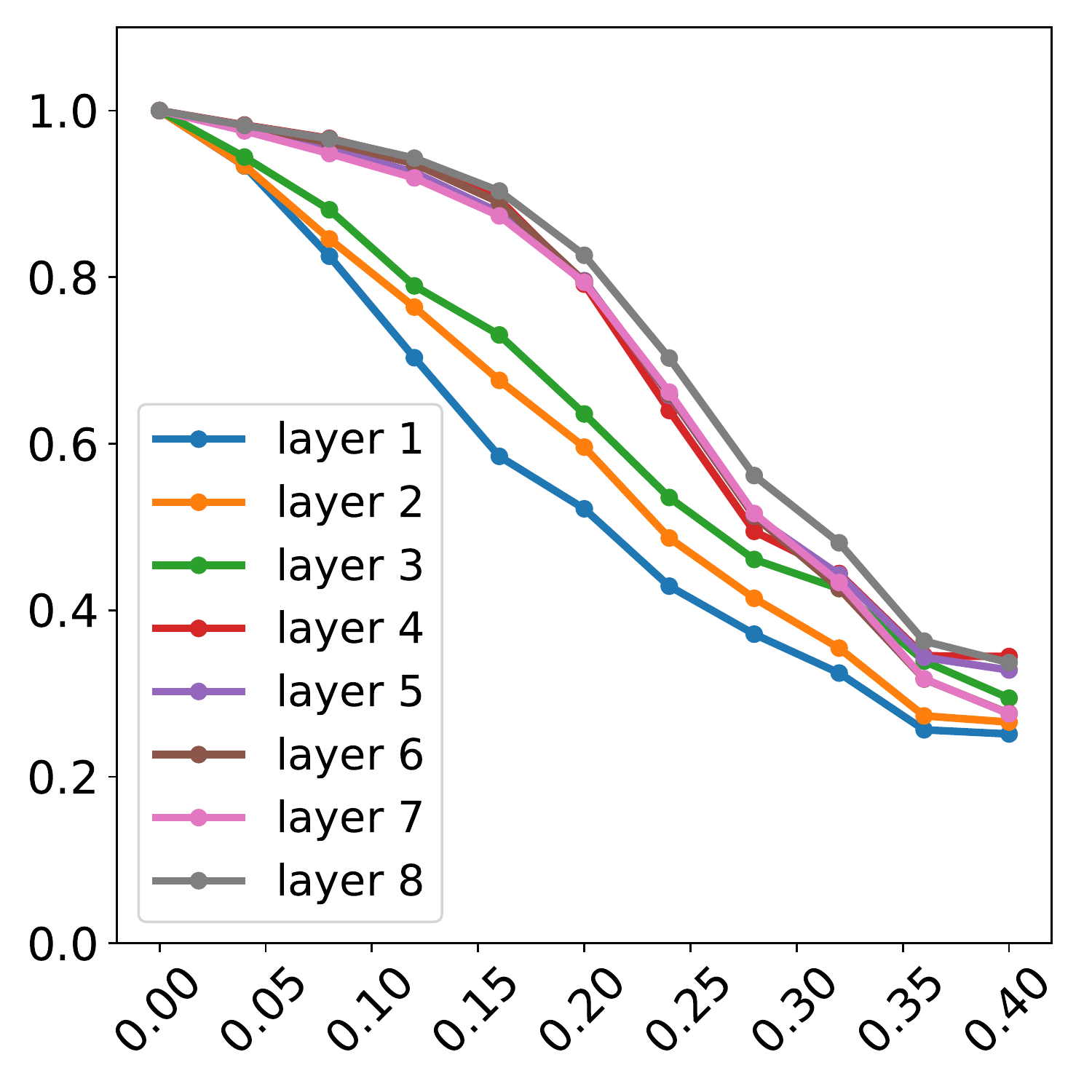}}
	\caption{Capsule activation correlation for a CapsNet with eight routing layers, trained on AffNIST. The correlation decreases with stronger affine transformation. Left to right: rotation, shearing, scaling, translation in x-direction, translation in y-direction. Shown are the means and the standard deviations of capsule activation for all layers except the last over ten trained models.}
	\label{fig:affnist:corss_corr:custom_16_32_8_model}
\end{figure*}

In this section, we report the results from additional experiments on the viewpoint invariance of the parse-tree. For experiment details, see Section~\ref{sec:viewpoint} of the main paper.
Figure~\ref{fig:affnist:corss_corr:hinton_orig_model} shows the capsule activation correlation for the original CapsNet model~\citep{nips/SabourFH17}.
Since it was defined for an input image size of 28$\times$28, we rescaled all the 40$\times$40 AffNIST images down to 28$\times$28.
Figure~\ref{fig:affnist:corss_corr:custom_16_8_1_model} shows the capsule activation correlation for a CapsNet with one routing layer, 16 PrimeCaps and a capsule dimension of eight.
Figure~\ref{fig:affnist:corss_corr:custom_16_32_8_model} shows the capsule activation correlation for an eight layer CapsNet with 16 capsules per layer and a capsule dimension of 32.
We observe a substantial decrease in capsule activation correlation for all models, similar to the results reported in the main paper.

\clearpage
\section{AffNIST: Additional Results for the Model from the Main Paper}\label{app:affnist_single}
\begin{figure}[!ht]
	\begin{center}
		\begin{tabular}{cccccc}
			\includegraphics[width=0.15\linewidth]{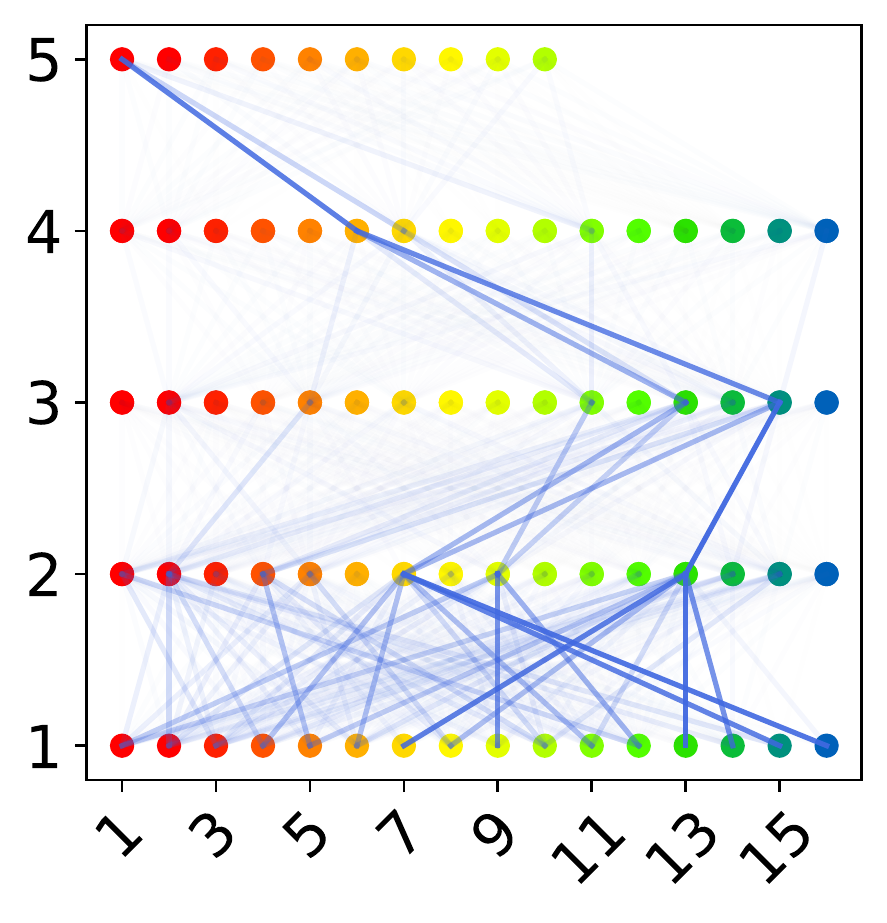} &
			\includegraphics[width=0.15\linewidth]{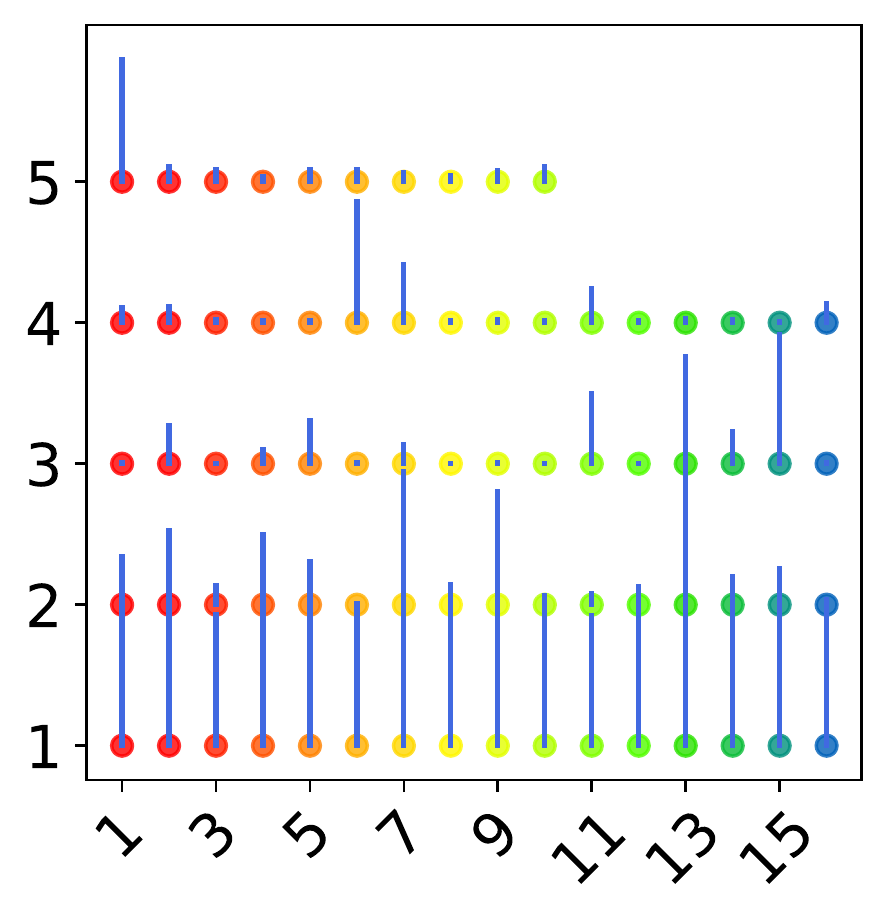} &
			\includegraphics[width=0.15\linewidth]{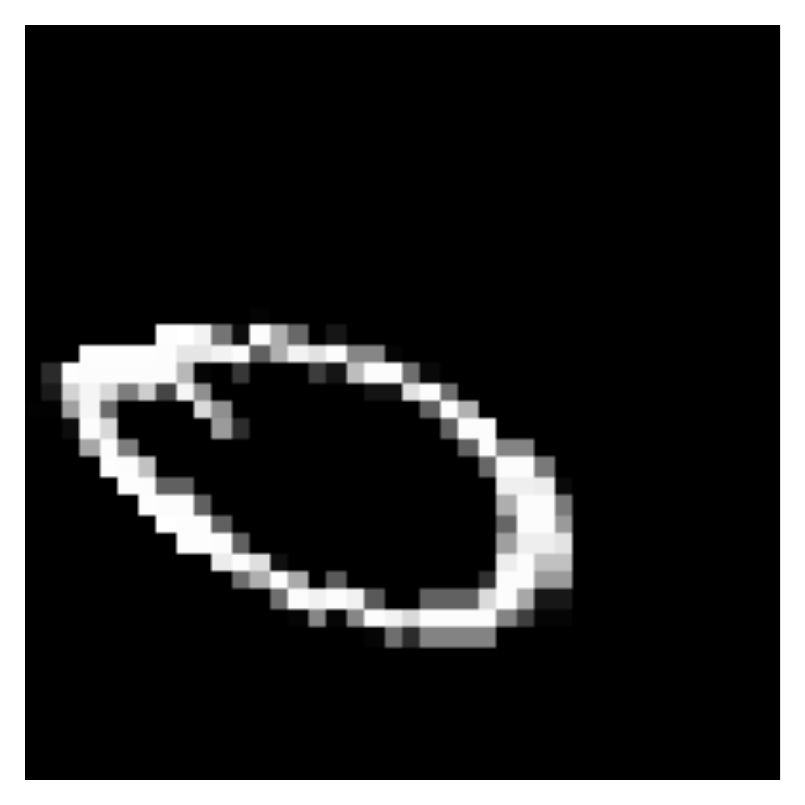} &
			\includegraphics[width=0.15\linewidth]{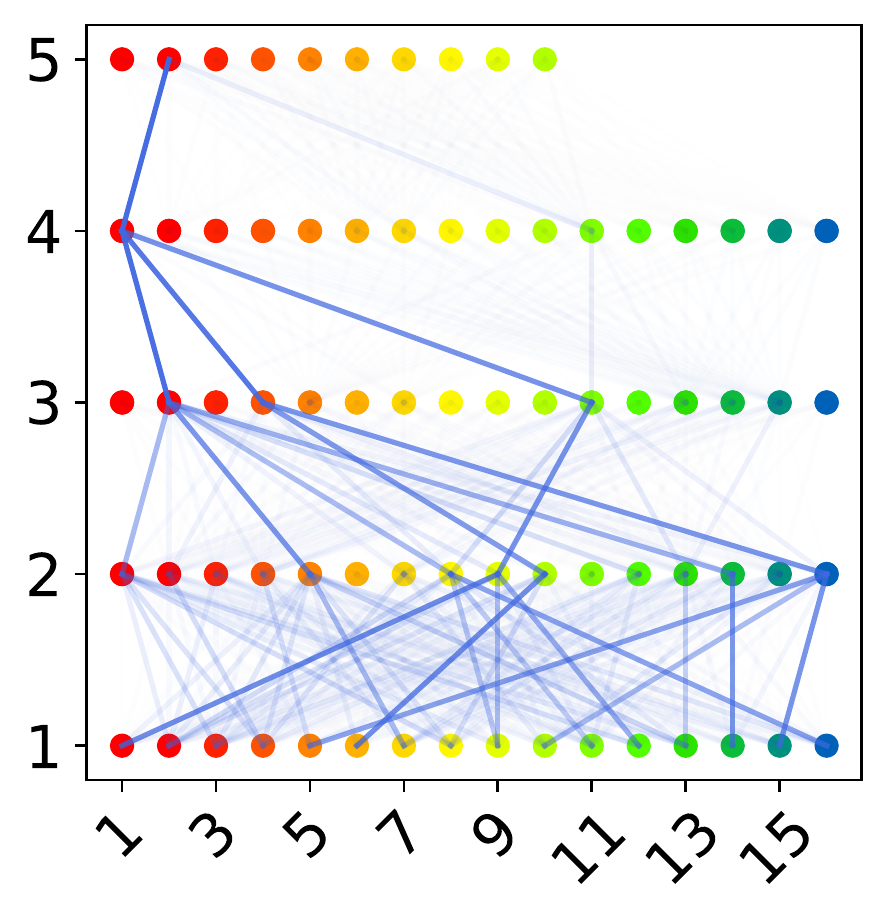} &
			\includegraphics[width=0.15\linewidth]{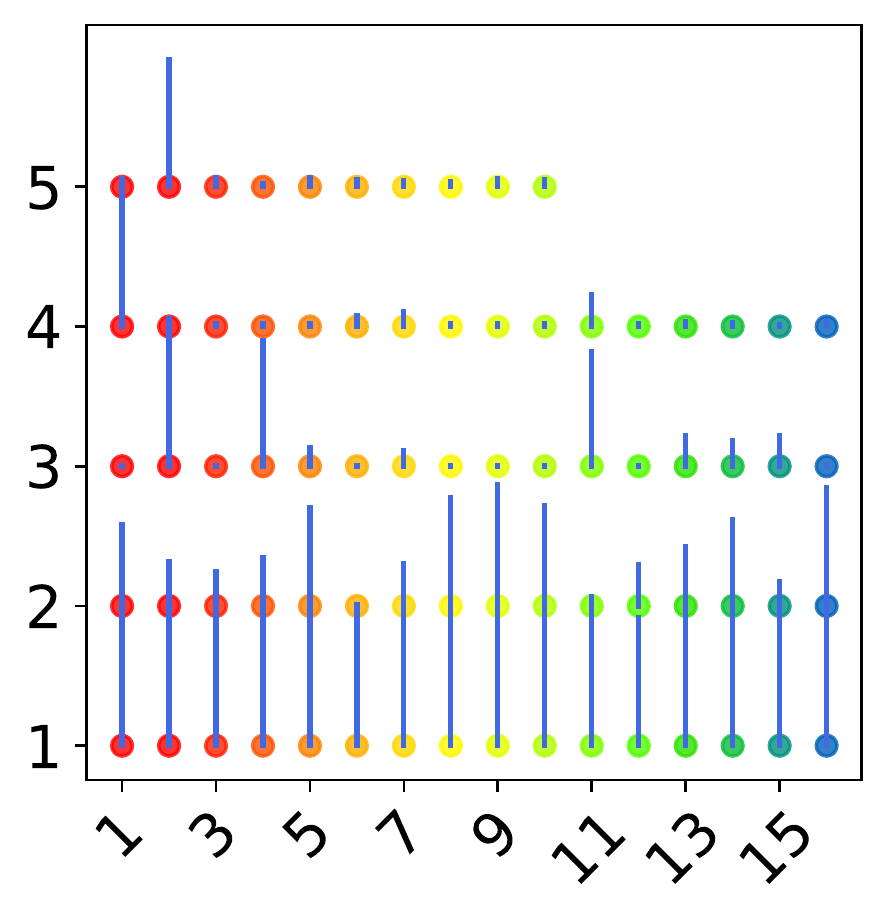} &
			\includegraphics[width=0.15\linewidth]{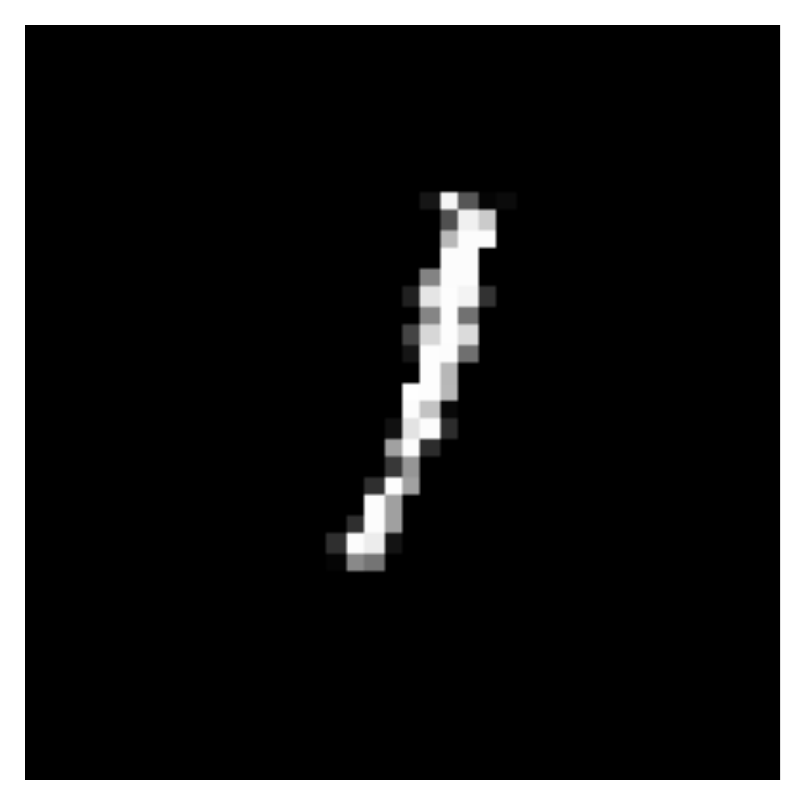} \\
			\includegraphics[width=0.15\linewidth]{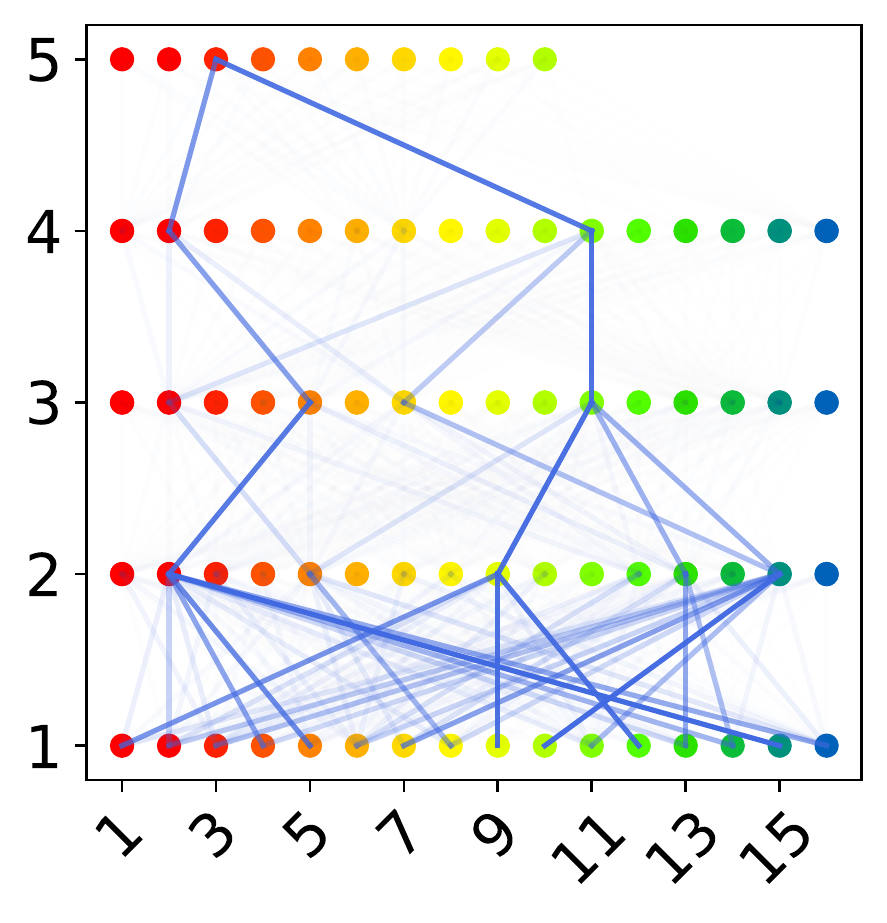} &
			\includegraphics[width=0.15\linewidth]{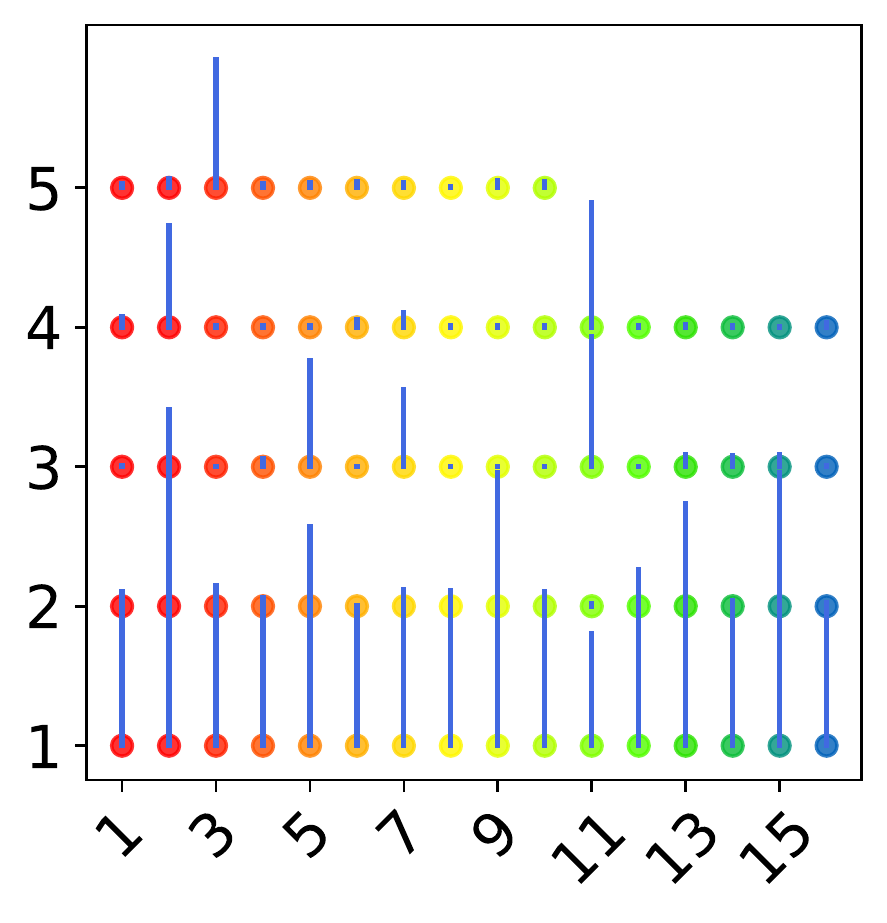} &
			\includegraphics[width=0.15\linewidth]{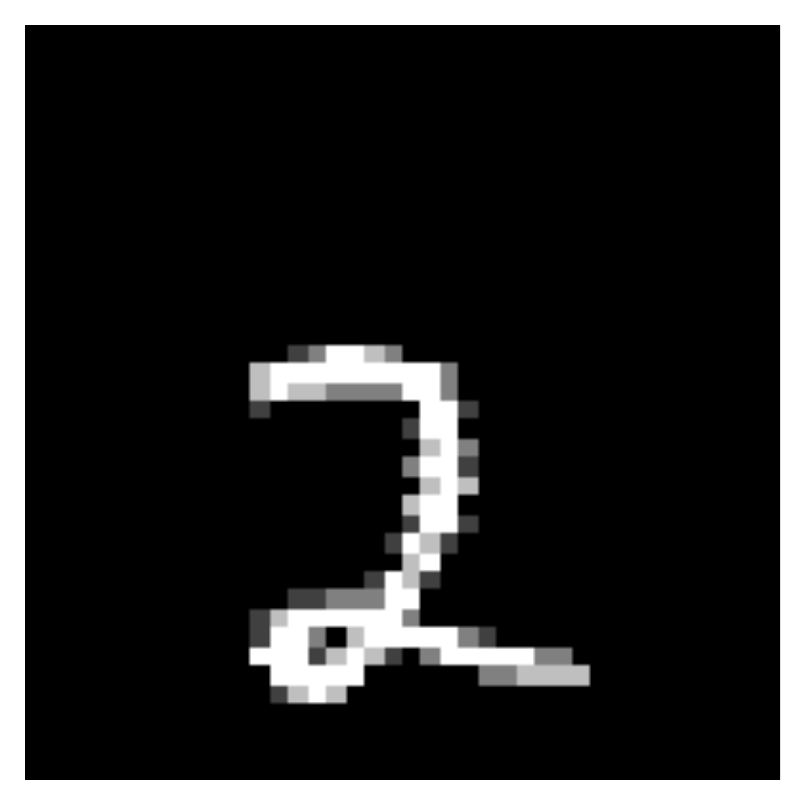} &
			\includegraphics[width=0.15\linewidth]{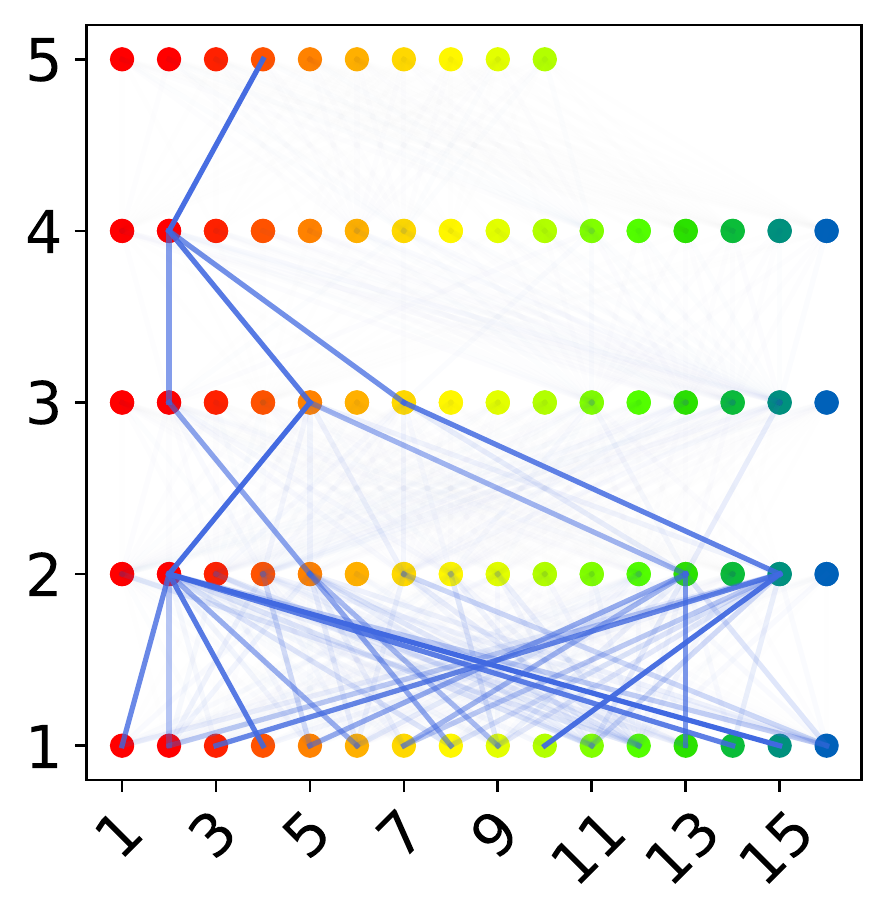} &
			\includegraphics[width=0.15\linewidth]{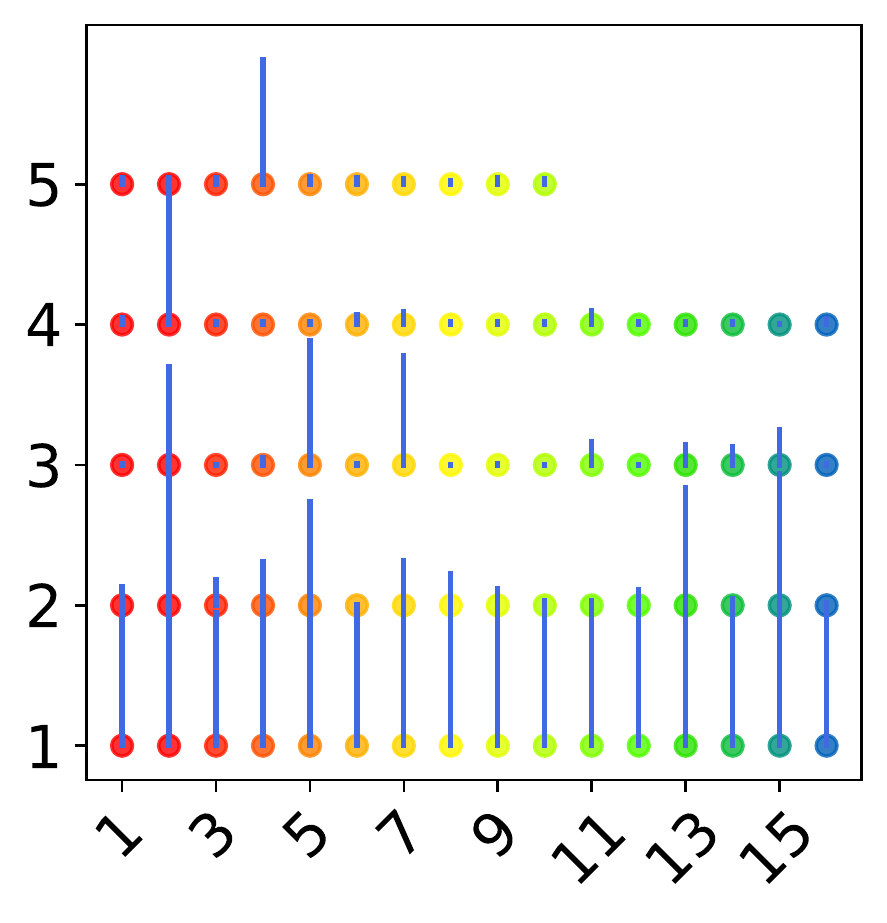} &
			\includegraphics[width=0.15\linewidth]{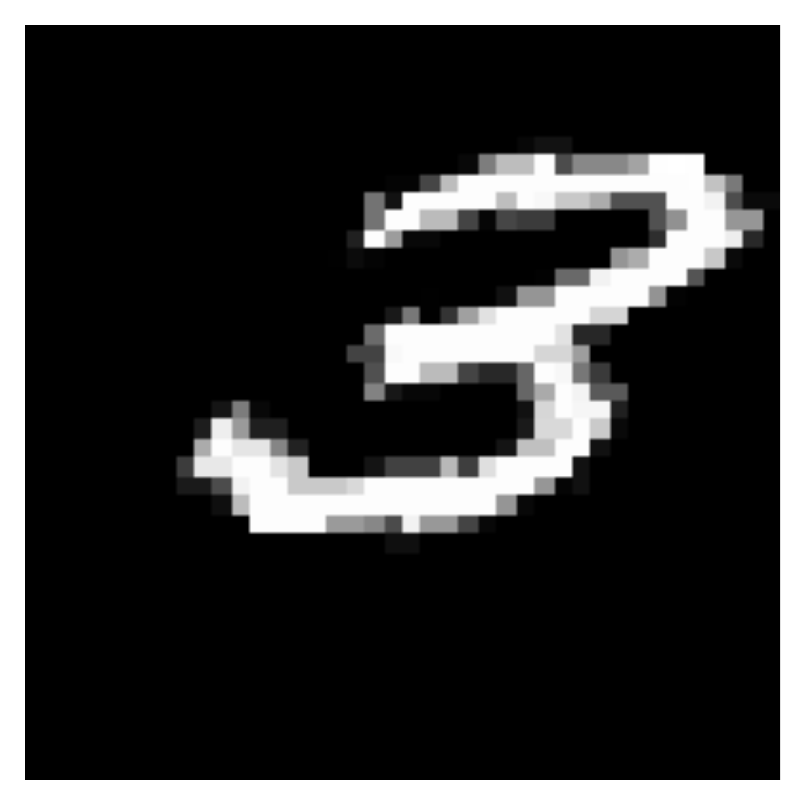} \\			
			\includegraphics[width=0.15\linewidth]{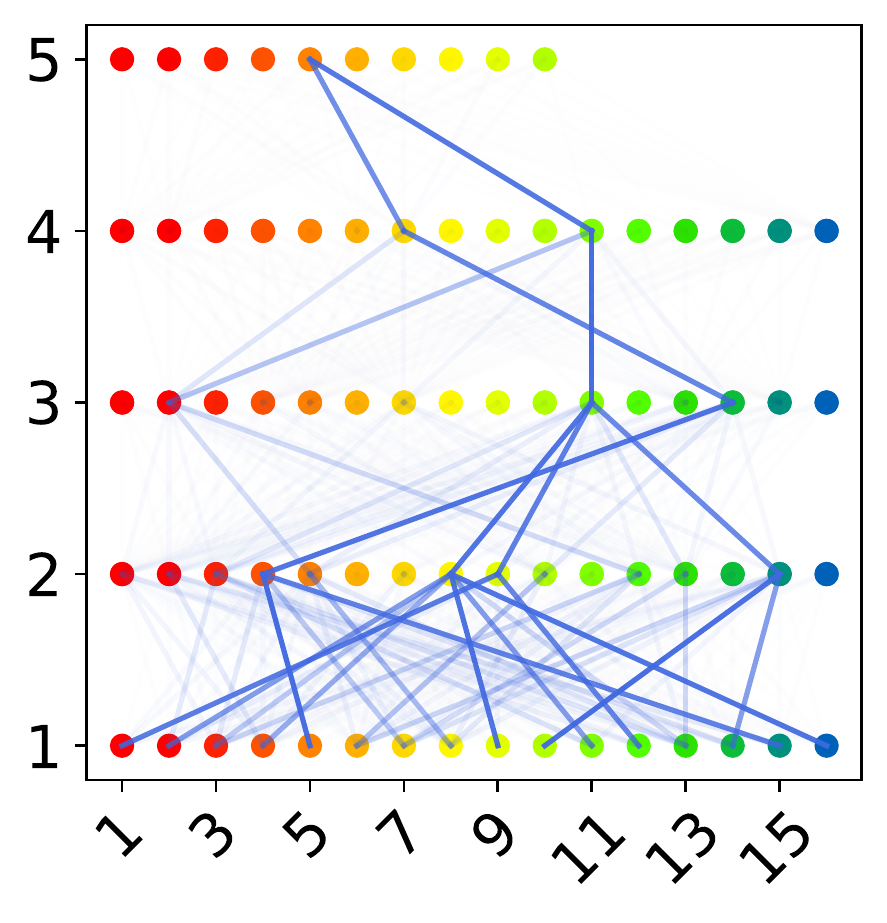} &
			\includegraphics[width=0.15\linewidth]{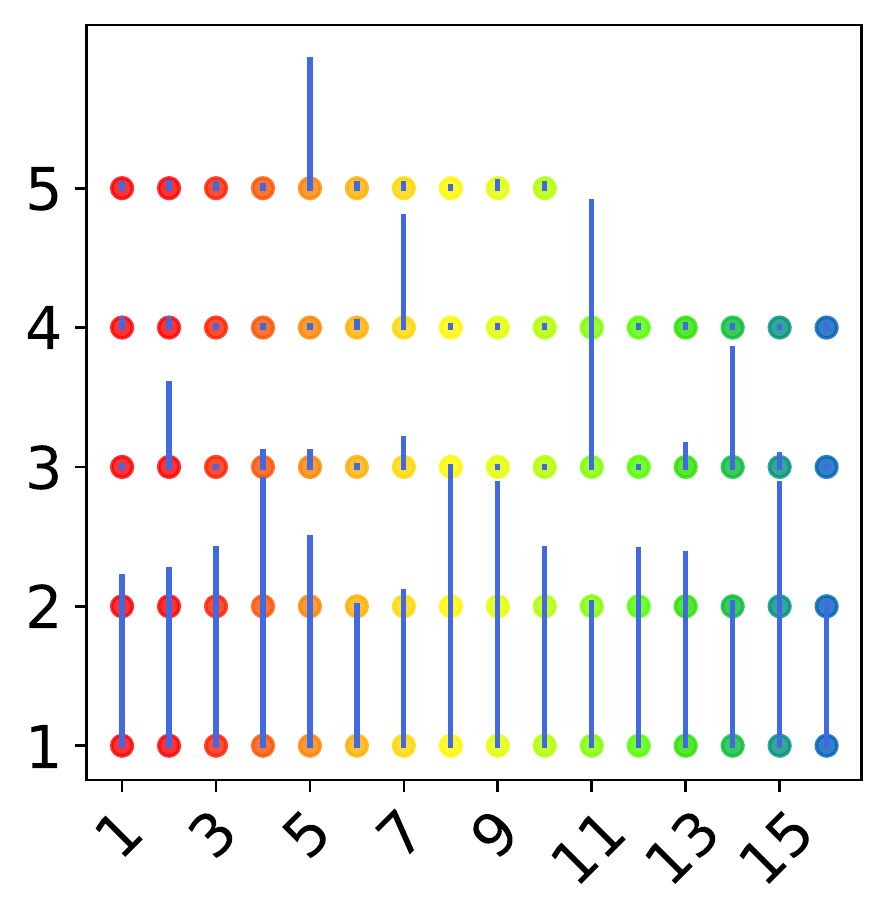} &
			\includegraphics[width=0.15\linewidth]{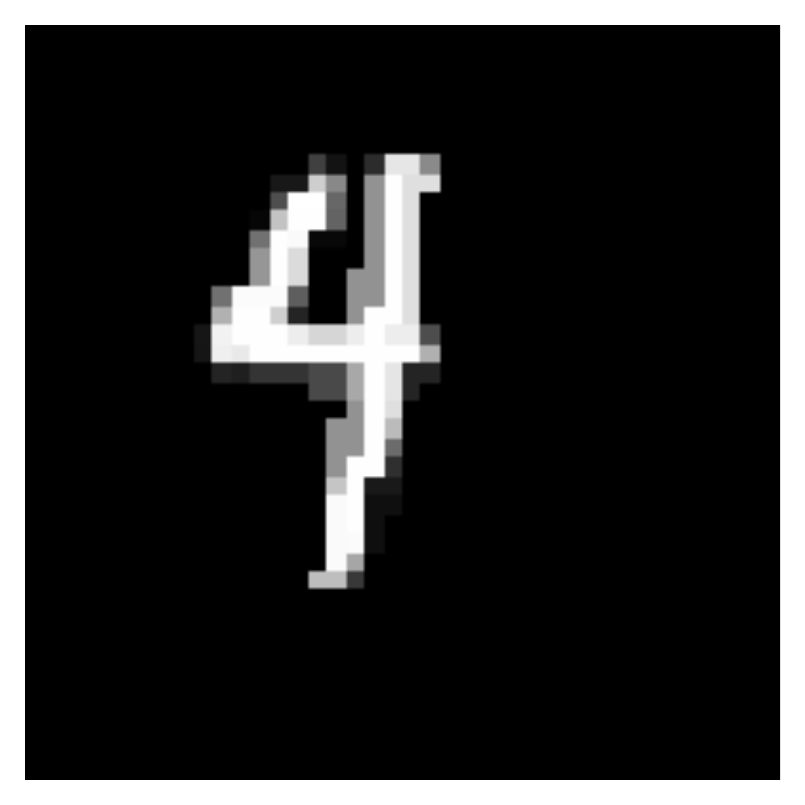} &
			\includegraphics[width=0.15\linewidth]{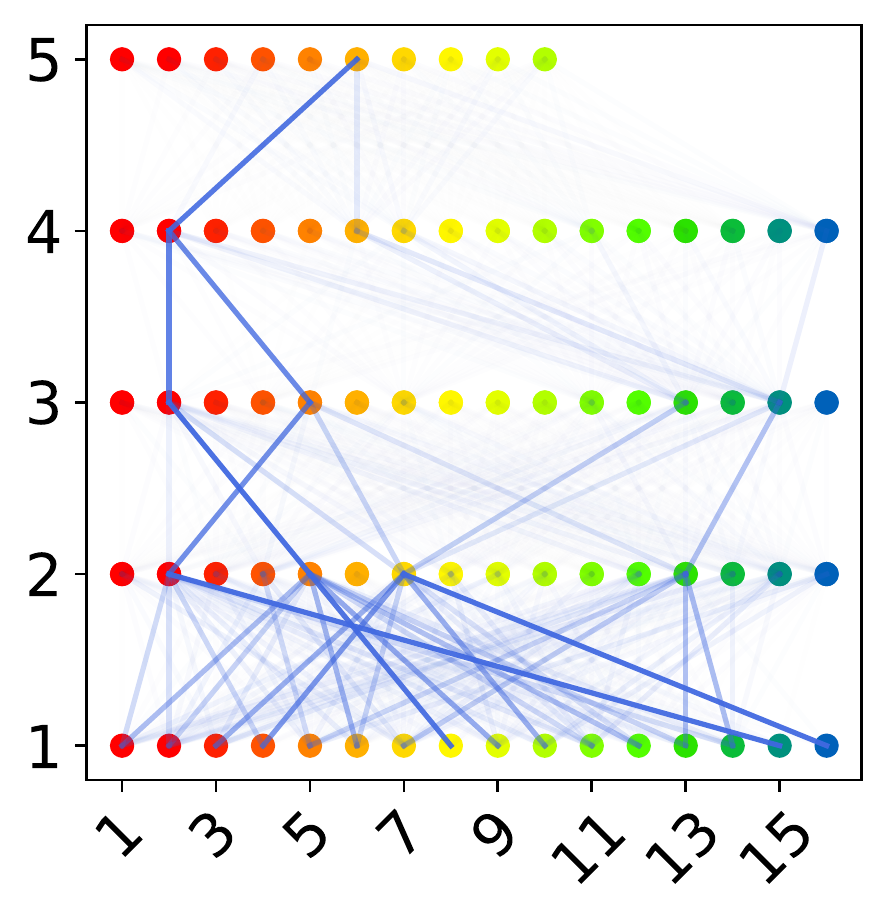} &
			\includegraphics[width=0.15\linewidth]{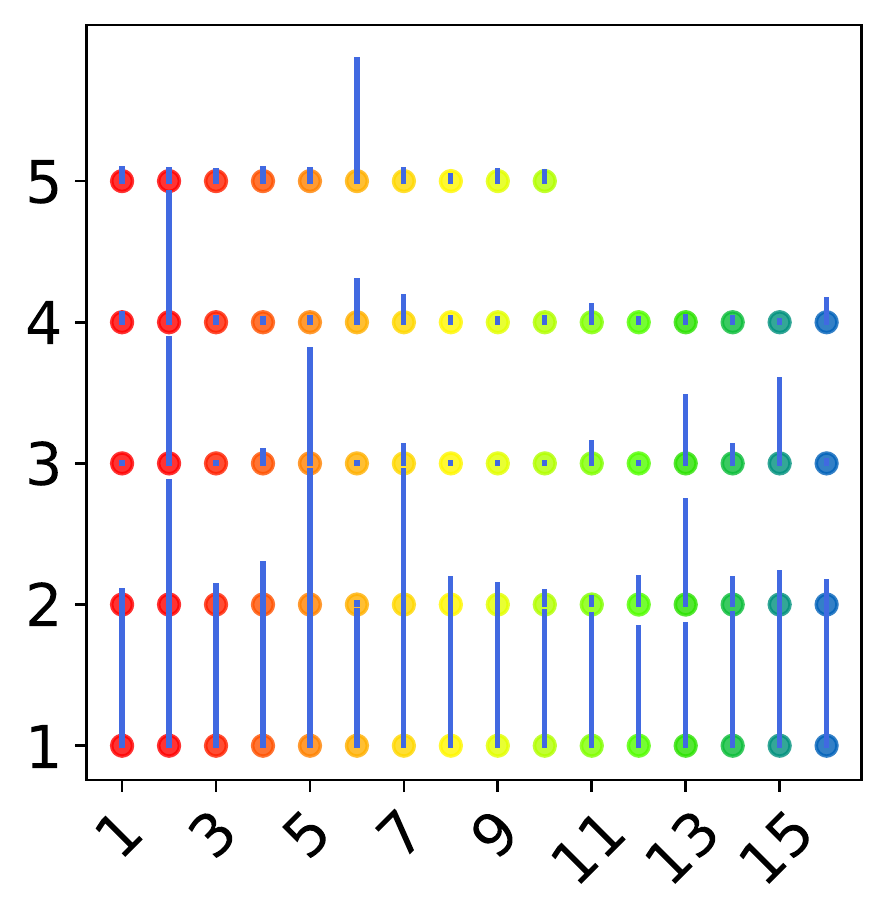} &
			\includegraphics[width=0.15\linewidth]{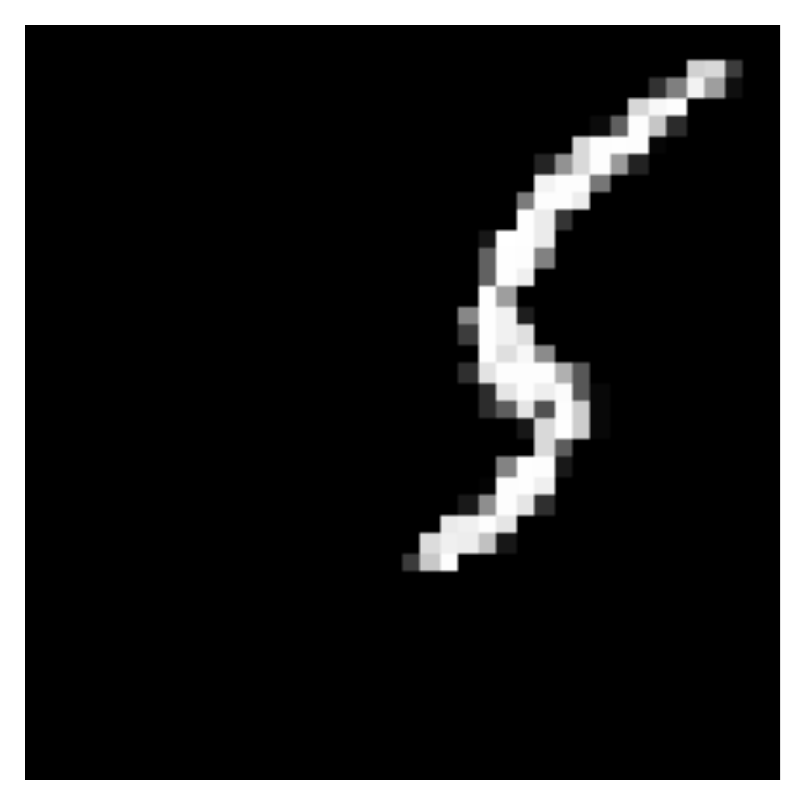} \\
			\includegraphics[width=0.15\linewidth]{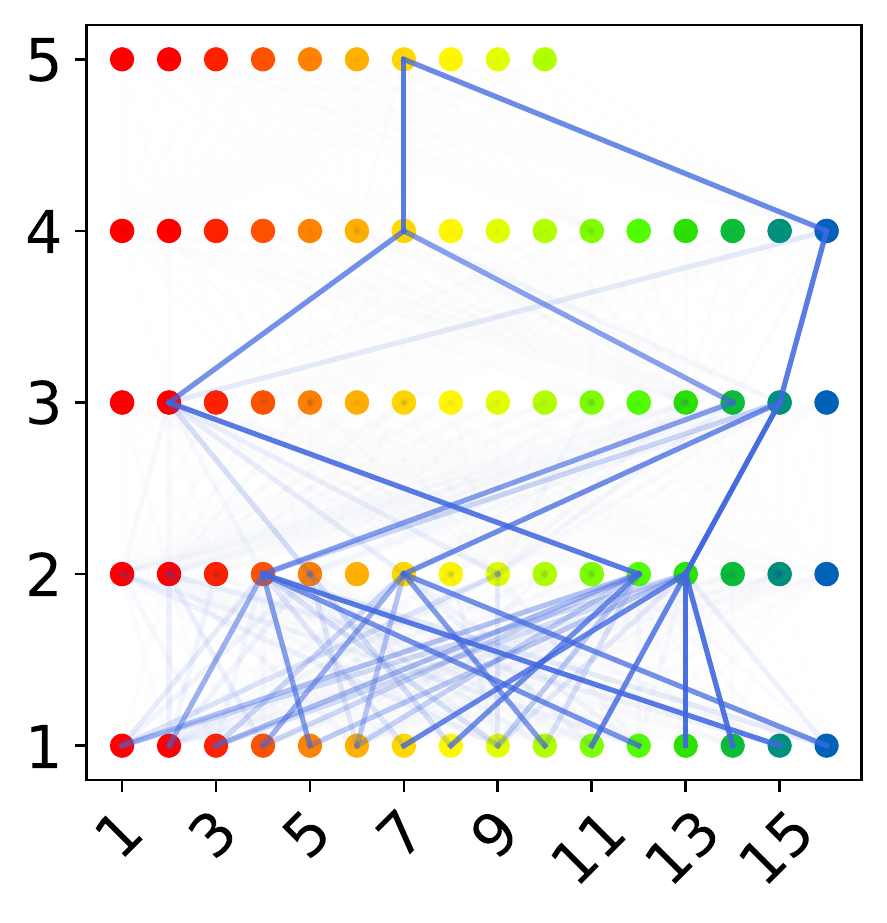} &
			\includegraphics[width=0.15\linewidth]{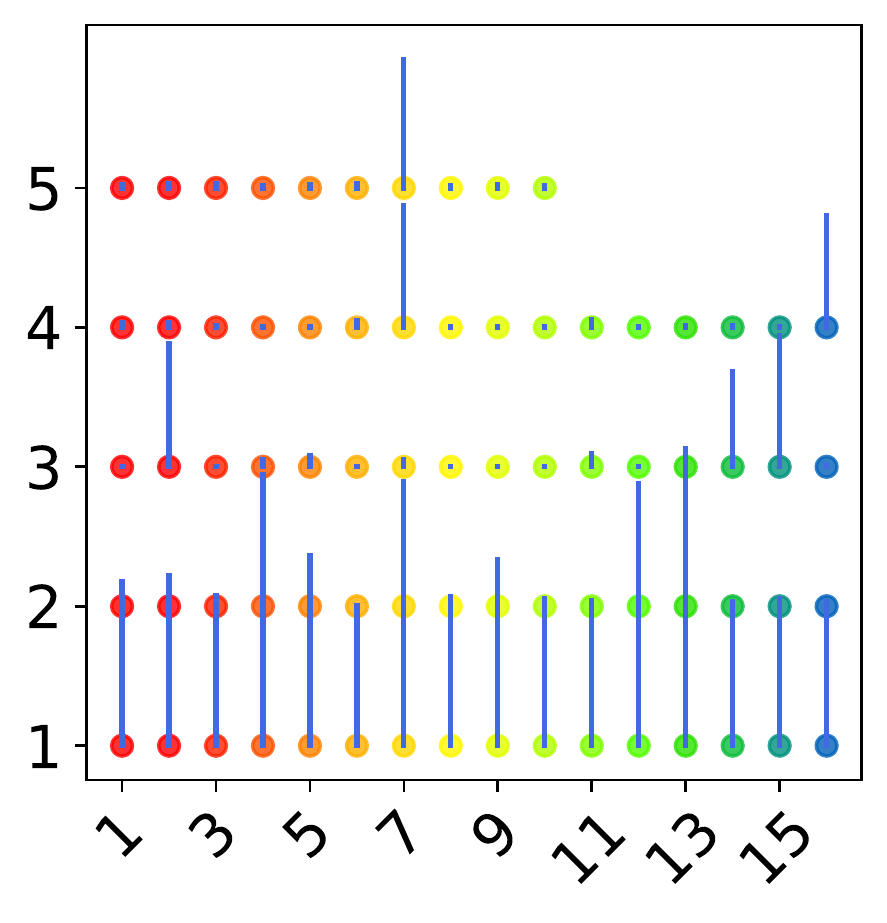} &
			\includegraphics[width=0.15\linewidth]{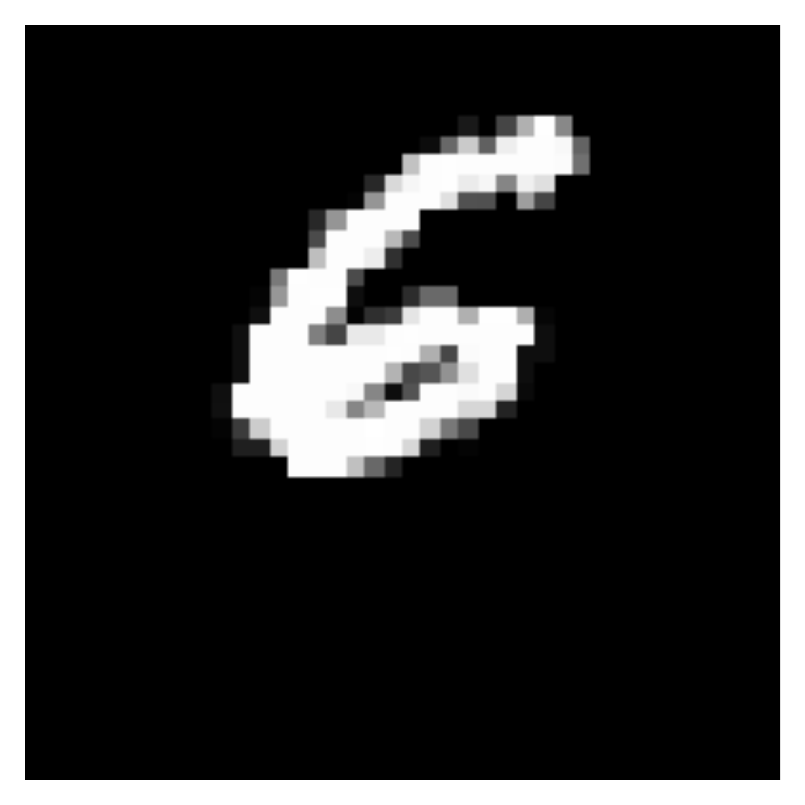} &
			\includegraphics[width=0.15\linewidth]{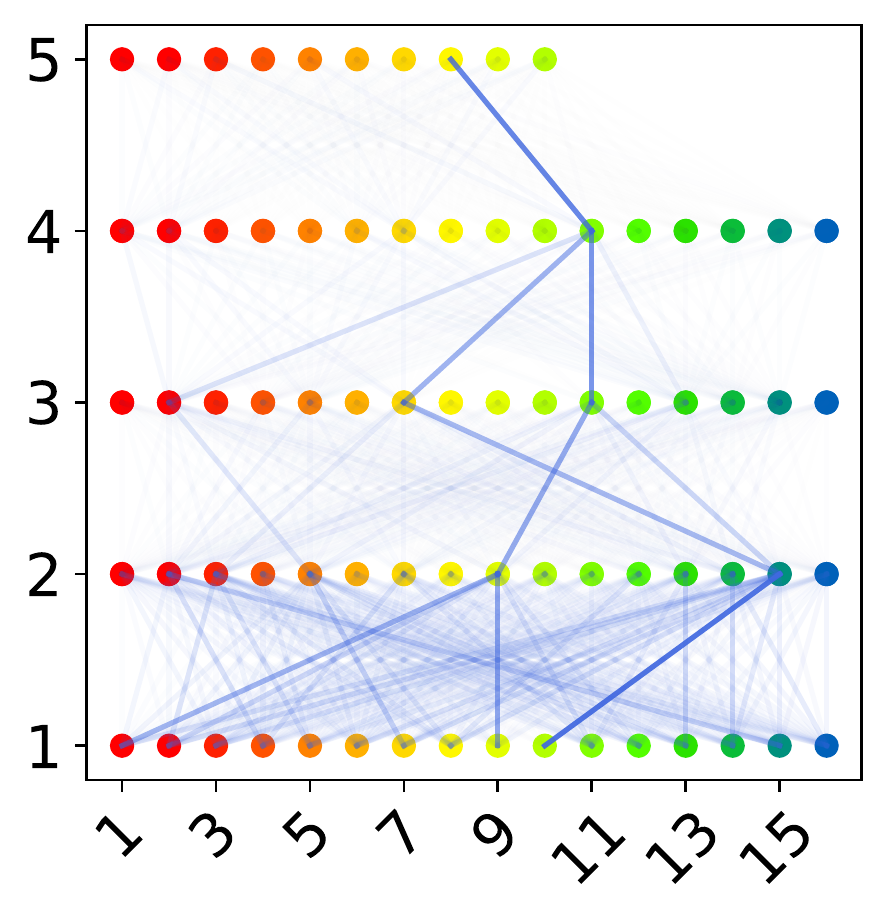} &
			\includegraphics[width=0.15\linewidth]{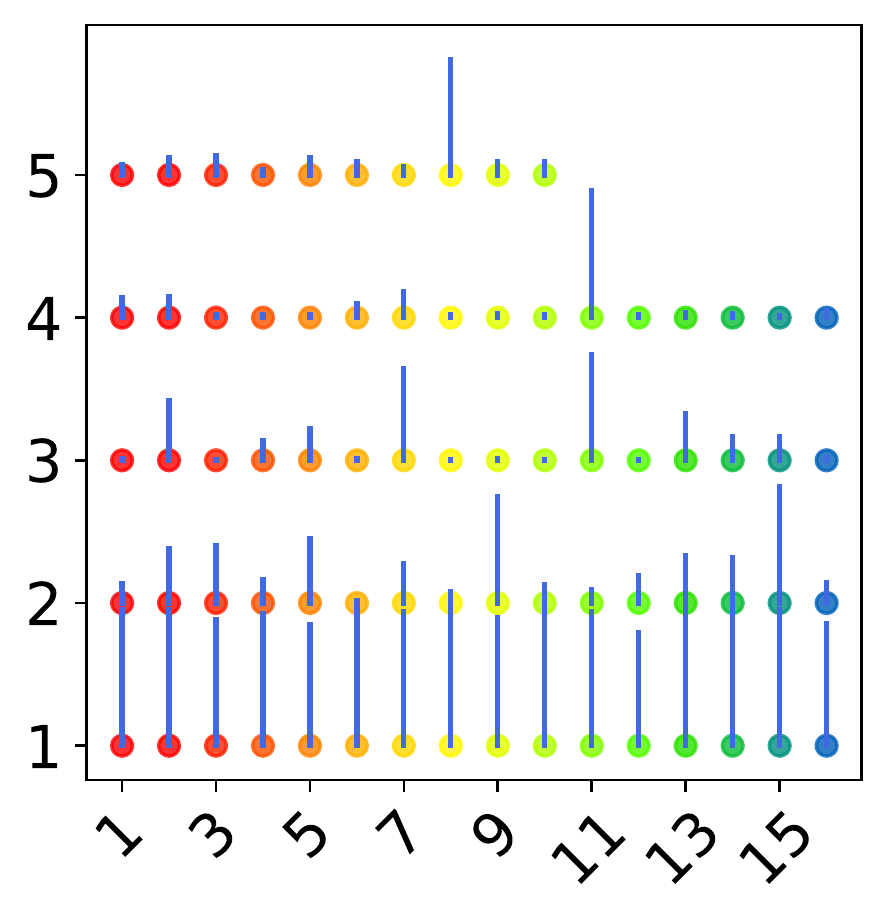} &
			\includegraphics[width=0.15\linewidth]{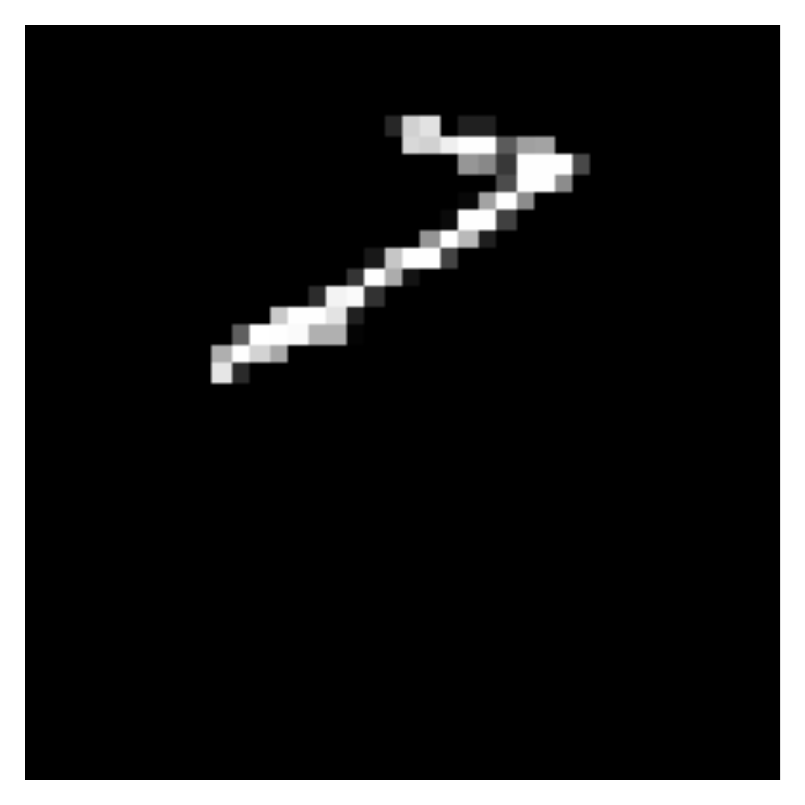} \\
			\includegraphics[width=0.15\linewidth]{figs/affnist/single_run/samples/8_sample_470_couplings.pdf} &
			\includegraphics[width=0.15\linewidth]{figs/affnist/single_run/samples/8_sample_470_capsules.pdf} &
			\includegraphics[width=0.15\linewidth]{figs/affnist/single_run/samples/8_sample_470_img.pdf} &
			\includegraphics[width=0.15\linewidth]{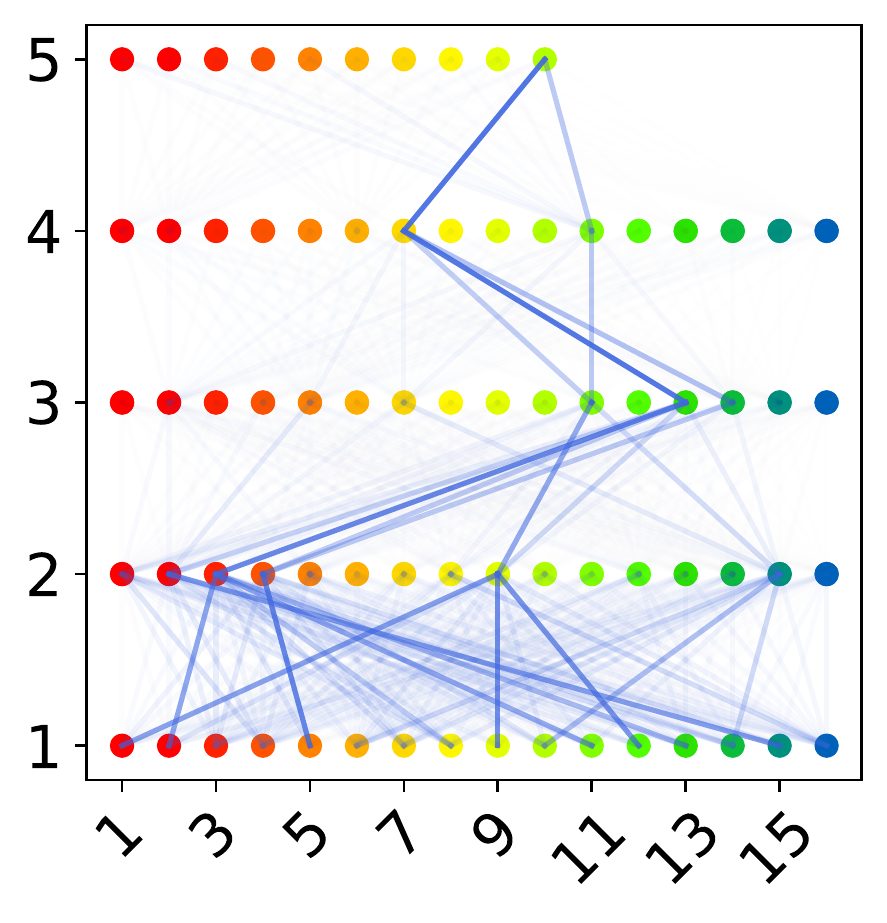} &
			\includegraphics[width=0.15\linewidth]{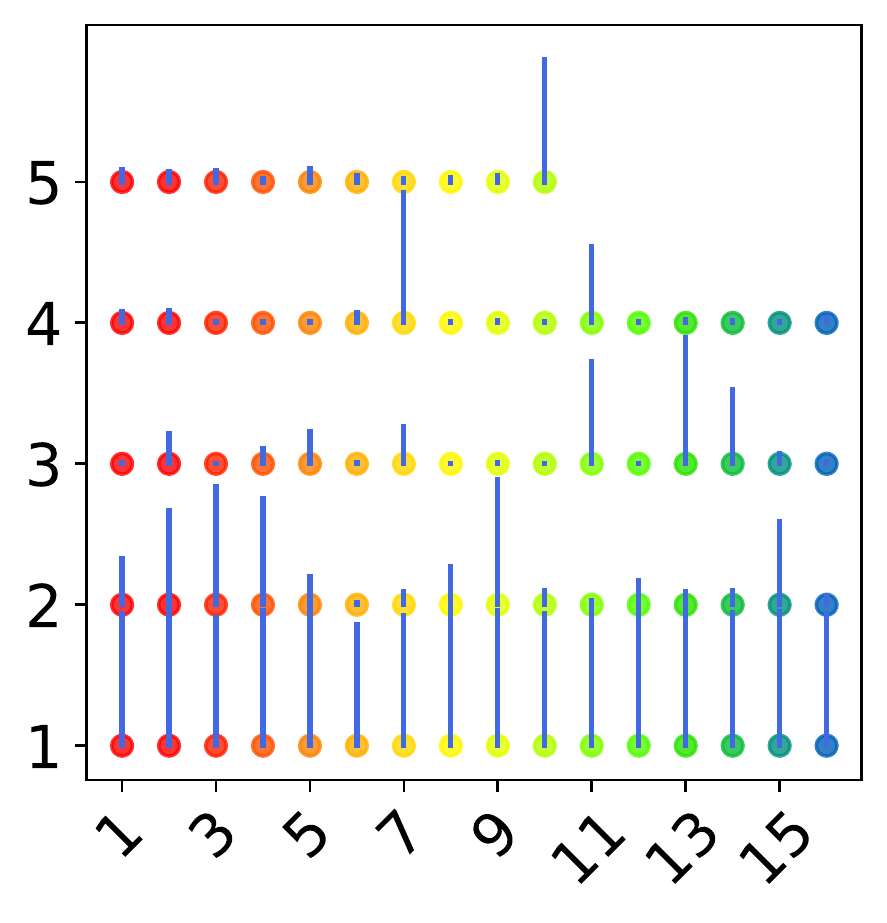} &
			\includegraphics[width=0.15\linewidth]{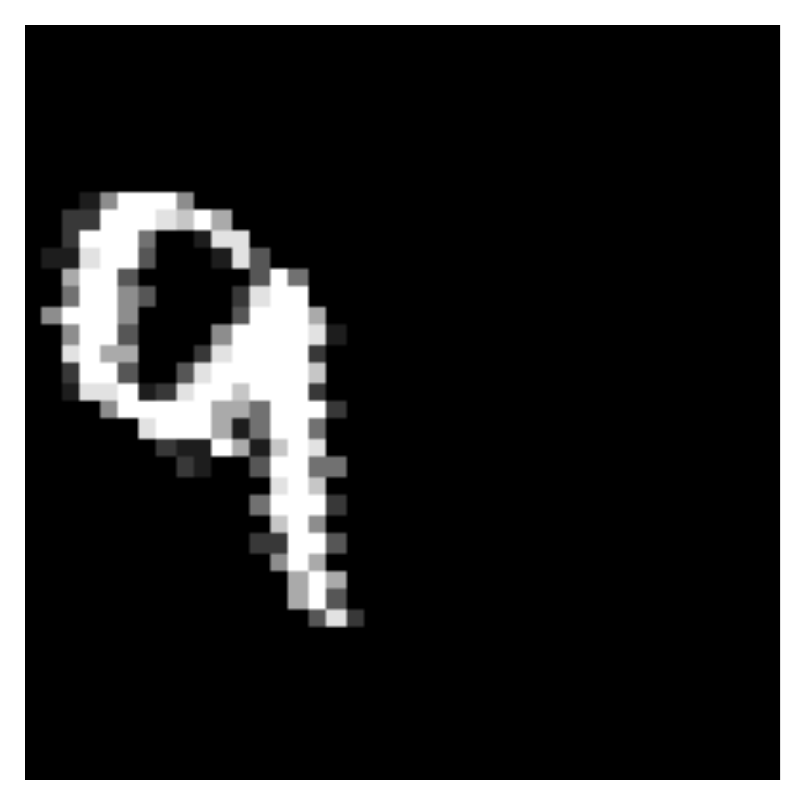} \\
		\end{tabular}
	\end{center}
	\caption{The parse-trees for AffNIST validation set samples.}
	\label{fig:affnist:main_model:all_classes}
\end{figure}

In this section, we report additional results about the model architecture used in the main paper.
The model architecture consists of five capsule layers. Each layer contains 16 capsules, except the last, where we set the number of capsules to match the number of digits, namely ten.
We use a capsule dimension of eight.
We trained a total of ten models with these settings.
We trained the model as described in Section~\ref{app:model_architectures}.
The activation and dynamics statistics of these ten models are reported in Tables~\ref{tab:affnist:main_model:dynamics} and~\ref{tab:affnist:main_model:activation} of the main paper.
Furthermore, we selected one of these models to create Figures~\ref{fig:affnist:main_model:class_samples}, ~\ref{fig:affnist:main_model:parse_tree_stats}, ~\ref{fig:affnist:main_model:view_point_invariance}, ~\ref{fig:affnist:main_model:sim_image_diff_parse_tree} and Figure~\ref{fig:affnist:main_model:all_classes}. 

\begin{figure}[!htb]%
	\centering
	\includegraphics[width=1.0\textwidth]{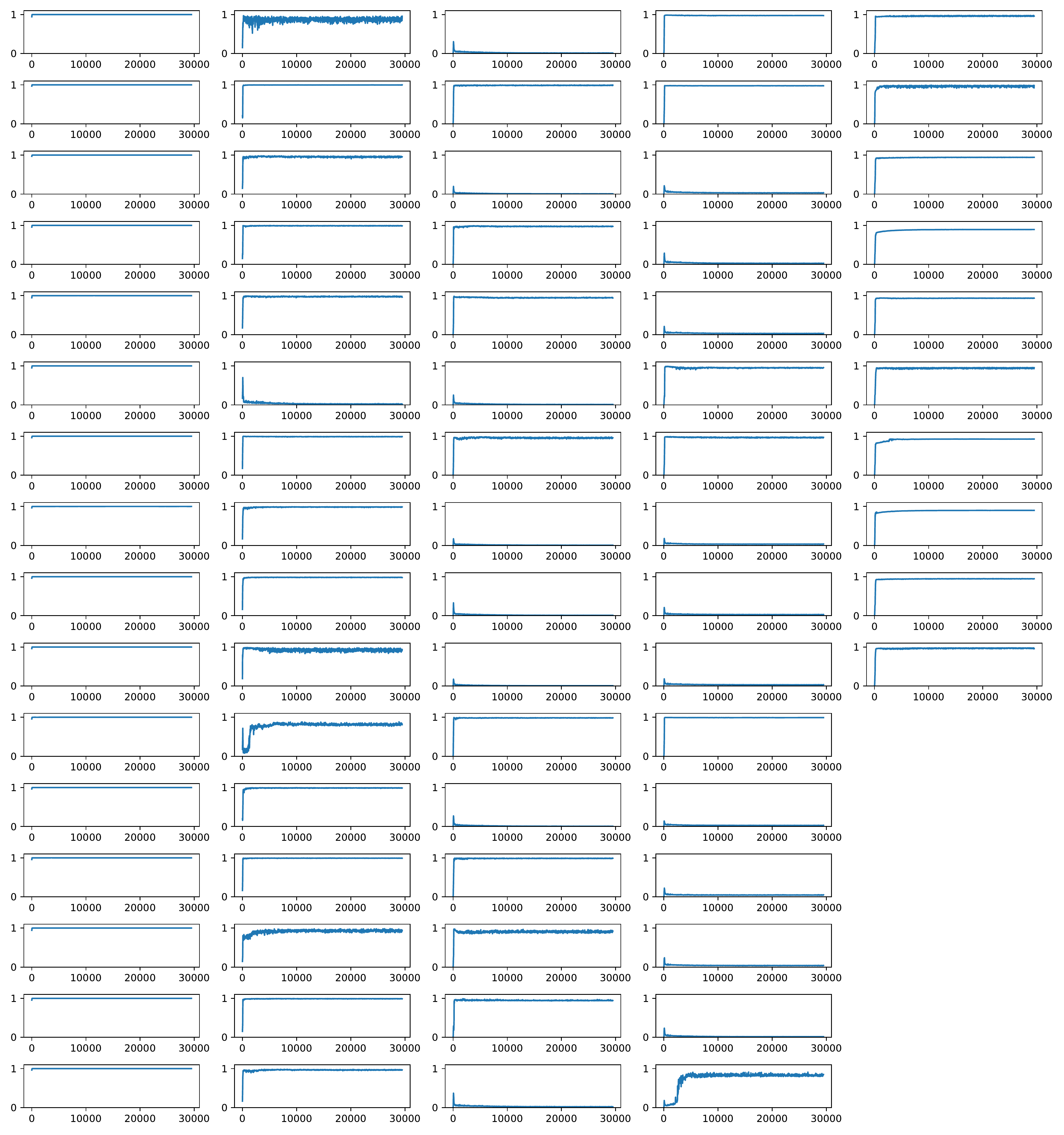}
	\caption{The maximal capsule norm/activation per batch over the training period of the model. The capsule layers are shown in the columns and the individual capsules per layer are shown in the rows.}%
	\label{fig:affnist:main_model:training:caps_norms_max}
\end{figure}

\begin{figure}[!htb]%
	\centering
	\makebox[\textwidth][c]{\includegraphics[width=1.3\textwidth]{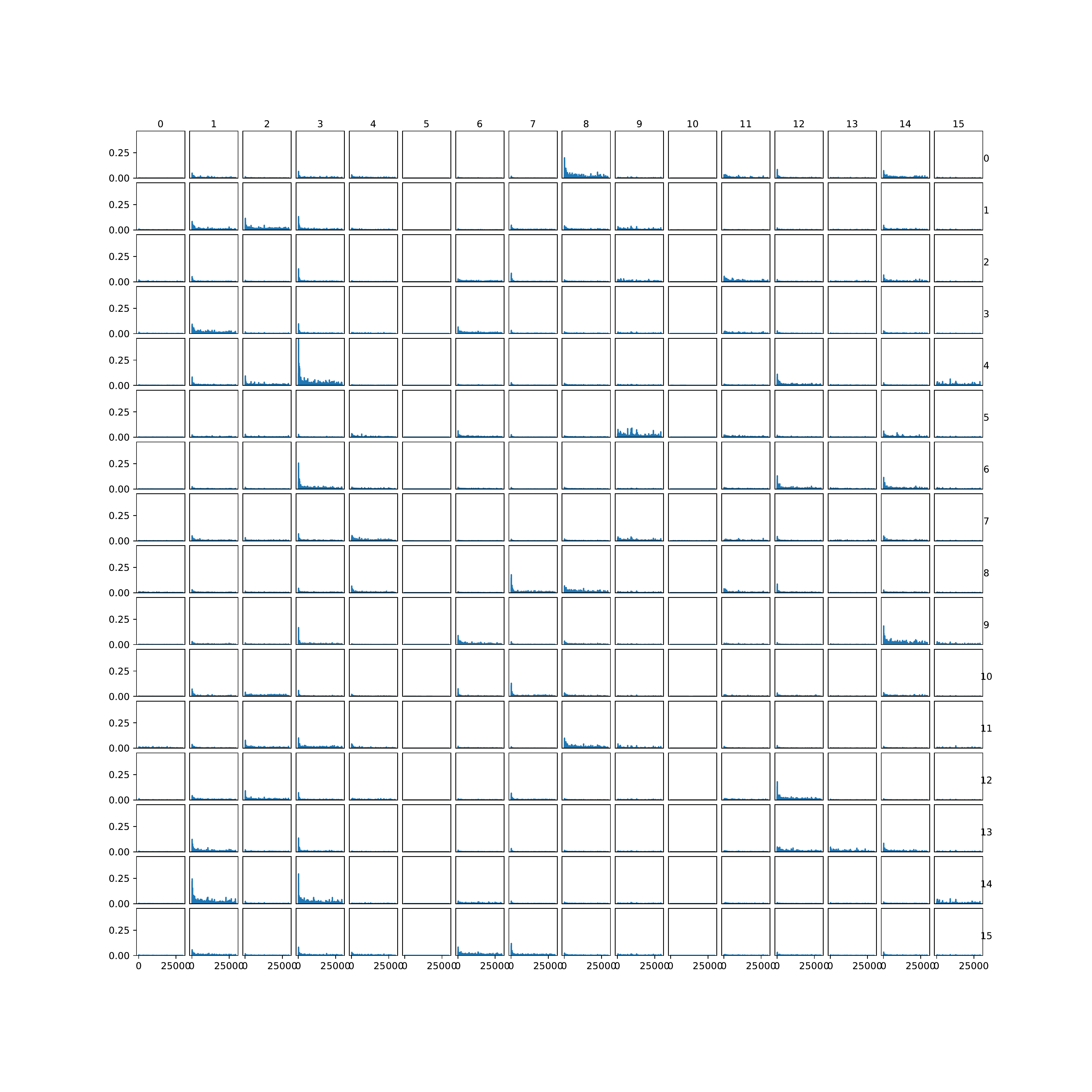}}%
	\vspace{-2cm}
	\caption{\textbf{First routing layer}: The gradient norms for the weight matrices $\frac{\partial L_m}{\partial W^1_{(j,i,:,:)}}$ over the training period of the model with lower layer capsules $i$ in the rows and upper layer capsules $j$ in the columns.}%
	\label{fig:affnist:main_model:training:grad_norms_0}
\end{figure}

\begin{figure}[!htb]%
	\centering
	\makebox[\textwidth][c]{\includegraphics[width=1.3\textwidth]{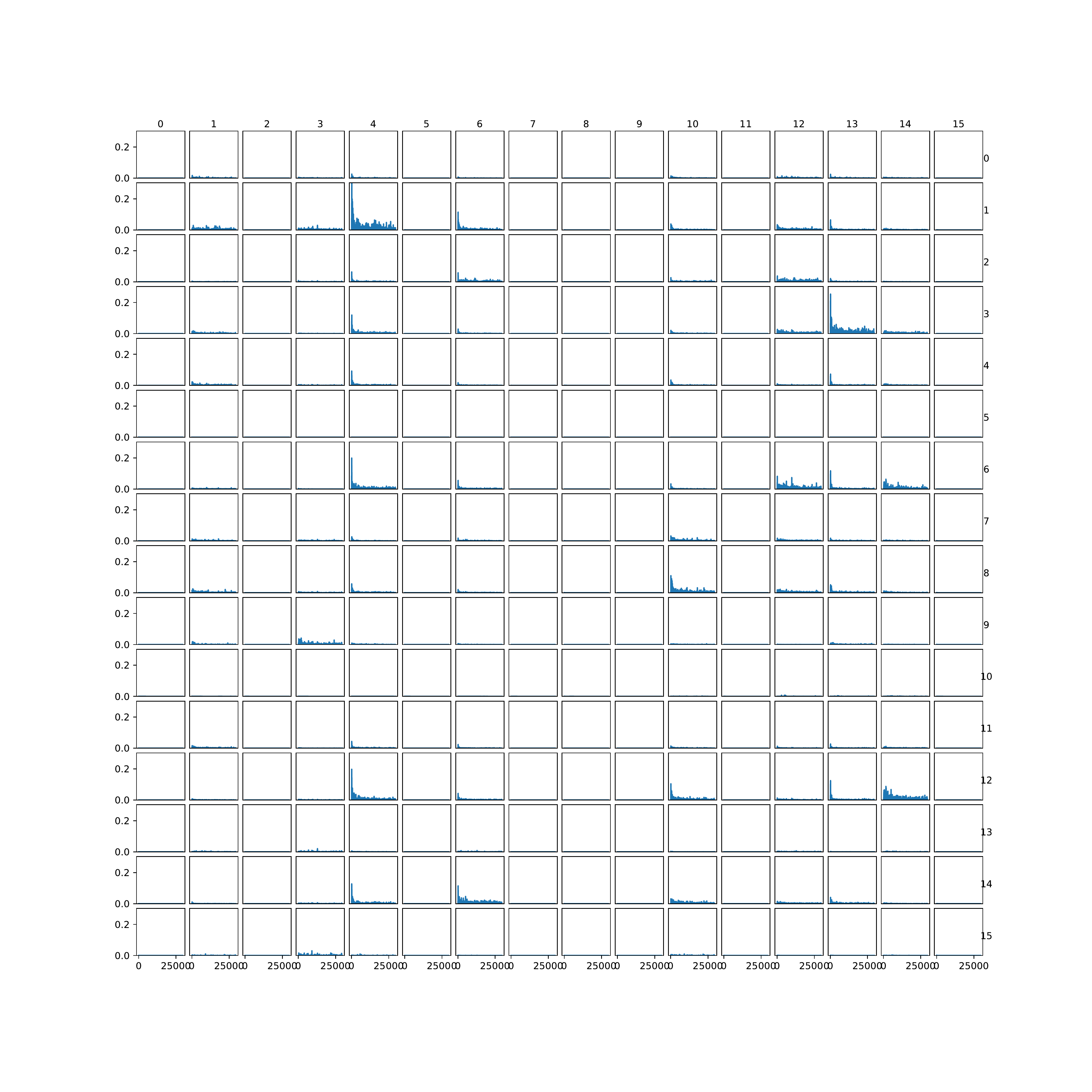}}%
	\vspace{-2cm}
	\caption{\textbf{Second routing layer}: The gradient norms for the weight matrices $\frac{\partial L_m}{\partial W^2_{(j,i,:,:)}}$ over the training period of the model with lower layer capsules $i$ in the rows and upper layer capsules $j$ in the columns.}%
	\label{fig:affnist:main_model:training:grad_norms_1}
\end{figure}

\begin{figure}[!htb]%
	\centering
	\makebox[\textwidth][c]{\includegraphics[width=1.3\textwidth]{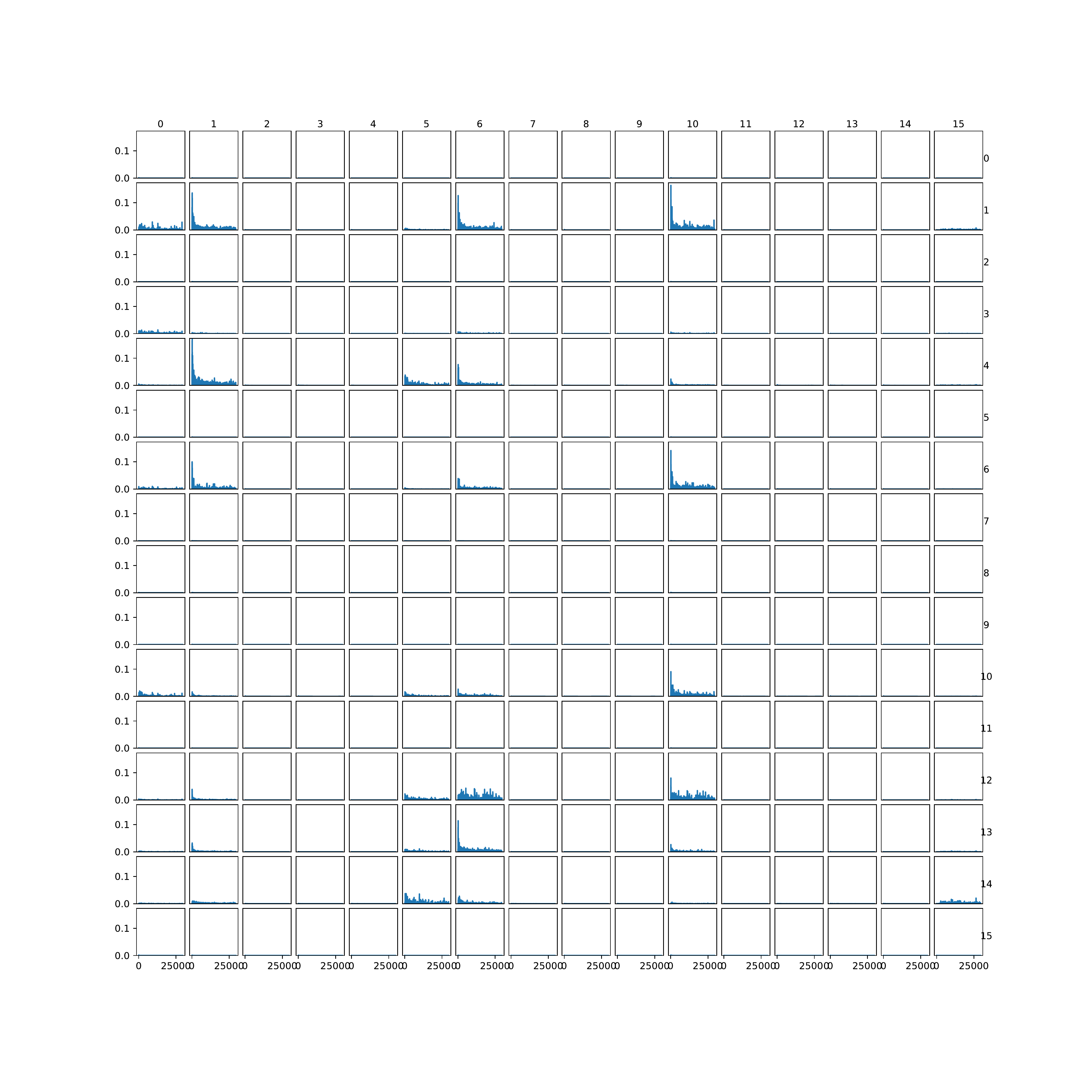}}%
	\vspace{-2cm}
	\caption{\textbf{Third routing layer}: The gradient norms for the weight matrices $\frac{\partial L_m}{\partial W^3_{(j,i,:,:)}}$ over the training period of the model with lower layer capsules $i$ in the rows and upper layer capsules $j$ in the columns.}%
	\label{fig:affnist:main_model:training:grad_norms_2}
\end{figure}

\begin{figure}[!htb]%
	\centering
	\makebox[\textwidth][c]{\includegraphics[width=0.8\textwidth]{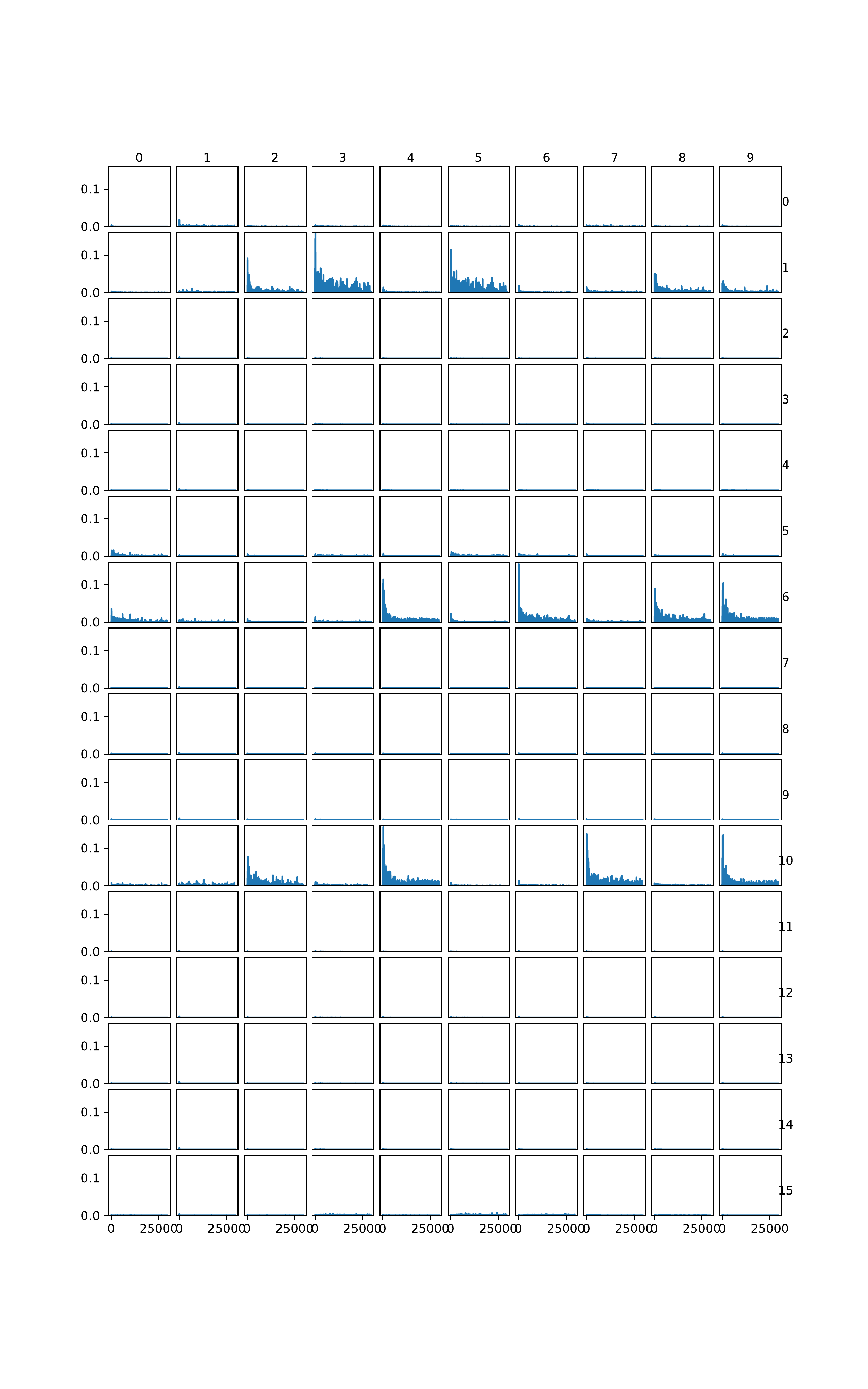}}%
	\vspace{-2cm}
	\caption{\textbf{Final routing layer}: The gradient norms for the weight matrices $\frac{\partial L_m}{\partial W^4_{(j,i,:,:)}}$ over the training period of the model with lower layer capsules $i$ in the rows and upper layer capsules $j$ in the columns.}%
	\label{fig:affnist:main_model:training:grad_norms_3}
\end{figure}

Here, we provide additional results for these models.
For the selected model, Figure~\ref{fig:affnist:main_model:training:caps_norms_max} shows the maximum norm of all the capsule vectors tracked over the training process.
We see that the norm of a capsule remains zero once it becomes zero.
This observation is explained by Theorem~\ref{thm:vanishing} in the main paper.
The norm of the gradients of all the capsule weight matrices, tracked over the training period is visualized in Figures~\ref{fig:affnist:main_model:training:grad_norms_0} to~\ref{fig:affnist:main_model:training:grad_norms_3}.
As can be seen, the weights referred to incoming and outgoing votes from and to dead capsules do not change since the norm of the gradients approaches zero, whereas, the gradient norms of weights from alive capsules to alive capsules show healthy updates.

\clearpage
\section{AffNIST: Exhaustive Experiments on Model Architectures}\label{app:affnist_grid}

\begin{figure}[!htb]%
	\centering
	\includegraphics[width=1.0\textwidth]{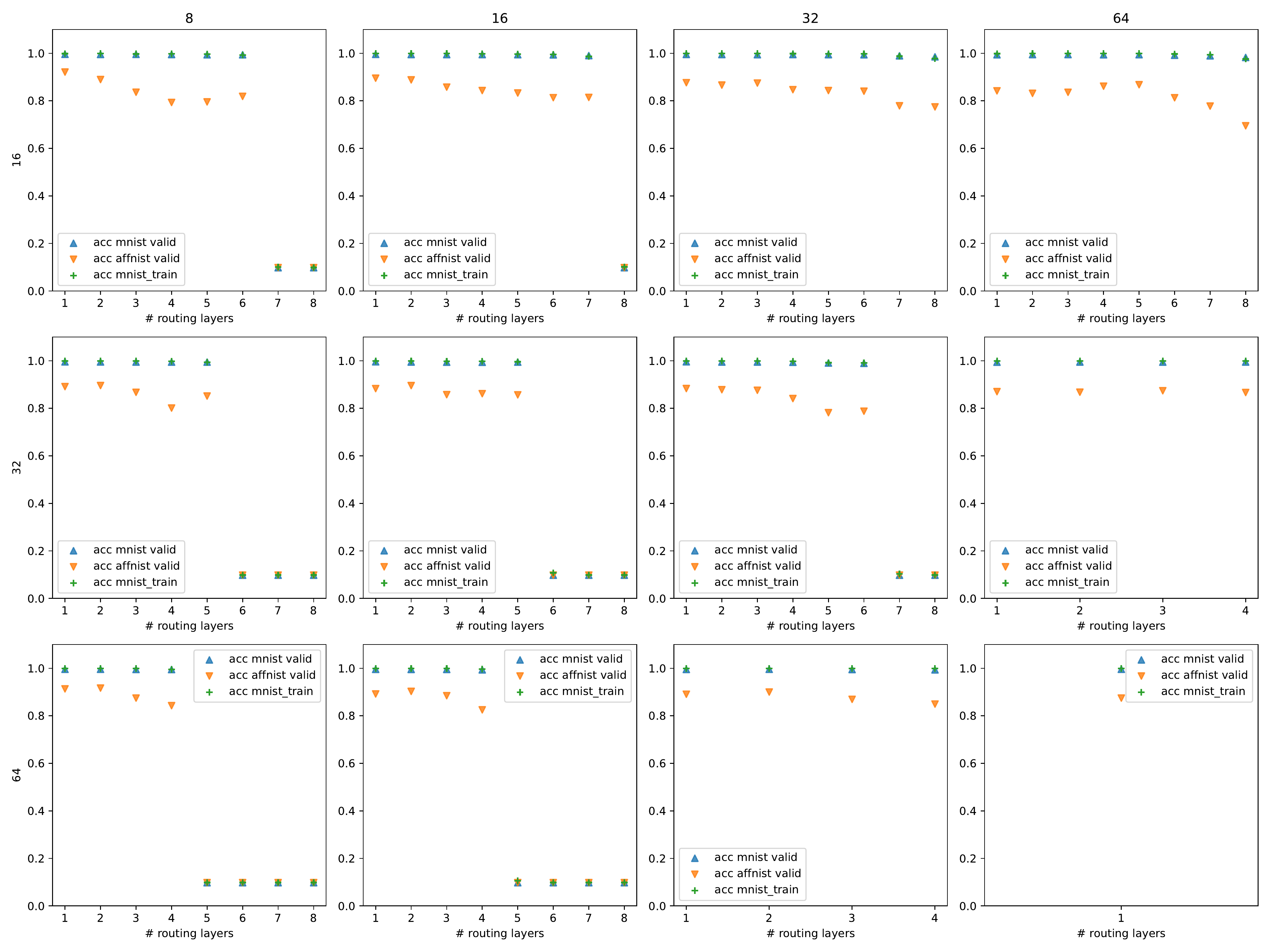}
	\caption{Reported accuracies for the AffNIST experiments.}%
	\label{fig:affnist_grid_run:all_models}
\end{figure}

In this section, we report the results of our exhaustive experiments on model architectures for the AffNIST benchmark. We trained all models following the training procedure as described in Appendix~\ref{app:model_architectures}.
We used a total of 81 different architectures with varying numbers of routing layers $\{1,2,3,4,5,6,7,8\}$, numbers of capsules per layer $\{15, 32, 64\}$ as well as capsule dimension $\{8, 32, 64\}$.
We chose these parameters to cover a broad range of settings, from simple one-layer models to complex models that used all the available GPU RAM (48GB).
We report the accuracy on the AffNIST validation set in Figure~\ref{fig:affnist_grid_run:all_models}.
Table~\ref{tab:affnist:grid_run:best_overall_models} lists the best overall models with architecture details, number of parameters and a uniform routing baselines.
The best models per depth are given in Table~\ref{tab:affnist:grid_run:best_models_depth} and the corresponding metrics and measurements are given in Tables~\ref{tab:affnist:model_d1:best},~\ref{tab:affnist:model_d2:best},~\ref{tab:affnist:model_d3:best},~\ref{tab:affnist:model_d4:best},~\ref{tab:affnist:model_d5:best},~\ref{tab:affnist:model_d6:best}.

\begin{table}[htb!]
	\centering
	\begin{tabular}{ccc|cc|cc}
		\toprule
		\multicolumn{3}{c}{Model Settings} &
		\multicolumn{2}{c}{Parameters} &
		\multicolumn{2}{c}{AffNIST Acc.} \\
		\#caps & dim & depth  & Routing  & Backbone & RBA  & Uniform  \\
		\midrule
		64 &  8 &  2 & 35 & 73 & 0.88 & 0.92 \\
		16 &  8 &  1 & 2 & 16 &  0.88 & 0.92 \\
		64 &  8 &  1 & 8 & 46 &  0.87 & 0.90 \\
		64 & 32 &  2 & 453 & 587 &  0.87 & 0.90 \\
		32 & 64 &  3 & 872 & 1006 &  0.86 & 0.89 \\
		16 & 16 &  2 & 11 & 33 &  0.86 & 0.92 \\
		32 &  8 &  2 & 11 & 33 &  0.86 & 0.90 \\
		16 &  8 &  2 & 4 & 18 &  0.86 & 0.91 \\
		64 & 16 &  2 & 122 & 192 &  0.86 & 0.90 \\
		32 & 16 &  2 & 35 & 72 &  0.85 & 0.90 \\
		32 & 16 &  1 & 8 & 46 &  0.85 & 0.89 \\
		32 &  8 &  1 & 4 & 26 &  0.85 & 0.90 \\
		64 & 32 &  1 & 33 & 167 &  0.85 & 0.88 \\
		32 & 64 &  1 & 33 & 167 &  0.85 & 0.89 \\
		64 & 64 &  1 & 66 & 329 &  0.85 & 0.88 \\
		16 & 64 &  4 & 331 & 401 &  0.85 & 0.90 \\
		32 & 16 &  4 & 87 & 125 &  0.84 & 0.92 \\
		64 & 32 &  3 & 873 & 1007 &  0.84 & 0.92 \\
		32 & 64 &  2 & 452 & 587 &  0.84 & 0.90 \\
		32 & 32 &  2 & 121 & 192 &  0.84 & 0.89 \\
		16 & 32 &  2 & 34 & 72 &  0.84 & 0.89 \\
		\bottomrule
	\end{tabular}
	\caption{Overview of the best overall models on the AffNIST benchmark. For a model, we list the number of capsules per layer (\#caps), the dimension of the capsules (dim), and the number of routing layers (depth).
		The number of backbone parameters and the sum of all routing layer parameters are listed separately in \textbf{10k}. We give the validation accuracy for the model when trained with uniform routing and RBA.}
	\label{tab:affnist:grid_run:best_overall_models}
\end{table}

\begin{table}[htb!]
	\centering
	\begin{tabular}{ccc|cc|c}
		\toprule
		\multicolumn{3}{c}{Model Settings} &
		\multicolumn{2}{c}{Parameters} &
		\multirow{2}{*}{AffNIST Acc.} \\
		depth  & \#caps & dim  & Routing  & Backbone &  \\
		\midrule
		1 & 16 &  8 & 2 & 16 & 0.88 \\
		2 & 64 &  8 & 35 & 73 & 0.88 \\
		3 & 32 & 64 & 872 & 1006 & 0.86 \\
		4 & 16 & 64 & 331 & 401 & 0.85 \\
		5 & 32 & 16 & 113 & 151 & 0.84 \\
		6 & 16 & 32 & 139 & 177 & 0.83 \\
		\bottomrule
	\end{tabular}
	\caption{Overview of the best models per depth on the AffNIST benchmark. For a model, we list the number of capsules per layer (\#caps), the dimension of the capsules (dim), and the number of routing layers (depth).
		The number of backbone parameters and the sum of all routing layer parameters are listed separately in \textbf{10k}.}
	\label{tab:affnist:grid_run:best_models_depth}
\end{table}

\begin{table}%
	\centering
	\subfloat[]{
	\begin{tabular}{ccc|cc|cc}
	\toprule
	\midrule
	\multirow{2}{*}{Capsule Layer} &
	\multicolumn{2}{c}{Capsule Norms} &
	\multicolumn{2}{c}{Capsule Activation} &
	\multicolumn{2}{c}{Capsule Deaths} \\
	& Mean ($\cnm$) & Sum ($\cns$) & Rate ($\car$) & Sum ($\cas$) & Rate ($\cdr$) & Sum ($\cds$) \\
	\midrule
	1 & 0.90  & 14.40 & 1.00  & 16.00 & 0.00  & 0.00\\
	2 & 0.18  & 1.79 & 0.42  & 4.23 & 0.00  & 0.00\\
	\bottomrule
	\end{tabular}
	}%
	\vspace{0.5cm}
	\subfloat[]{
	\begin{tabular}{ccc|cc}
	\toprule
	\multirow{2}{*}{Routing Layer} &
	\multicolumn{2}{c}{Capsules Alive} &
	\multicolumn{2}{c}{Routing Dynamics} \\
	& From lower layer & To higher layer & Rate ($\dyr$) & Mean ($\dys$) \\
	\midrule
	1 & 16 & 10 & 0.31 & 3.05 \\
	\midrule
	\bottomrule
	\end{tabular}
	}
	\caption{Capsule activation and routing dynamics for the best model with one routing layer.}%
	\label{tab:affnist:model_d1:best}%
\end{table}

\begin{table}%
	\centering
	\subfloat[]{
	\begin{tabular}{ccc|cc|cc}
	\toprule
	\midrule
	\multirow{2}{*}{Capsule Layer} &
	\multicolumn{2}{c}{Capsule Norms} &
	\multicolumn{2}{c}{Capsule Activation} &
	\multicolumn{2}{c}{Capsule Deaths} \\
	& Mean ($\cnm$) & Sum ($\cns$) & Rate ($\car$) & Sum ($\cas$) & Rate ($\cdr$) & Sum ($\cds$) \\
	\midrule
	1 & 0.93  & 59.73 & 1.00  & 64.00 & 0.00  & 0.00\\
	2 & 0.08  & 5.35 & 0.14  & 8.68 & 0.75  & 48.00\\
	3 & 0.13  & 1.31 & 0.22  & 2.16 & 0.00  & 0.00\\
	\bottomrule
\end{tabular}
	}%
	\vspace{0.5cm}
	\subfloat[]{
	\begin{tabular}{ccc|cc}
	\toprule
	\multirow{2}{*}{Routing Layer} &
	\multicolumn{2}{c}{Capsules Alive} &
	\multicolumn{2}{c}{Routing Dynamics} \\
	& From lower layer & To higher layer & Rate ($\dyr$) & Mean ($\dys$) \\
	\midrule
	1 & 64 & 16 & 0.14 & 2.27 \\
	2 & 16 & 10 & 0.26 & 2.63 \\
	\midrule
	\bottomrule
\end{tabular}
	}
	\caption{Capsule activation and routing dynamics for the best model with two routing layers.}%
	\label{tab:affnist:model_d2:best}%
\end{table}
\begin{table}%
	\centering
	\subfloat[]{
	\begin{tabular}{ccc|cc|cc}
	\toprule
	\midrule
	\multirow{2}{*}{Capsule Layer} &
	\multicolumn{2}{c}{Capsule Norms} &
	\multicolumn{2}{c}{Capsule Activation} &
	\multicolumn{2}{c}{Capsule Deaths} \\
	& Mean ($\cnm$) & Sum ($\cns$) & Rate ($\car$) & Sum ($\cas$) & Rate ($\cdr$) & Sum ($\cds$) \\
	\midrule
	1 & 0.99  & 31.76 & 1.00  & 32.00 & 0.00  & 0.00\\
	2 & 0.21  & 6.71 & 0.37  & 11.81 & 0.22  & 7.00\\
	3 & 0.07  & 2.18 & 0.09  & 3.02 & 0.72  & 23.00\\
	4 & 0.13  & 1.25 & 0.16  & 1.58 & 0.00  & 0.00\\
	\bottomrule
\end{tabular}
	}%
	\vspace{0.5cm}
	\subfloat[]{
	\begin{tabular}{ccc|cc}
	\toprule
	\multirow{2}{*}{Routing Layer} &
	\multicolumn{2}{c}{Capsules Alive} &
	\multicolumn{2}{c}{Routing Dynamics} \\
	& From lower layer & To higher layer & Rate ($\dyr$) & Mean ($\dys$) \\
	\midrule
	1 & 32 & 25 & 0.41 & 10.18 \\
	2 & 25 & 9 & 0.20 & 1.80 \\
	3 & 9 & 10 & 0.38 & 3.80 \\
	\midrule
	\bottomrule
\end{tabular}
	}
	\caption{Capsule activation and routing dynamics for the best model with three routing layers.}%
	\label{tab:affnist:model_d3:best}%
\end{table}

\begin{table}%
	\centering
	\subfloat[]{
	\begin{tabular}{ccc|cc|cc}
	\toprule
	\midrule
	\multirow{2}{*}{Capsule Layer} &
	\multicolumn{2}{c}{Capsule Norms} &
	\multicolumn{2}{c}{Capsule Activation} &
	\multicolumn{2}{c}{Capsule Deaths} \\
	& Mean ($\cnm$) & Sum ($\cns$) & Rate ($\car$) & Sum ($\cas$) & Rate ($\cdr$) & Sum ($\cds$) \\
	\midrule
	1 & 0.99  & 15.92 & 1.00  & 16.00 & 0.00  & 0.00\\
	2 & 0.41  & 6.60 & 0.69  & 10.98 & 0.00  & 0.00\\
	3 & 0.18  & 2.83 & 0.25  & 3.96 & 0.50  & 8.00\\
	4 & 0.10  & 1.52 & 0.14  & 2.24 & 0.56  & 9.00\\
	5 & 0.12  & 1.23 & 0.17  & 1.71 & 0.00  & 0.00\\
	\bottomrule
\end{tabular}
	}%
	\vspace{0.5cm}
	\subfloat[]{
	\begin{tabular}{ccc|cc}
	\toprule
	\multirow{2}{*}{Routing Layer} &
	\multicolumn{2}{c}{Capsules Alive} &
	\multicolumn{2}{c}{Routing Dynamics} \\
	& From lower layer & To higher layer & Rate ($\dyr$) & Mean ($\dys$) \\
	\midrule
	1 & 16 & 16 & 0.49 & 7.78 \\
	2 & 16 & 8 & 0.31 & 2.51 \\
	3 & 8 & 7 & 0.34 & 2.41 \\
	4 & 7 & 10 & 0.37 & 3.66 \\
	\midrule
	\bottomrule
\end{tabular}
	}
	\caption{Capsule activation and routing dynamics for the best model with four routing layers.}%
	\label{tab:affnist:model_d4:best}%
\end{table}

\begin{table}%
	\centering
	\subfloat[]{
	\begin{tabular}{ccc|cc|cc}
	\toprule
	\midrule
	\multirow{2}{*}{Capsule Layer} &
	\multicolumn{2}{c}{Capsule Norms} &
	\multicolumn{2}{c}{Capsule Activation} &
	\multicolumn{2}{c}{Capsule Deaths} \\
	& Mean ($\cnm$) & Sum ($\cns$) & Rate ($\car$) & Sum ($\cas$) & Rate ($\cdr$) & Sum ($\cds$) \\
	\midrule
	1 & 0.96  & 30.83 & 1.00  & 32.00 & 0.00  & 0.00\\
	2 & 0.17  & 5.29 & 0.27  & 8.79 & 0.34  & 11.00\\
	3 & 0.07  & 2.14 & 0.08  & 2.56 & 0.72  & 23.00\\
	4 & 0.05  & 1.50 & 0.06  & 2.05 & 0.91  & 29.00\\
	5 & 0.05  & 1.45 & 0.08  & 2.55 & 0.75  & 24.00\\
	6 & 0.14  & 1.39 & 0.19  & 1.89 & 0.00  & 0.00\\
	\bottomrule
\end{tabular}
	}%
	\vspace{0.5cm}
	\subfloat[]{
	\begin{tabular}{ccc|cc}
	\toprule
	\multirow{2}{*}{Routing Layer} &
	\multicolumn{2}{c}{Capsules Alive} &
	\multicolumn{2}{c}{Routing Dynamics} \\
	& From lower layer & To higher layer & Rate ($\dyr$) & Mean ($\dys$) \\
	\midrule
	1 & 32 & 21 & 0.28 & 5.81 \\
	2 & 21 & 9 & 0.20 & 1.82 \\
	3 & 9 & 3 & 0.19 & 0.57 \\
	4 & 3 & 8 & 0.20 & 1.63 \\
	5 & 8 & 10 & 0.31 & 3.10 \\
	\midrule
	\bottomrule
\end{tabular}
	}
	\caption{Capsule activation and routing dynamics for the best model with five routing layers.}%
	\label{tab:affnist:model_d5:best}%
\end{table}

\begin{table}%
	\centering
	\subfloat[]{
	\begin{tabular}{ccc|cc|cc}
	\toprule
	\midrule
	\multirow{2}{*}{Capsule Layer} &
	\multicolumn{2}{c}{Capsule Norms} &
	\multicolumn{2}{c}{Capsule Activation} &
	\multicolumn{2}{c}{Capsule Deaths} \\
	& Mean ($\cnm$) & Sum ($\cns$) & Rate ($\car$) & Sum ($\cas$) & Rate ($\cdr$) & Sum ($\cds$) \\
	\midrule
	1 & 0.98  & 15.75 & 1.00  & 16.00 & 0.00  & 0.00\\
	2 & 0.27  & 4.36 & 0.45  & 7.13 & 0.19  & 3.00\\
	3 & 0.14  & 2.18 & 0.18  & 2.89 & 0.62  & 10.00\\
	4 & 0.09  & 1.50 & 0.12  & 1.92 & 0.69  & 11.00\\
	5 & 0.09  & 1.41 & 0.13  & 2.08 & 0.56  & 9.00\\
	6 & 0.09  & 1.42 & 0.15  & 2.34 & 0.56  & 9.00\\
	7 & 0.14  & 1.41 & 0.24  & 2.40 & 0.00  & 0.00\\
	\bottomrule
\end{tabular}
	}%
	\vspace{0.5cm}
	\subfloat[]{
	\begin{tabular}{ccc|cc}
	\toprule
	\multirow{2}{*}{Routing Layer} &
	\multicolumn{2}{c}{Capsules Alive} &
	\multicolumn{2}{c}{Routing Dynamics} \\
	& From lower layer & To higher layer & Rate ($\dyr$) & Mean ($\dys$) \\
	\midrule
	1 & 16 & 13 & 0.40 & 5.26 \\
	2 & 13 & 6 & 0.23 & 1.39 \\
	3 & 6 & 5 & 0.25 & 1.26 \\
	4 & 5 & 7 & 0.26 & 1.80 \\
	5 & 7 & 7 & 0.31 & 2.17 \\
	6 & 7 & 10 & 0.35 & 3.46 \\
	\midrule
	\bottomrule
\end{tabular}
	}
	\caption{Capsule activation and routing dynamics for the best model with six routing layers.}%
	\label{tab:affnist:model_d6:best}%
\end{table}

\clearpage
\section{CIFAR10: Complete Results for a Single Model}\label{app:cifar10_single}
\begin{figure}[!ht]%
	\centering
	\subfloat[Mean Couplings]
	{\includegraphics[width=0.22\linewidth]{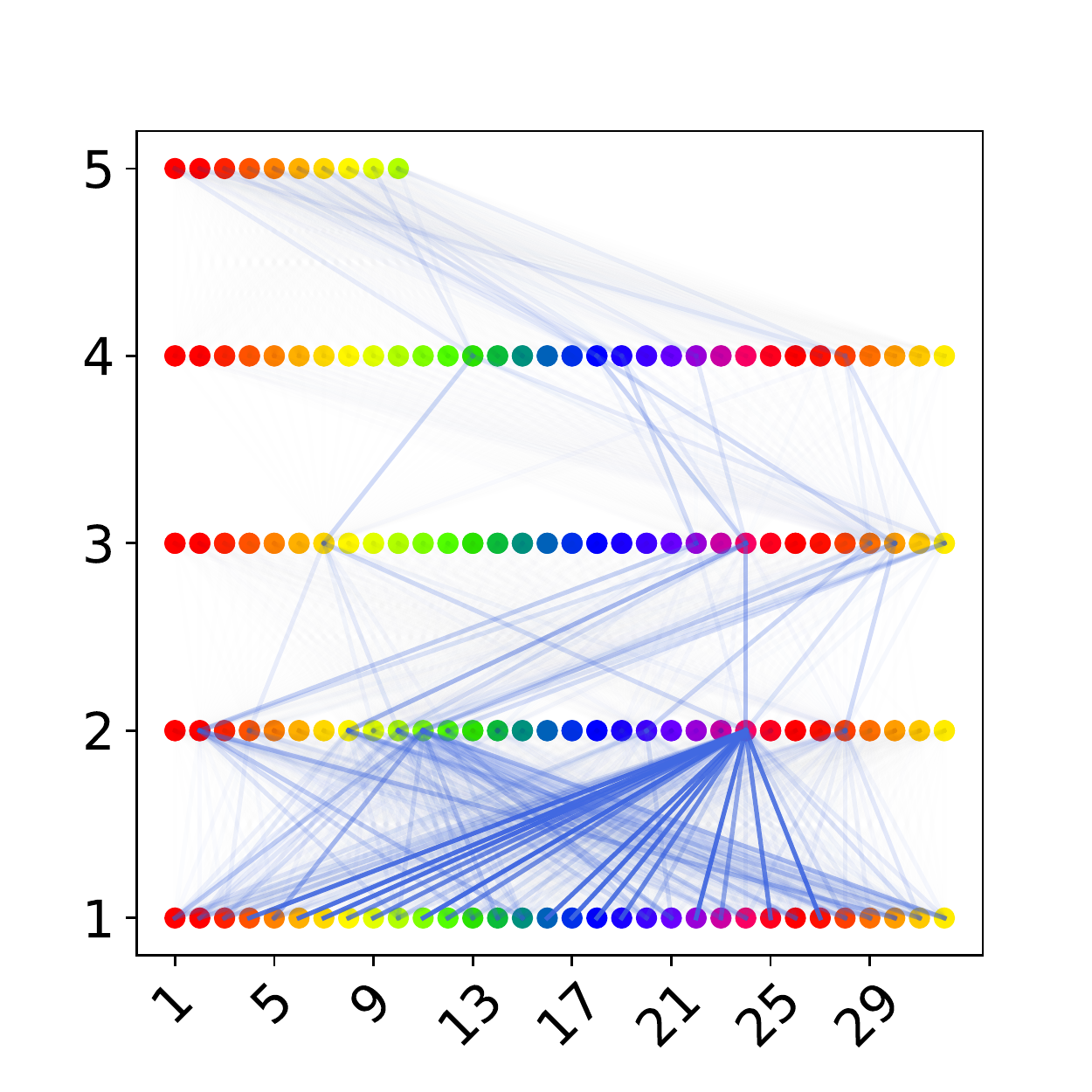}}
	\subfloat[Std Couplings]
	{\includegraphics[width=0.22\linewidth]{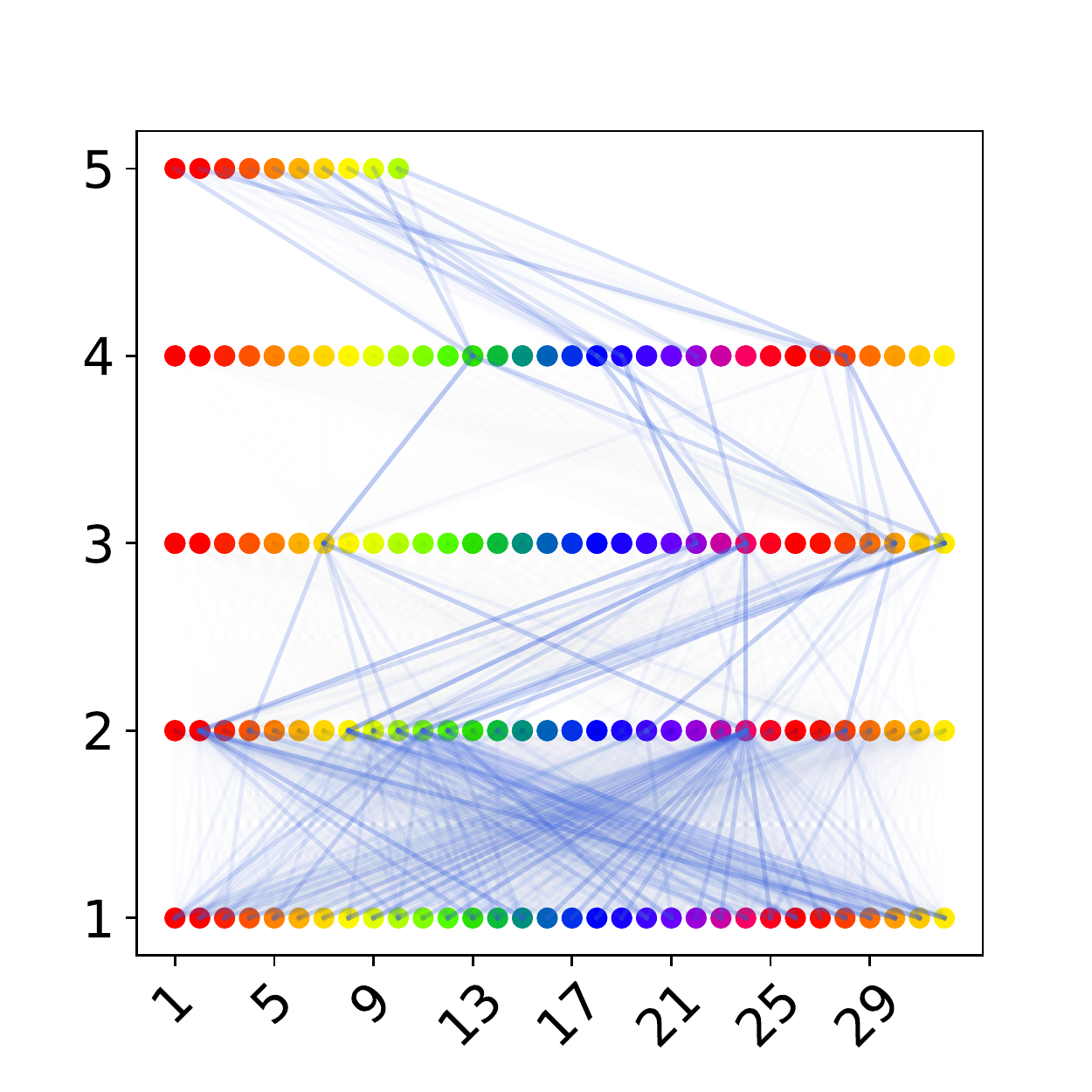}}
	\subfloat[Mean Activations]
	{\includegraphics[width=0.22\linewidth]{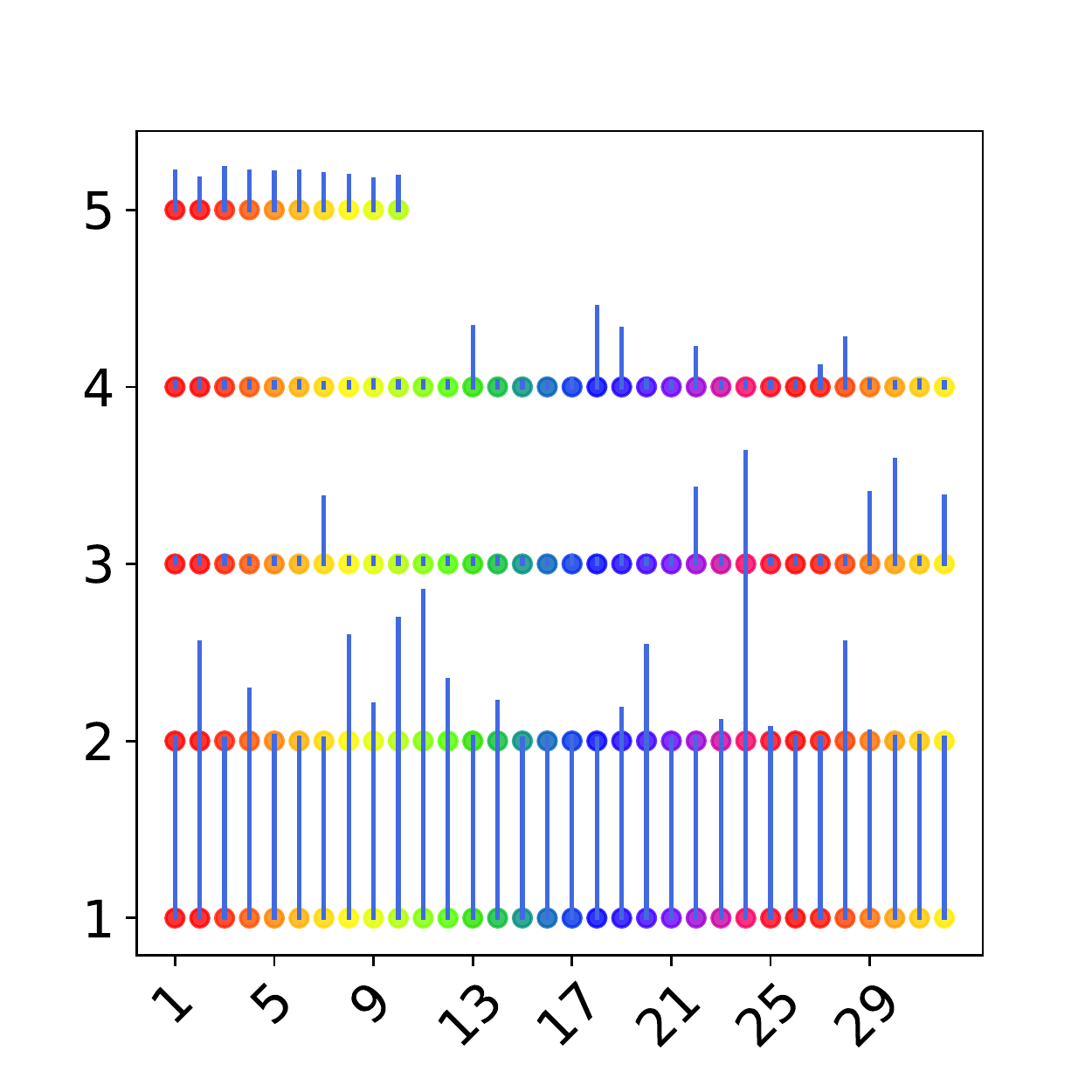}}
	\subfloat[Std Activations]
	{\includegraphics[width=0.22\linewidth]{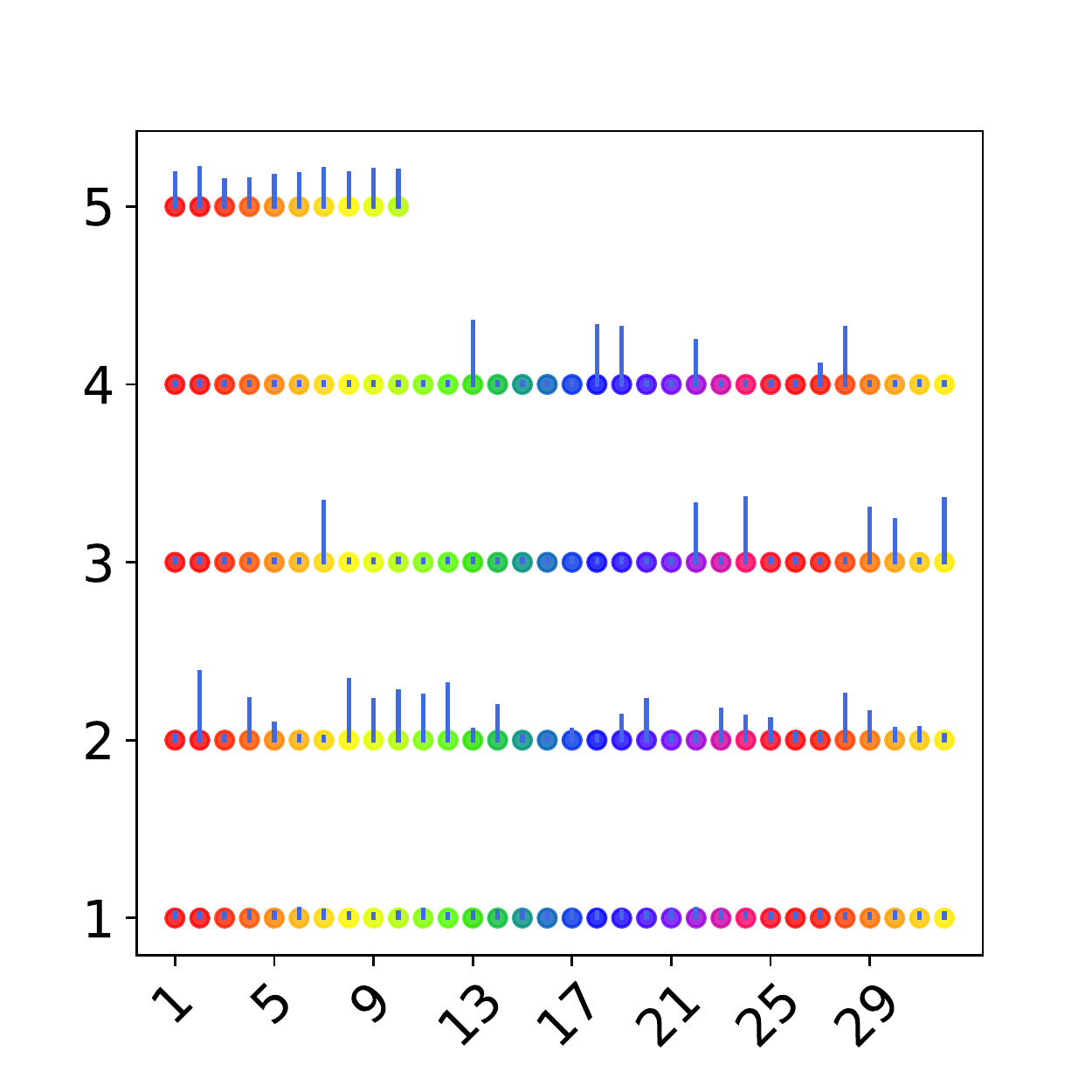}}
	\subfloat[Dead Capsules]
	{\includegraphics[width=0.22\linewidth]{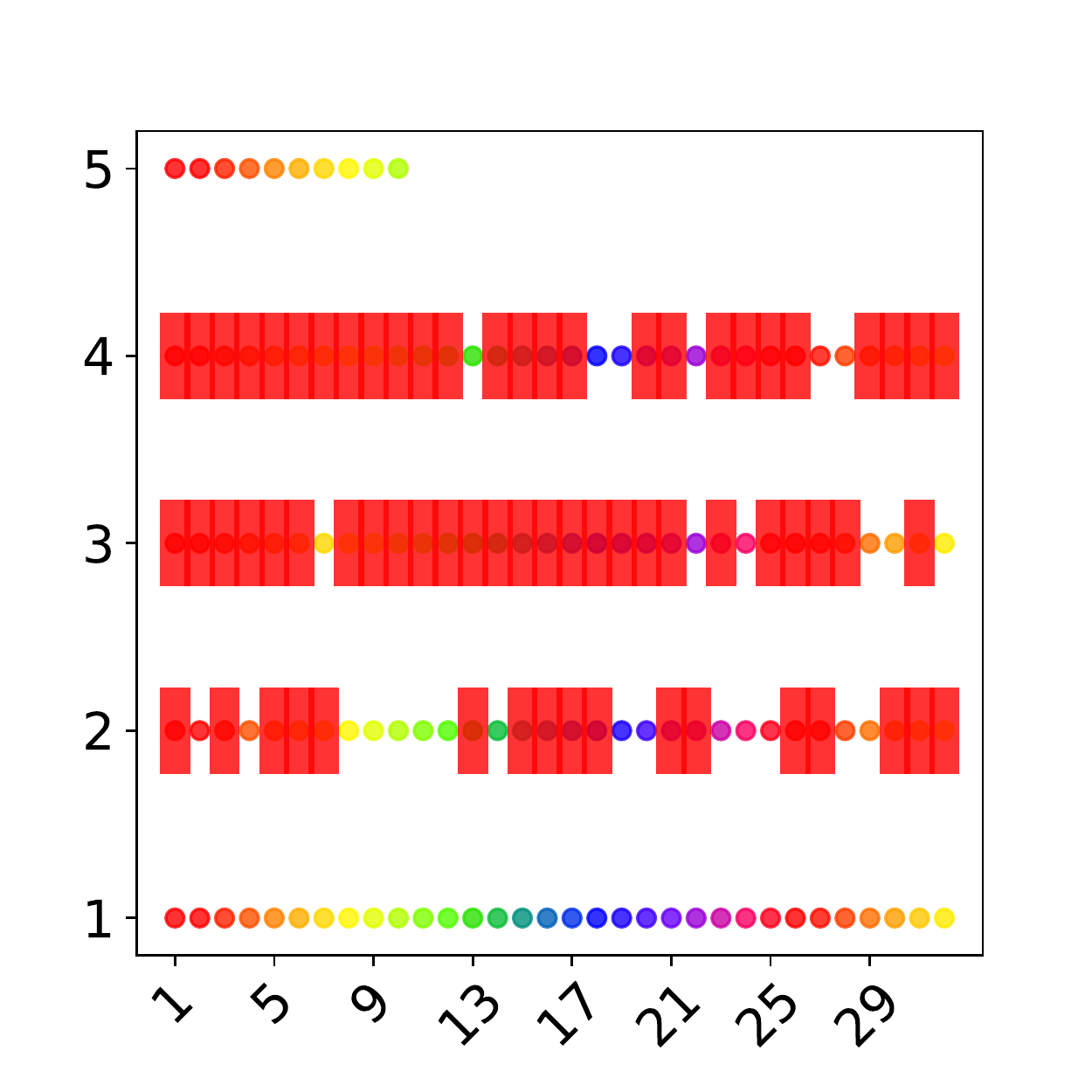}}
	\caption{Parse-tree statistics for the complete CIFAR10 validation dataset for a five-layer CapsNet model with 32/10 capsules. The mean~(a) and the standard deviation~(b) of the coupling coefficient matrices for each layer are visualized as connections between capsules. Higher coupling coefficients have a darker blue tone. The capsule norms' mean~(c) and standard deviation~(d)
		are visualized by bars. Dead capsules~(e) are highlighted with a red bar.}%
	\label{fig:cifar:main_model:training:parse_tree}
\end{figure}

\begin{table}[!ht]%
	\centering
	\subfloat[]{
	\begin{tabular}{ccc|cc|cc}
	\toprule
	\midrule
	\multirow{2}{*}{Capsule Layer} &
	\multicolumn{2}{c}{Capsule Norms} &
	\multicolumn{2}{c}{Capsule Activation} &
	\multicolumn{2}{c}{Capsule Deaths} \\
	& Mean ($\cnm$) & Sum ($\cns$) & Rate ($\car$) & Sum ($\cas$) & Rate ($\cdr$) & Sum ($\cds$) \\
	\midrule
	1 & 0.98  & 31.49 & 1.00  & 32.00 & 0.00  & 0.00\\
	2 & 0.20  & 6.55 & 0.35  & 11.30 & 0.53  & 17.00\\
	3 & 0.12  & 3.69 & 0.16  & 5.16 & 0.81  & 26.00\\
	4 & 0.08  & 2.52 & 0.10  & 3.36 & 0.81  & 26.00\\
	5 & 0.20  & 2.02 & 0.68  & 6.79 & 0.00  & 0.00\\
	\bottomrule
	\end{tabular}
	}%
	\vspace{0.5cm}
	\subfloat[]{
	\begin{tabular}{ccc|cc}
	\toprule
	\multirow{2}{*}{Routing Layer} &
	\multicolumn{2}{c}{Capsules Alive} &
	\multicolumn{2}{c}{Routing Dynamics} \\
	& From lower layer & To higher layer & Rate ($\dyr$) & Mean ($\dys$) \\
	\midrule
	1 & 32 & 15 & 0.19 & 2.90 \\
	2 & 15 & 6 & 0.23 & 1.37 \\
	3 & 6 & 6 & 0.27 & 1.61 \\
	4 & 6 & 10 & 0.26 & 2.60 \\
	\midrule
	\bottomrule
	\end{tabular}
	}
	\caption{Capsule activation and routing dynamics.}%
	\label{tab:cifar:main_model:activation_dynamics}%
\end{table}

\begin{figure}[!ht]
	\begin{center}
		\begin{tabular}{cccccc}
			\includegraphics[width=0.15\linewidth]{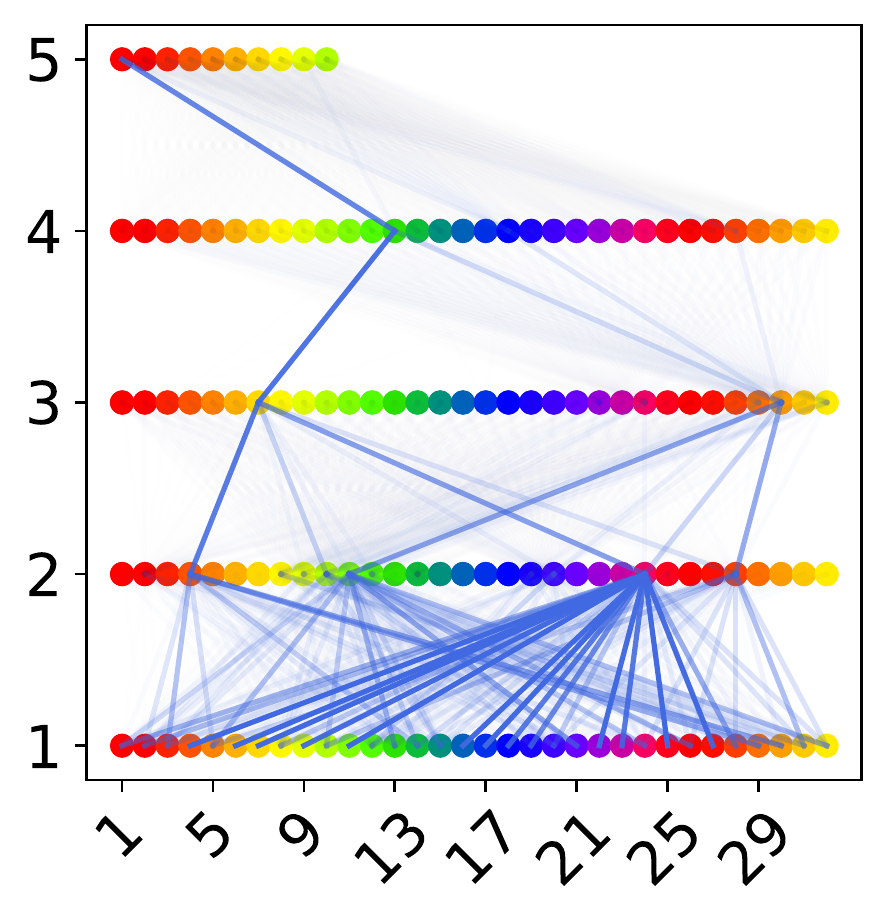} &
			\includegraphics[width=0.15\linewidth]{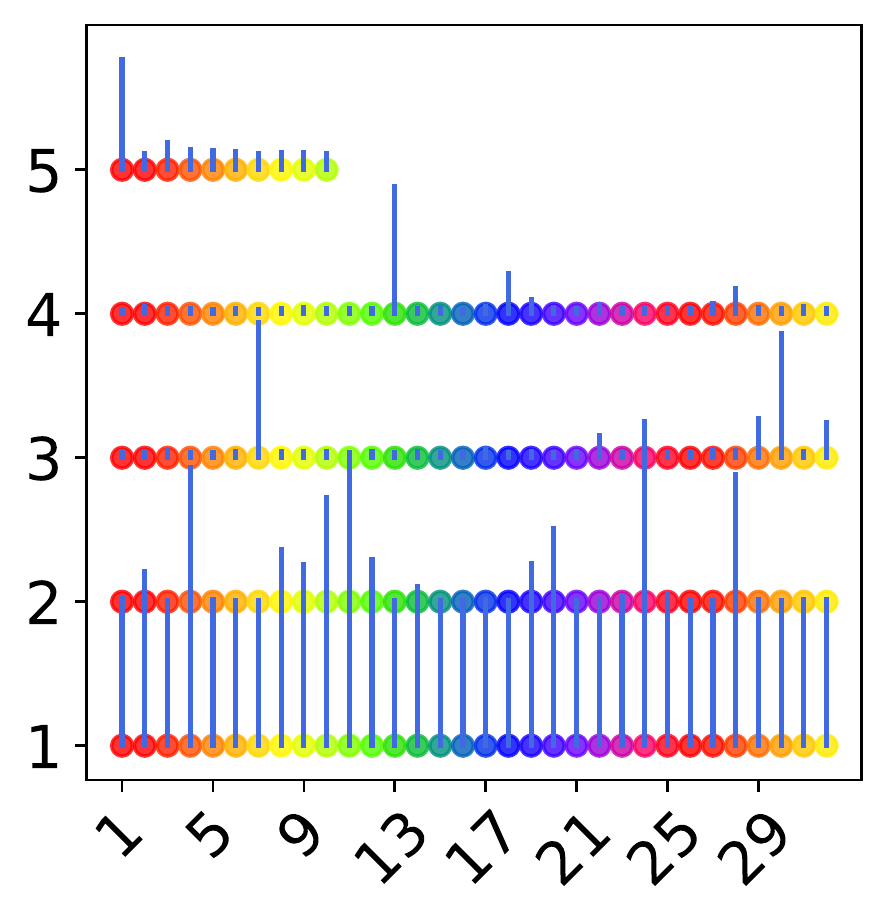} &
			\includegraphics[width=0.15\linewidth]{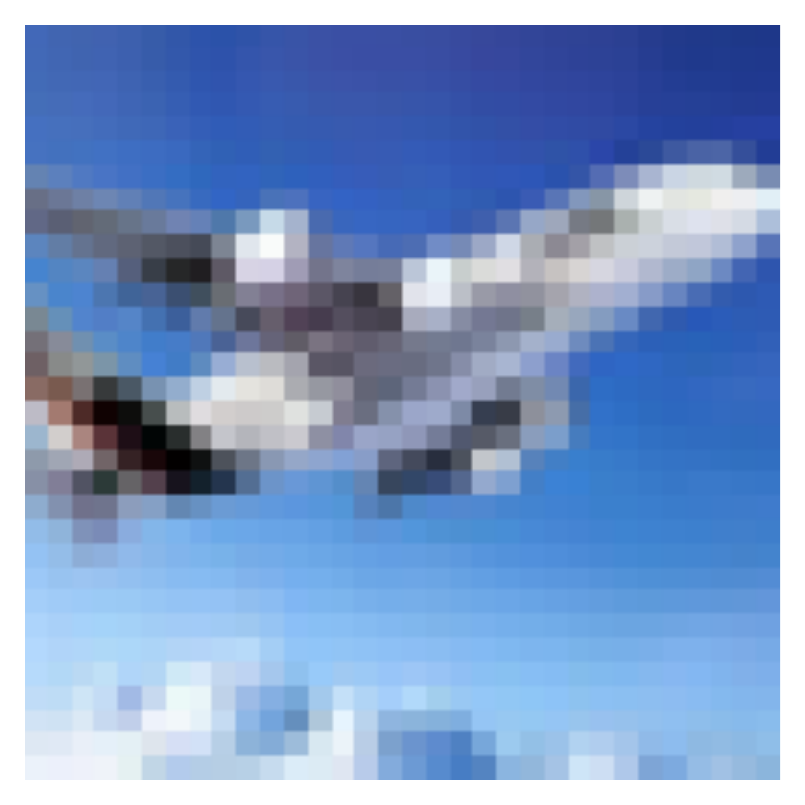} &
			\includegraphics[width=0.15\linewidth]{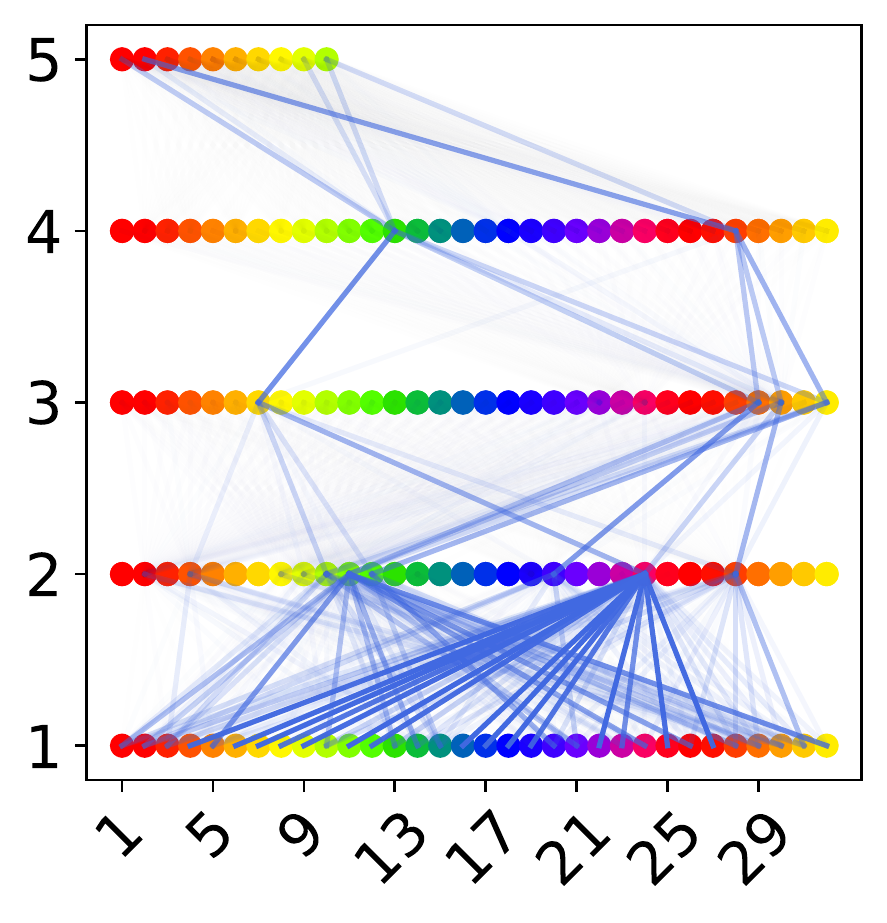} &
			\includegraphics[width=0.15\linewidth]{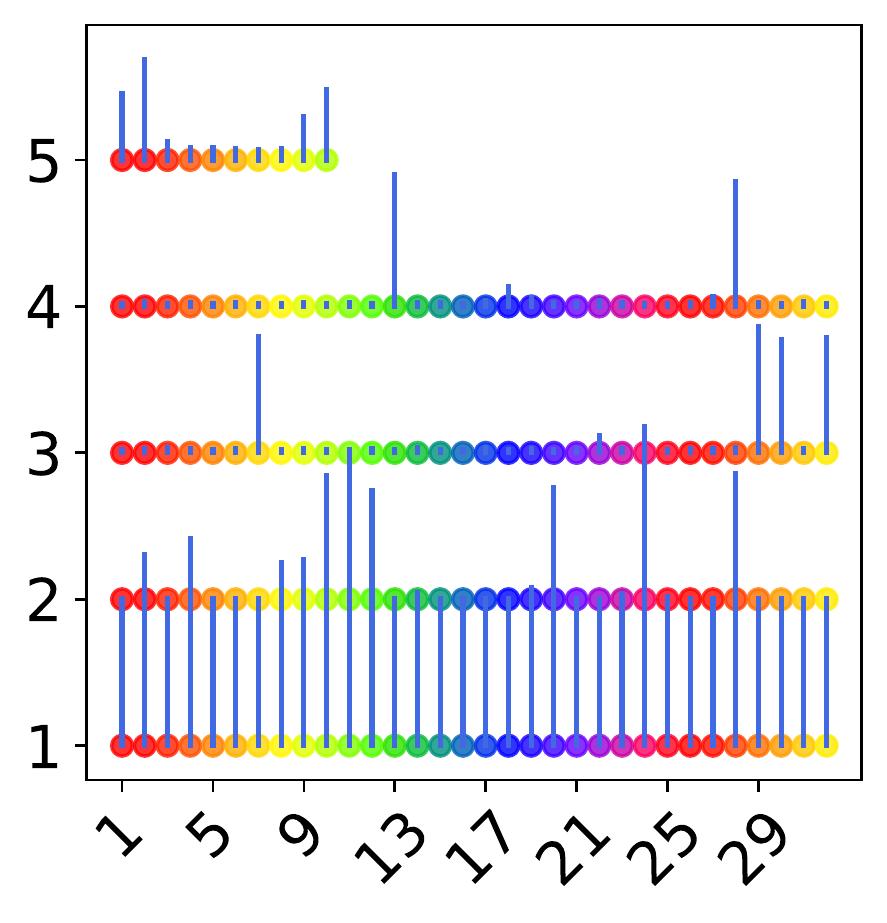} &
			\includegraphics[width=0.15\linewidth]{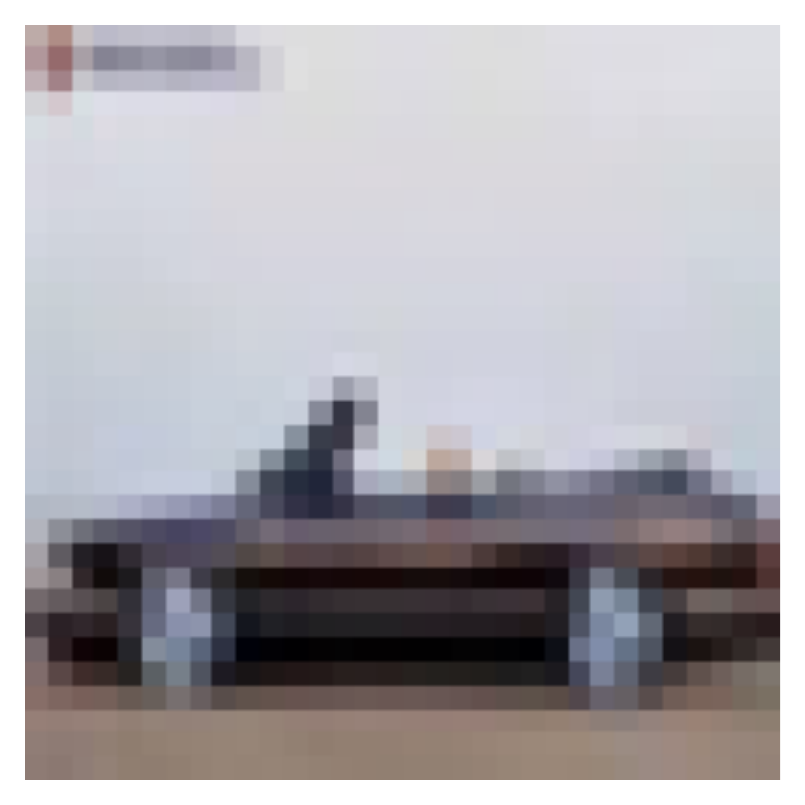} \\
			\includegraphics[width=0.15\linewidth]{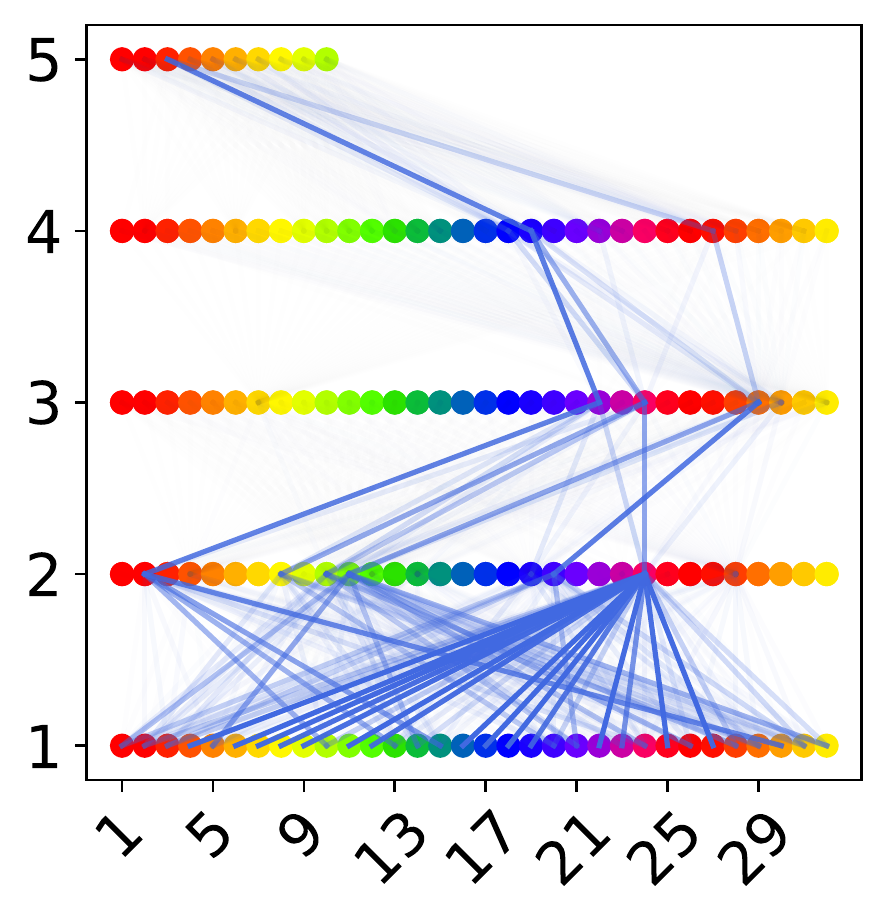} &
			\includegraphics[width=0.15\linewidth]{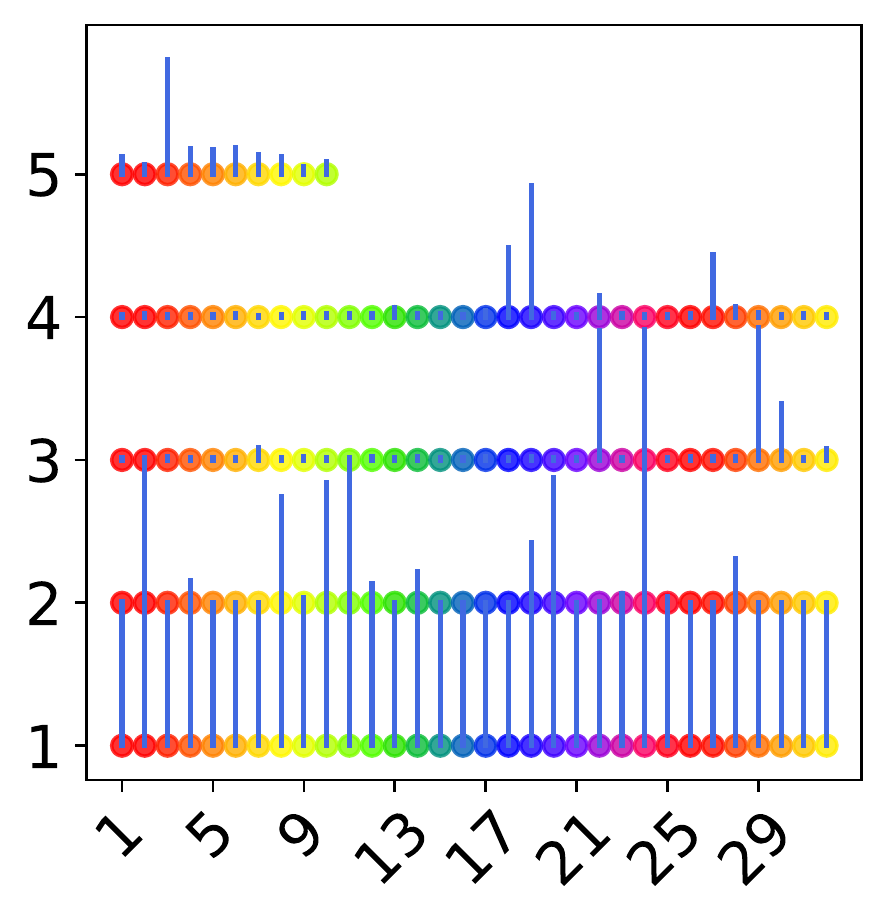} &
			\includegraphics[width=0.15\linewidth]{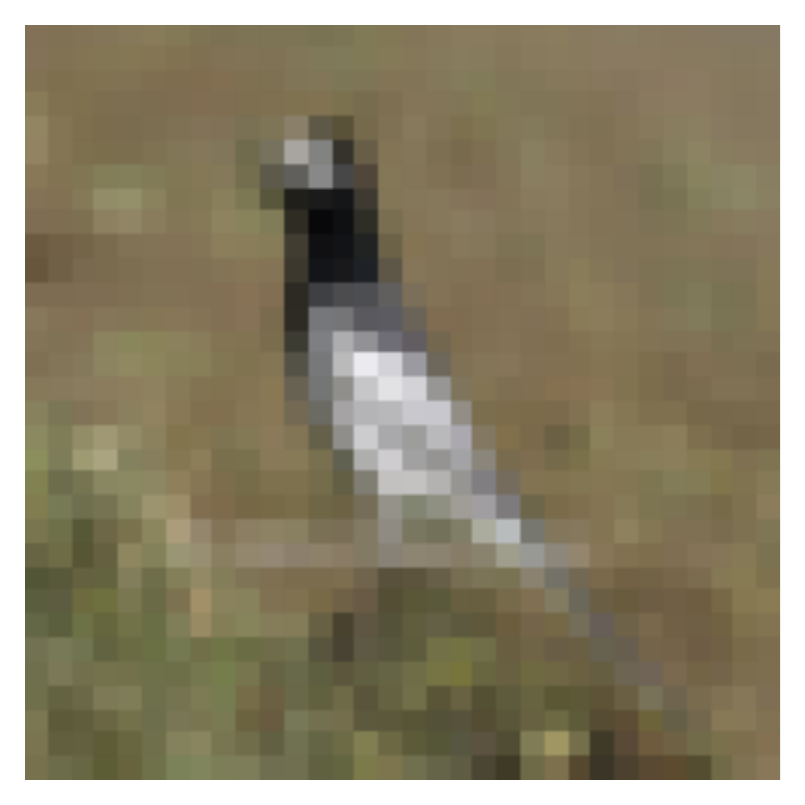} &
			\includegraphics[width=0.15\linewidth]{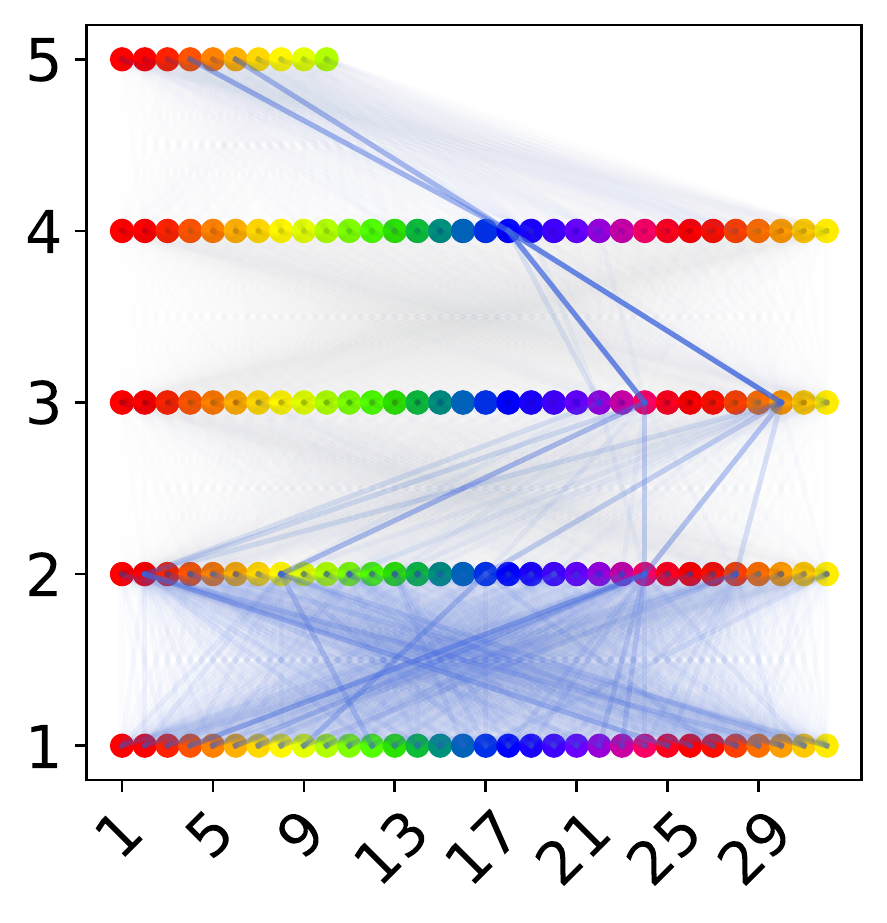} &
			\includegraphics[width=0.15\linewidth]{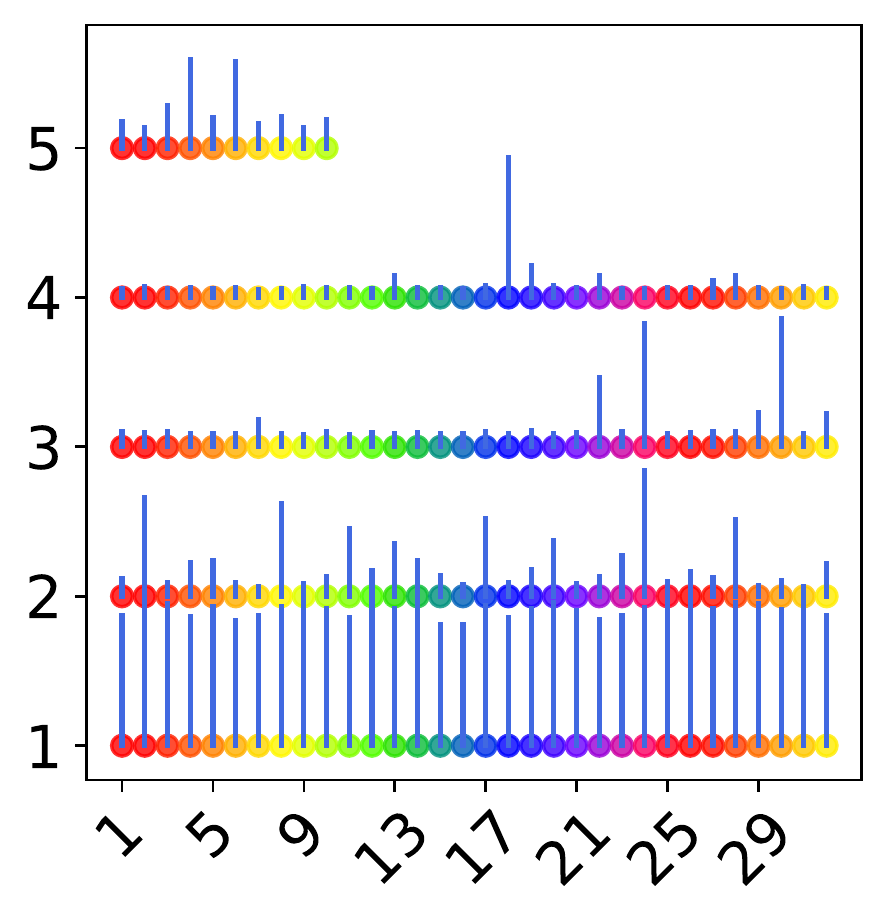} &
			\includegraphics[width=0.15\linewidth]{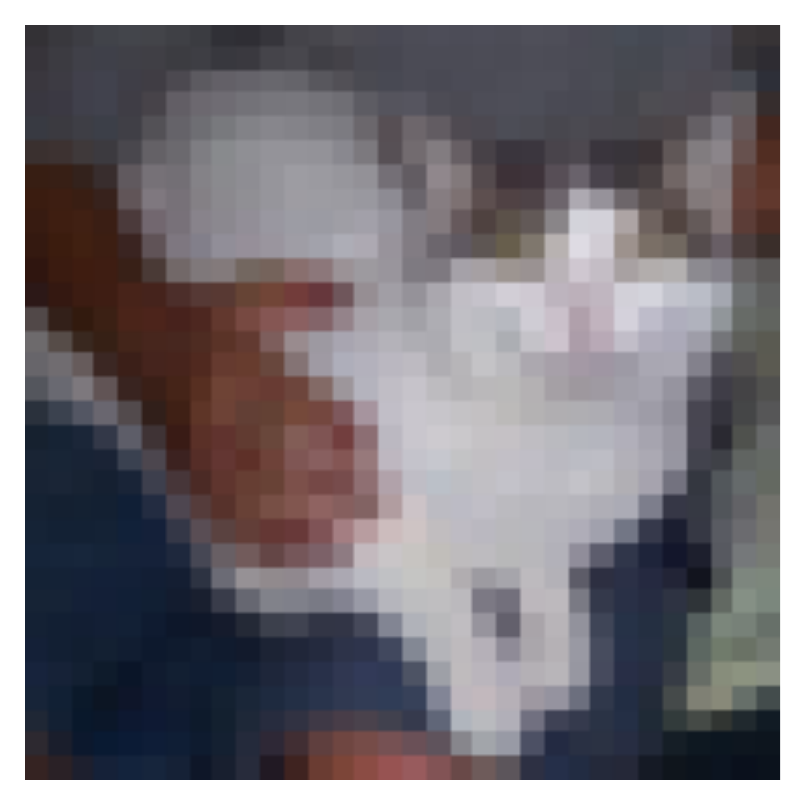} \\			
			\includegraphics[width=0.15\linewidth]{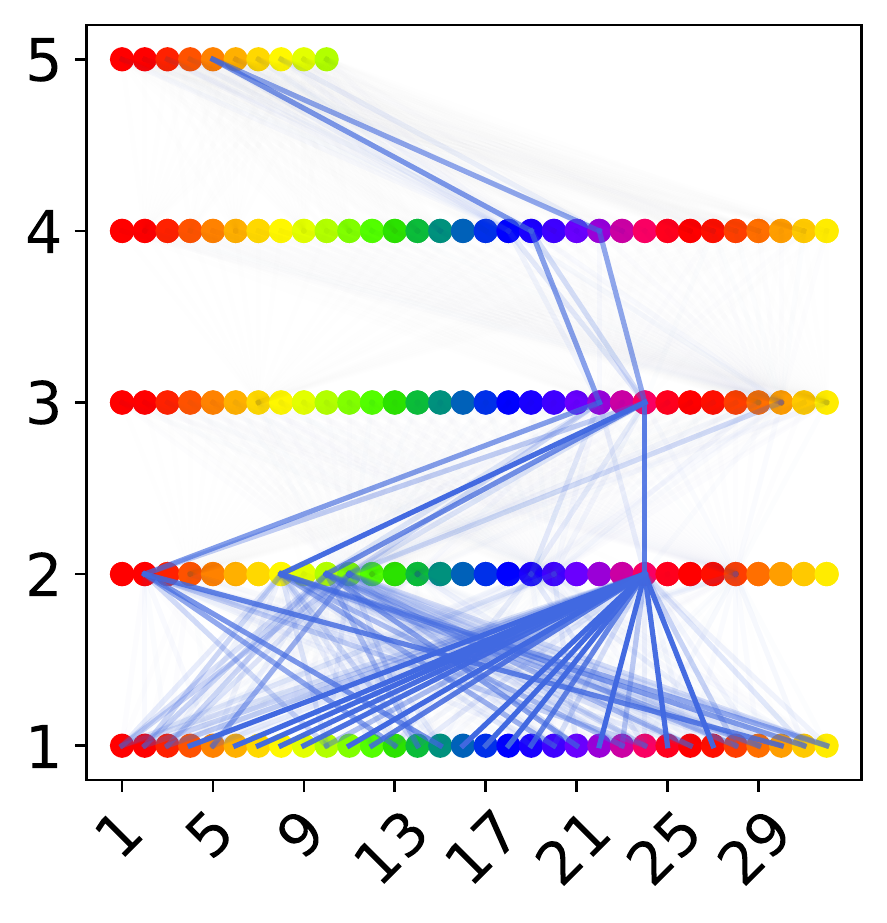} &
			\includegraphics[width=0.15\linewidth]{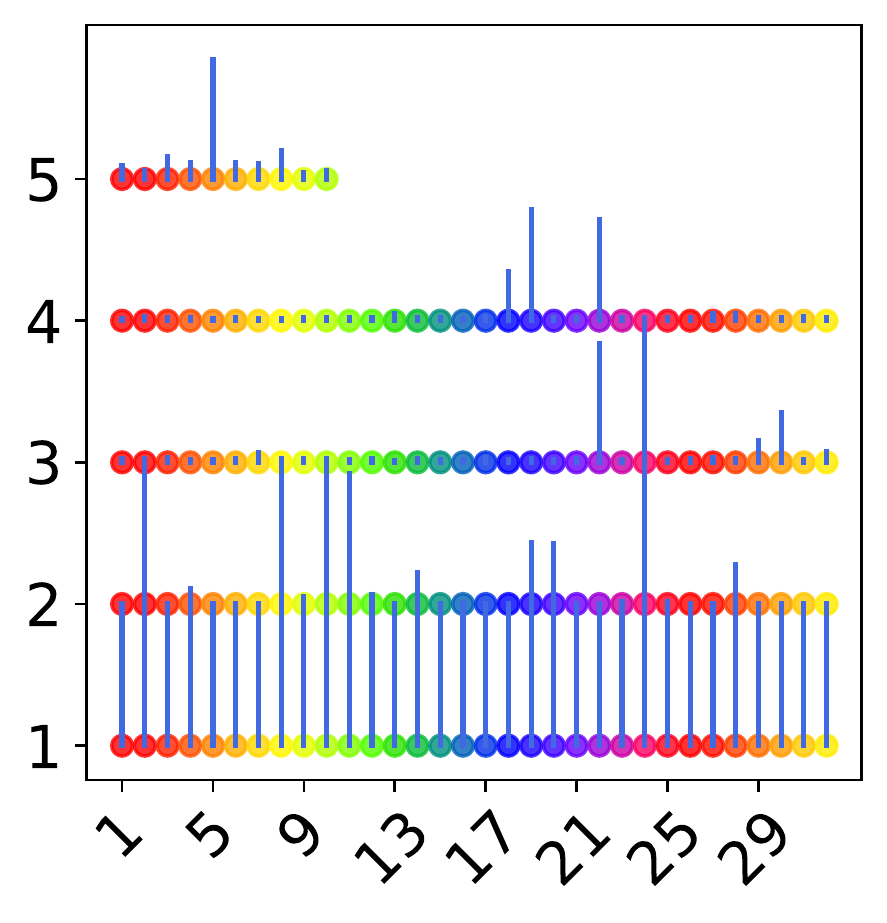} &
			\includegraphics[width=0.15\linewidth]{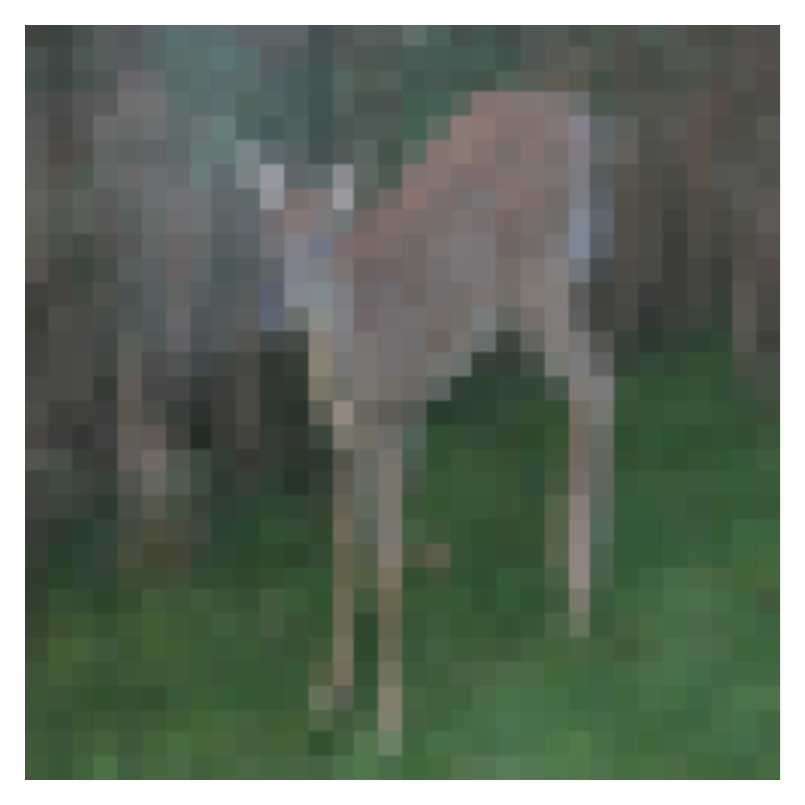} &
			\includegraphics[width=0.15\linewidth]{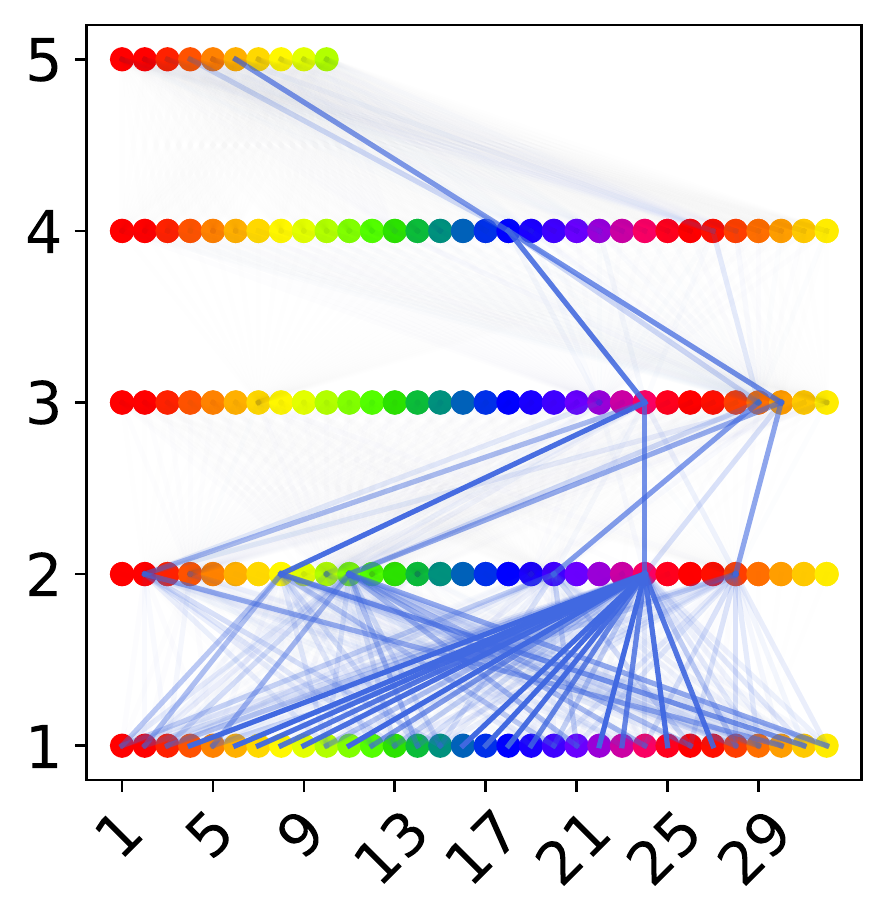} &
			\includegraphics[width=0.15\linewidth]{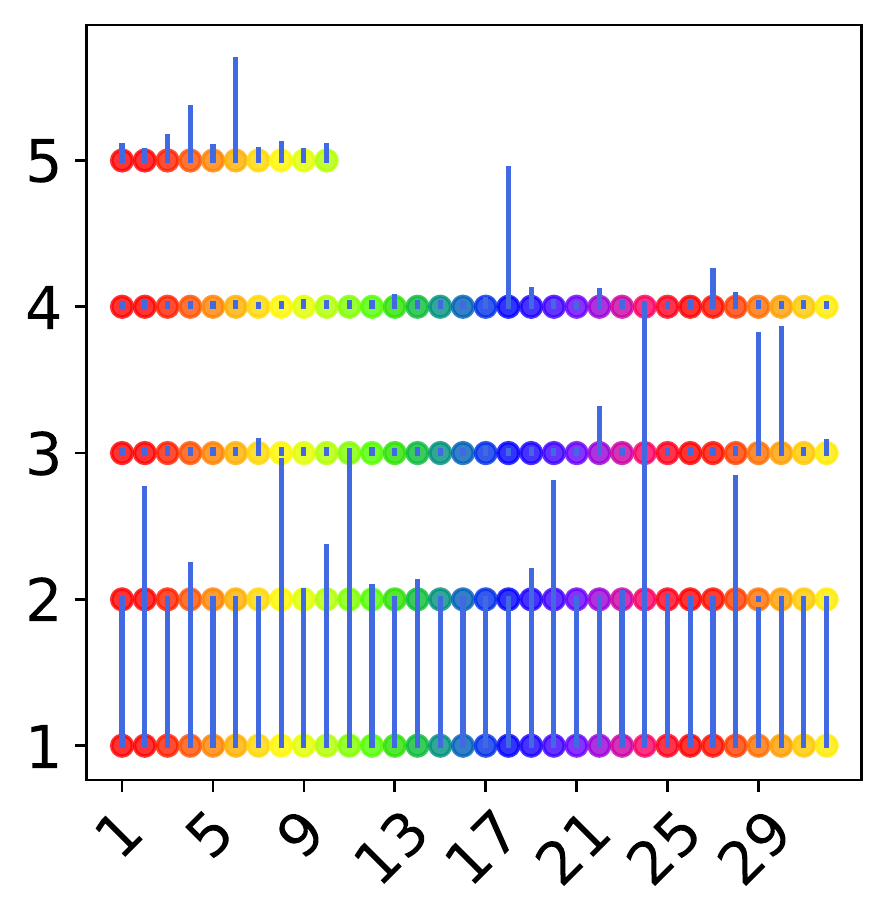} &
			\includegraphics[width=0.15\linewidth]{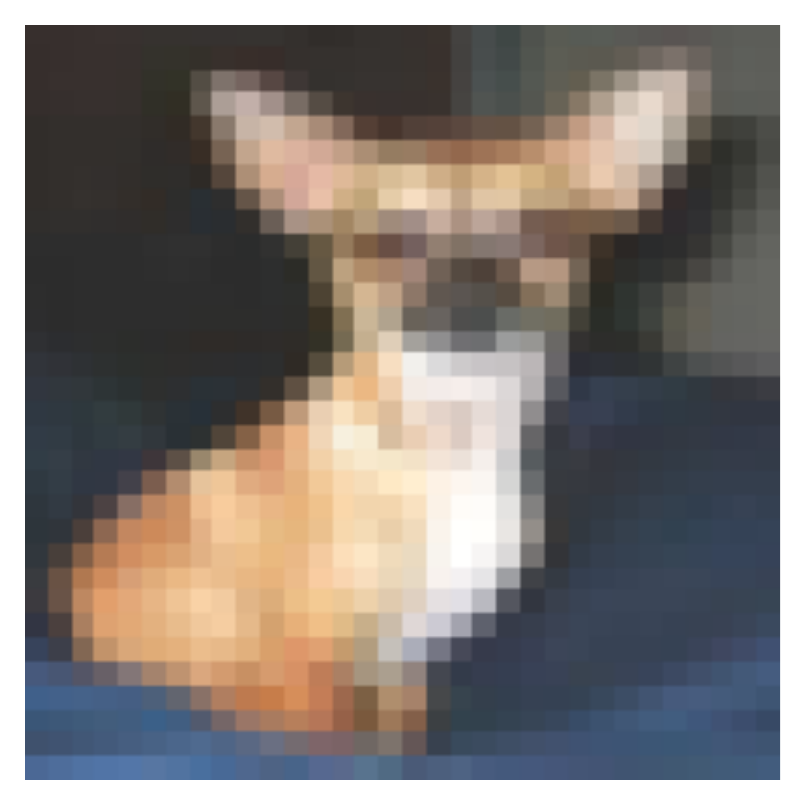} \\
			\includegraphics[width=0.15\linewidth]{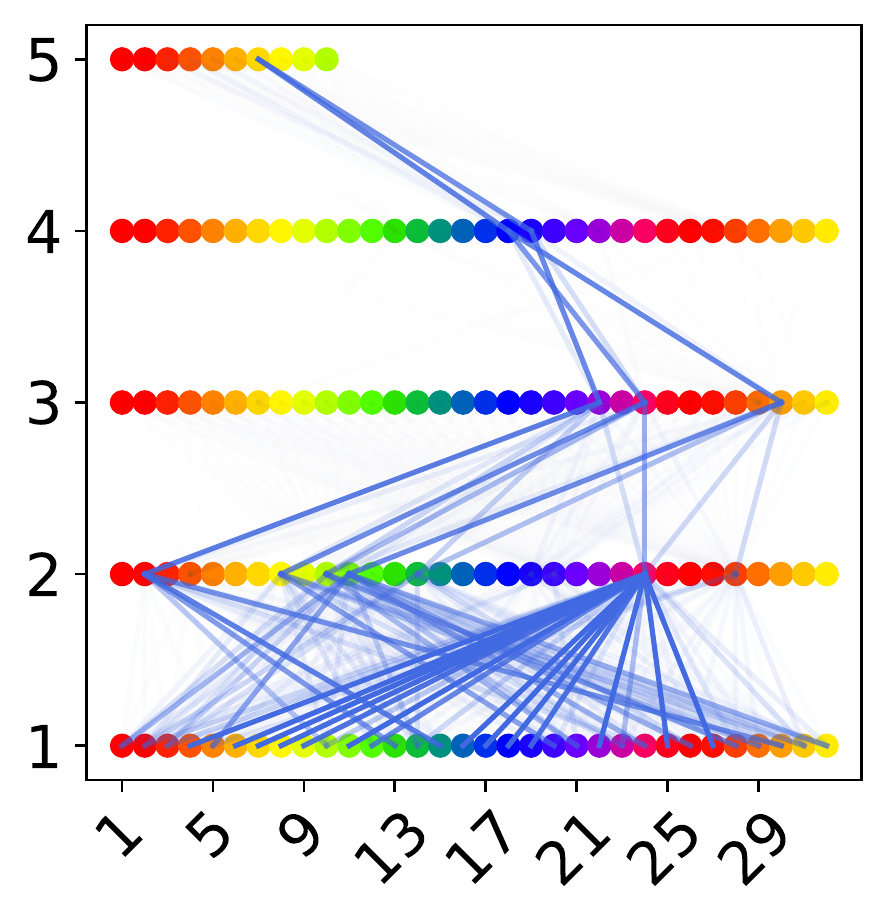} &
			\includegraphics[width=0.15\linewidth]{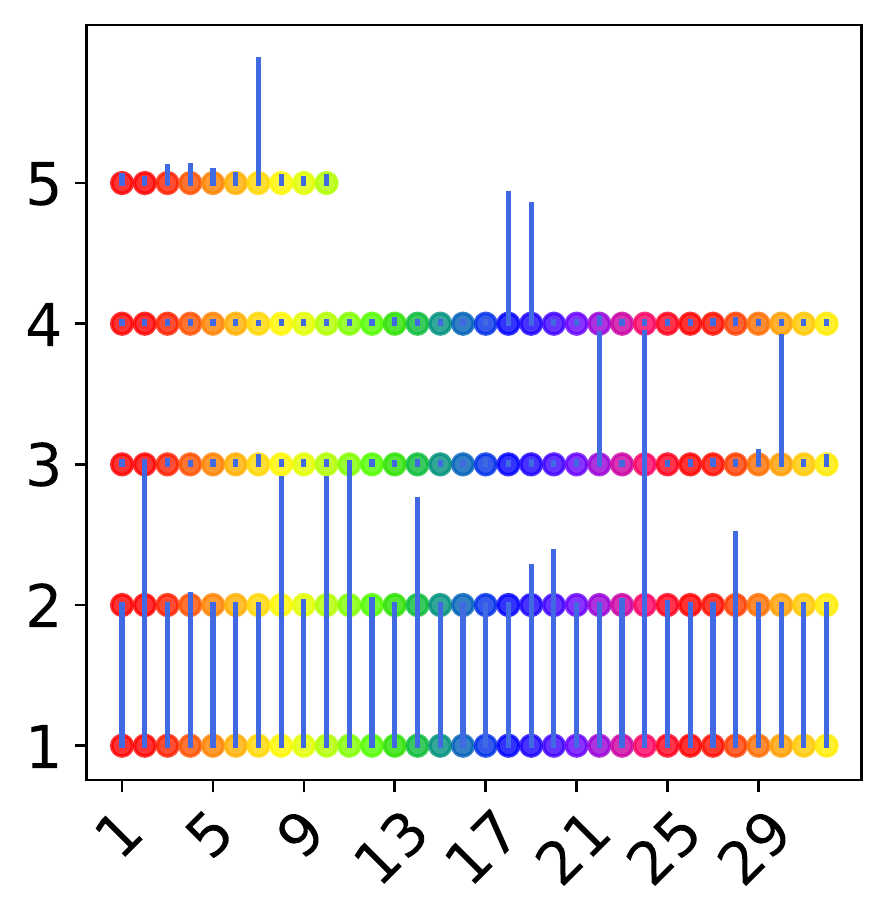} &
			\includegraphics[width=0.15\linewidth]{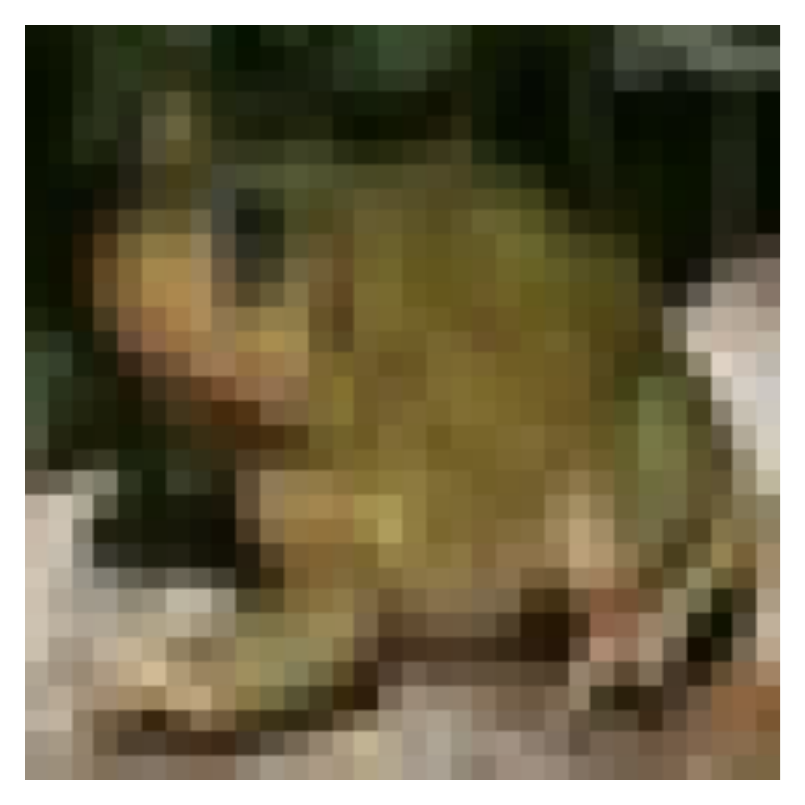} &
			\includegraphics[width=0.15\linewidth]{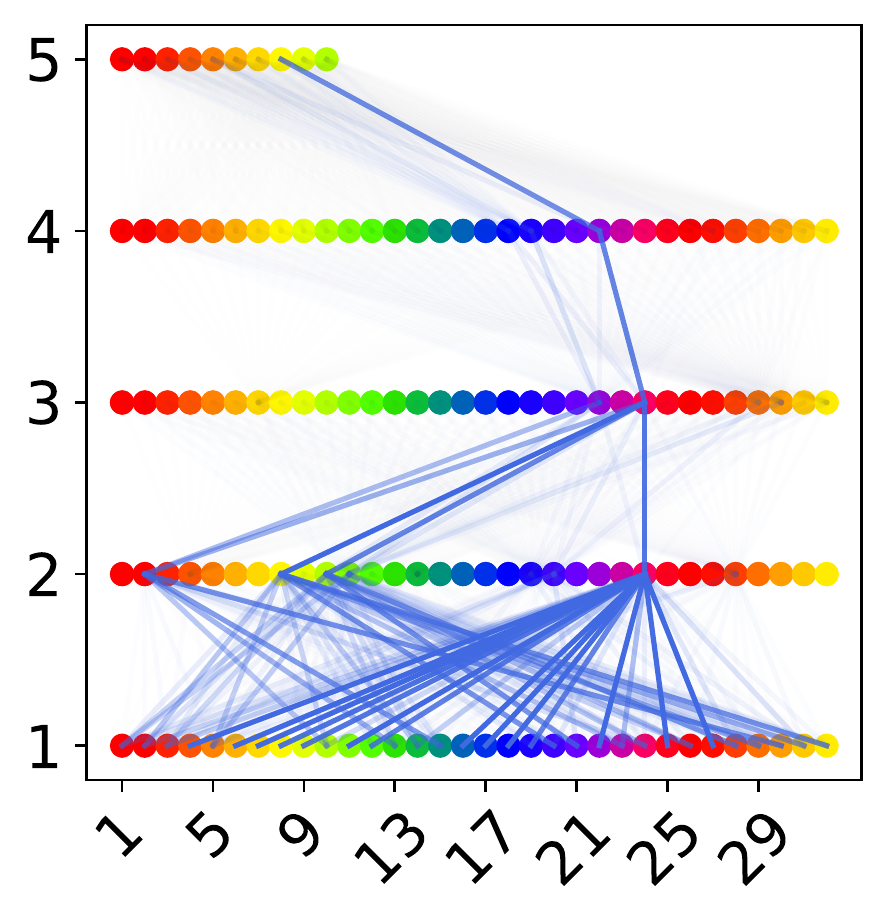} &
			\includegraphics[width=0.15\linewidth]{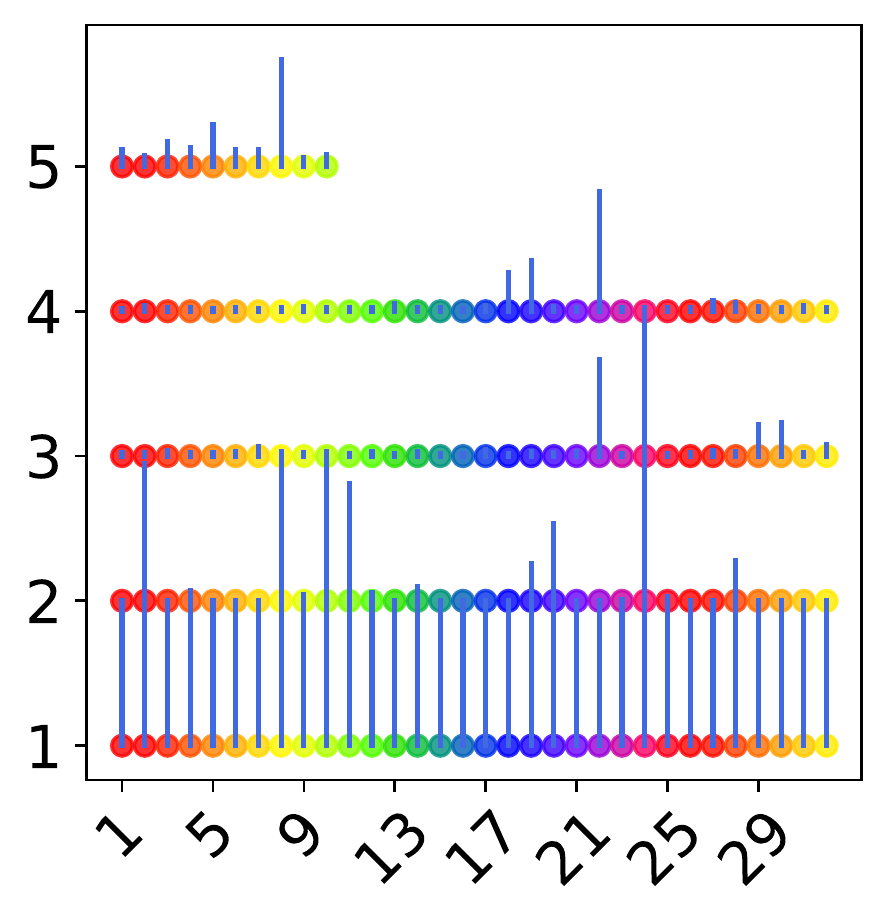} &
			\includegraphics[width=0.15\linewidth]{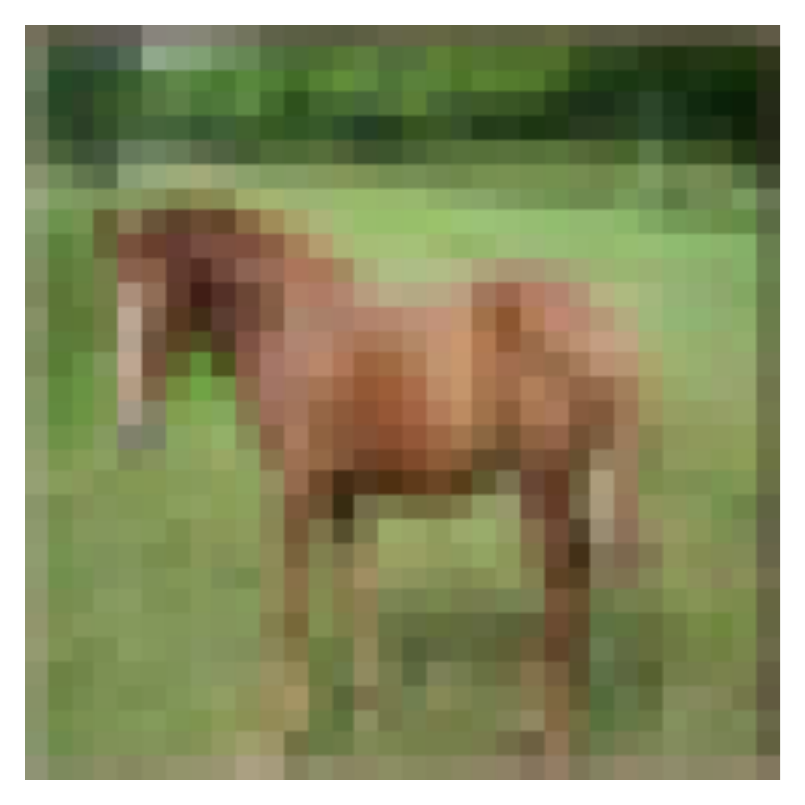} \\
			\includegraphics[width=0.15\linewidth]{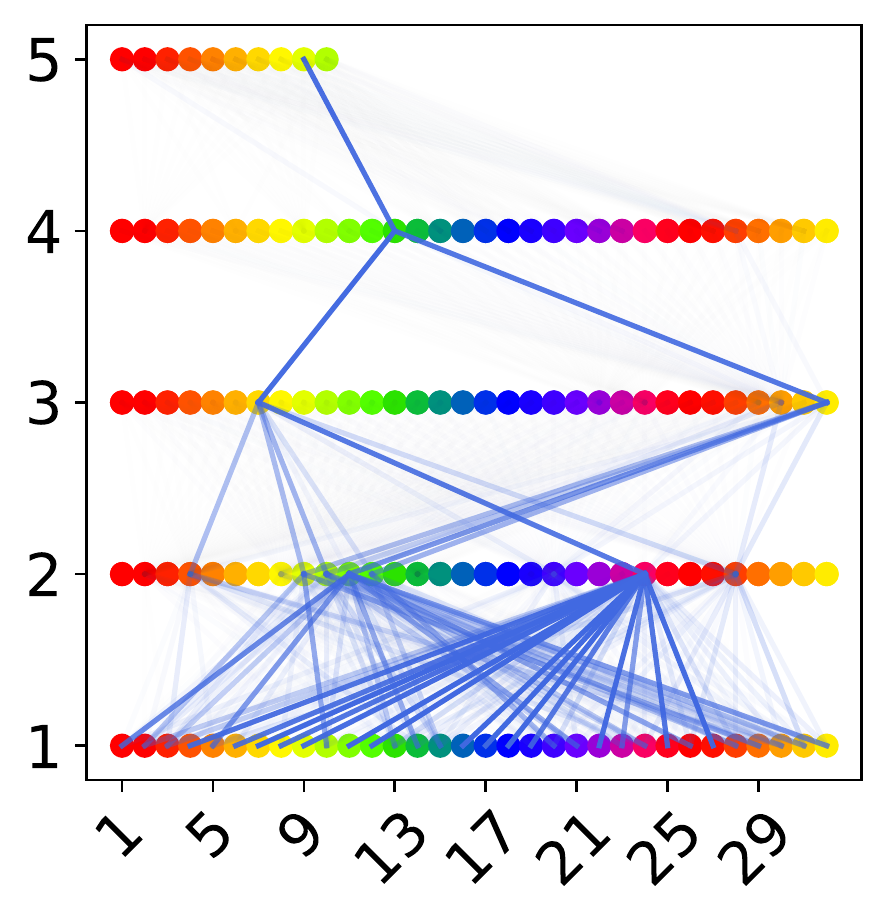} &
			\includegraphics[width=0.15\linewidth]{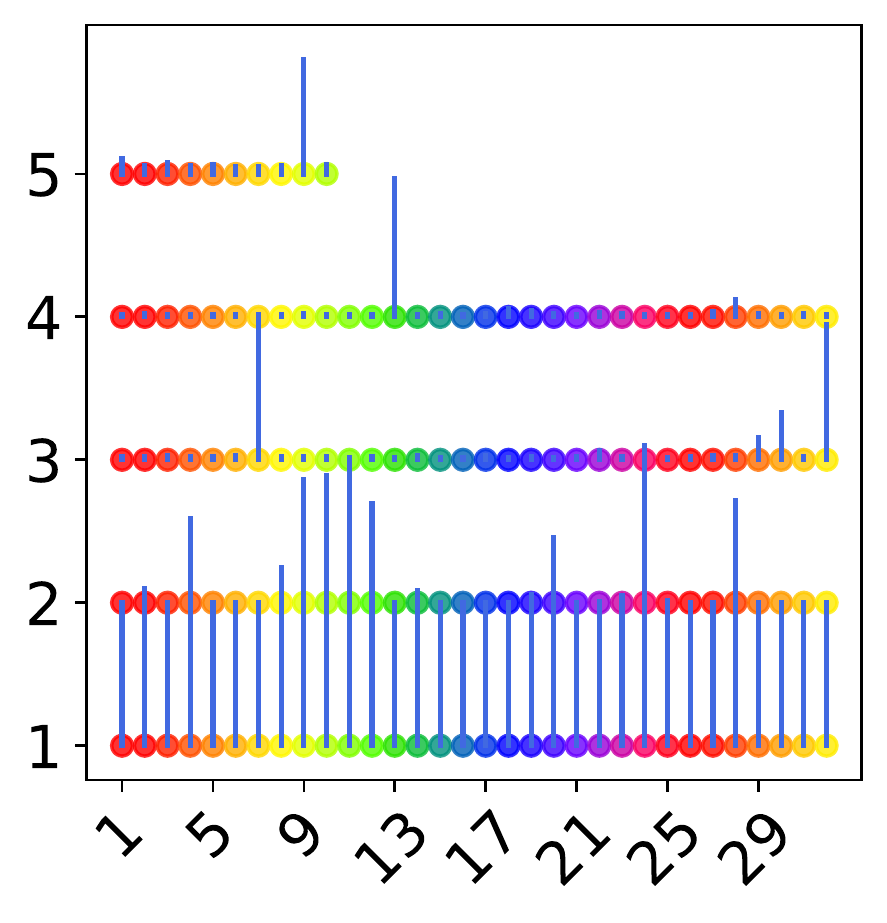} &
			\includegraphics[width=0.15\linewidth]{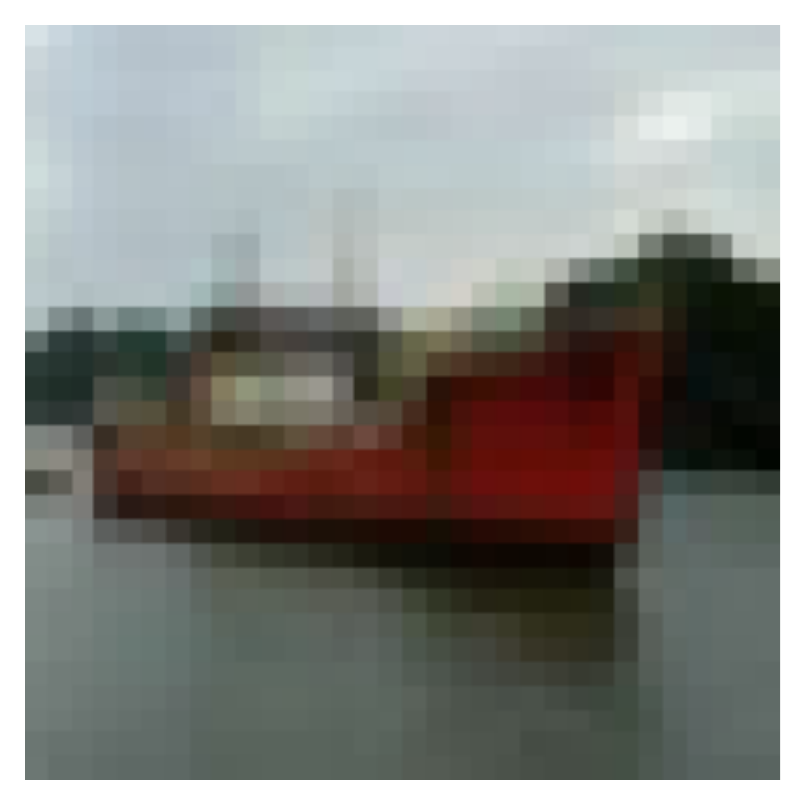} &
			\includegraphics[width=0.15\linewidth]{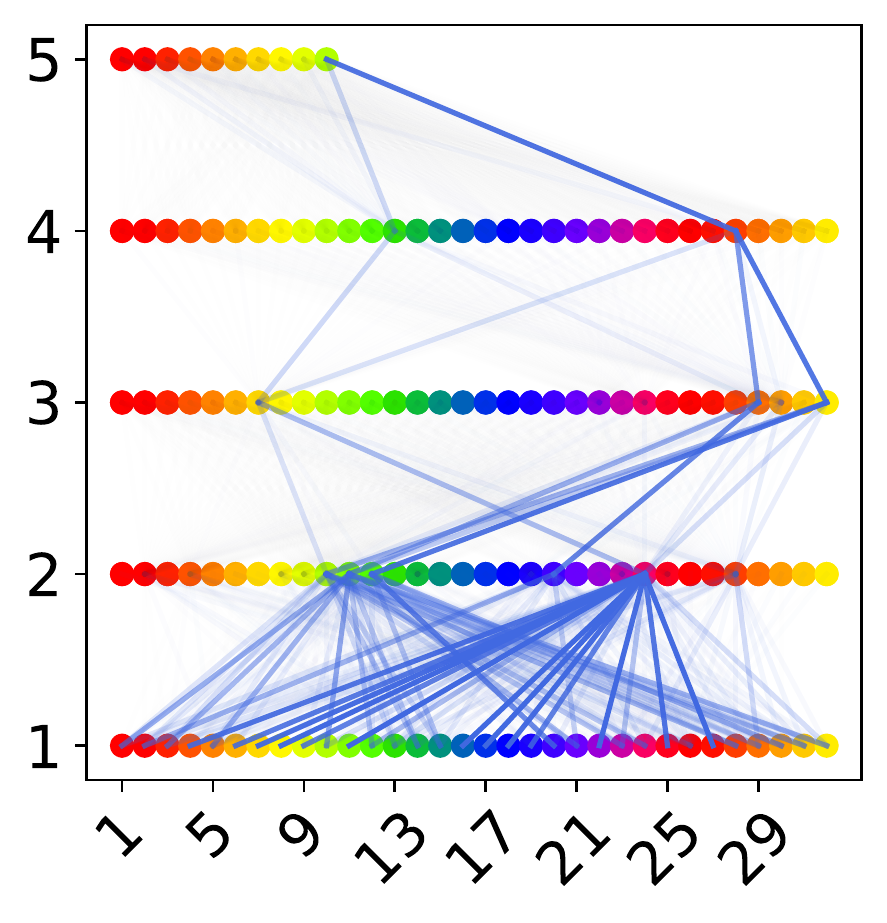} &
			\includegraphics[width=0.15\linewidth]{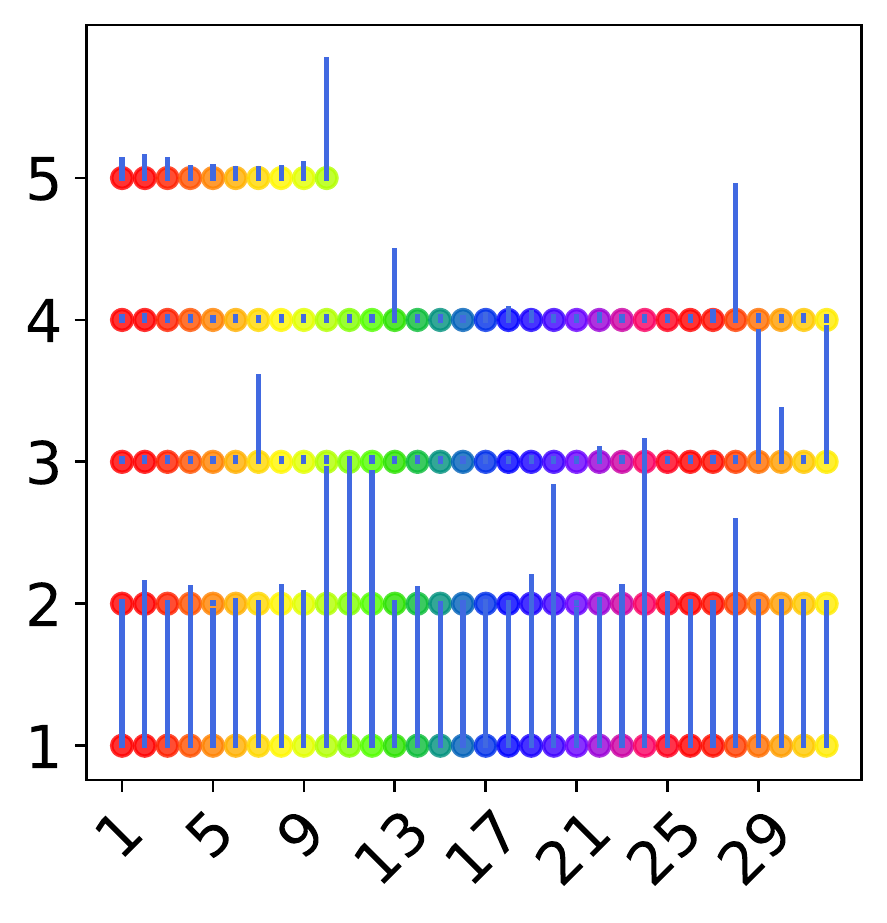} &
			\includegraphics[width=0.15\linewidth]{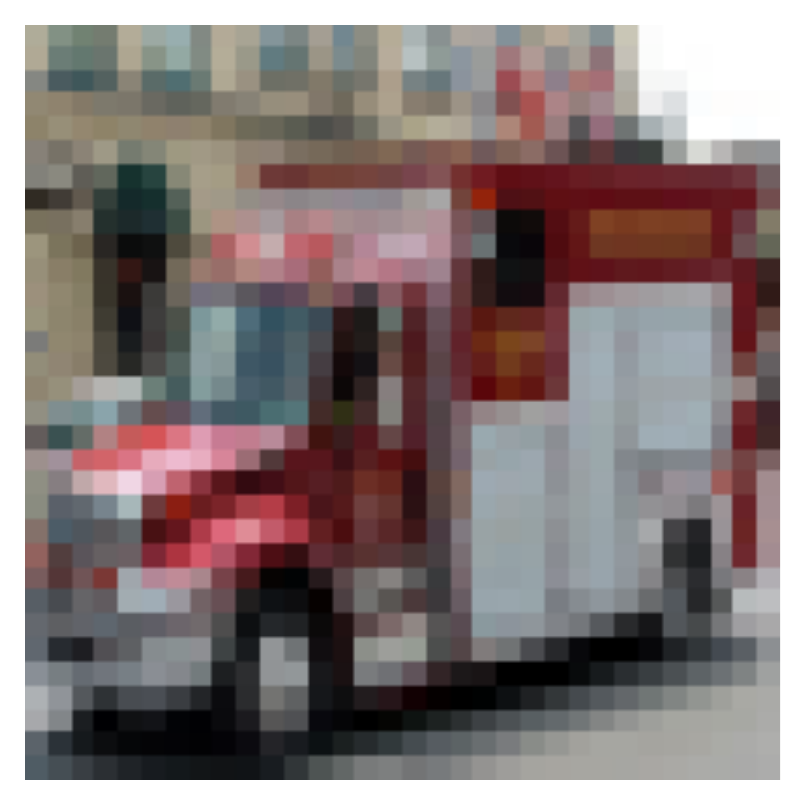} \\
		\end{tabular}
	\end{center}
	\caption{The parse-trees for CIFAR10 validation samples.}
	\label{fig:cifar10:main_model:all_classes}
\end{figure}

\begin{figure}[!htb]%
	\centering
	\includegraphics[width=0.6\textwidth]{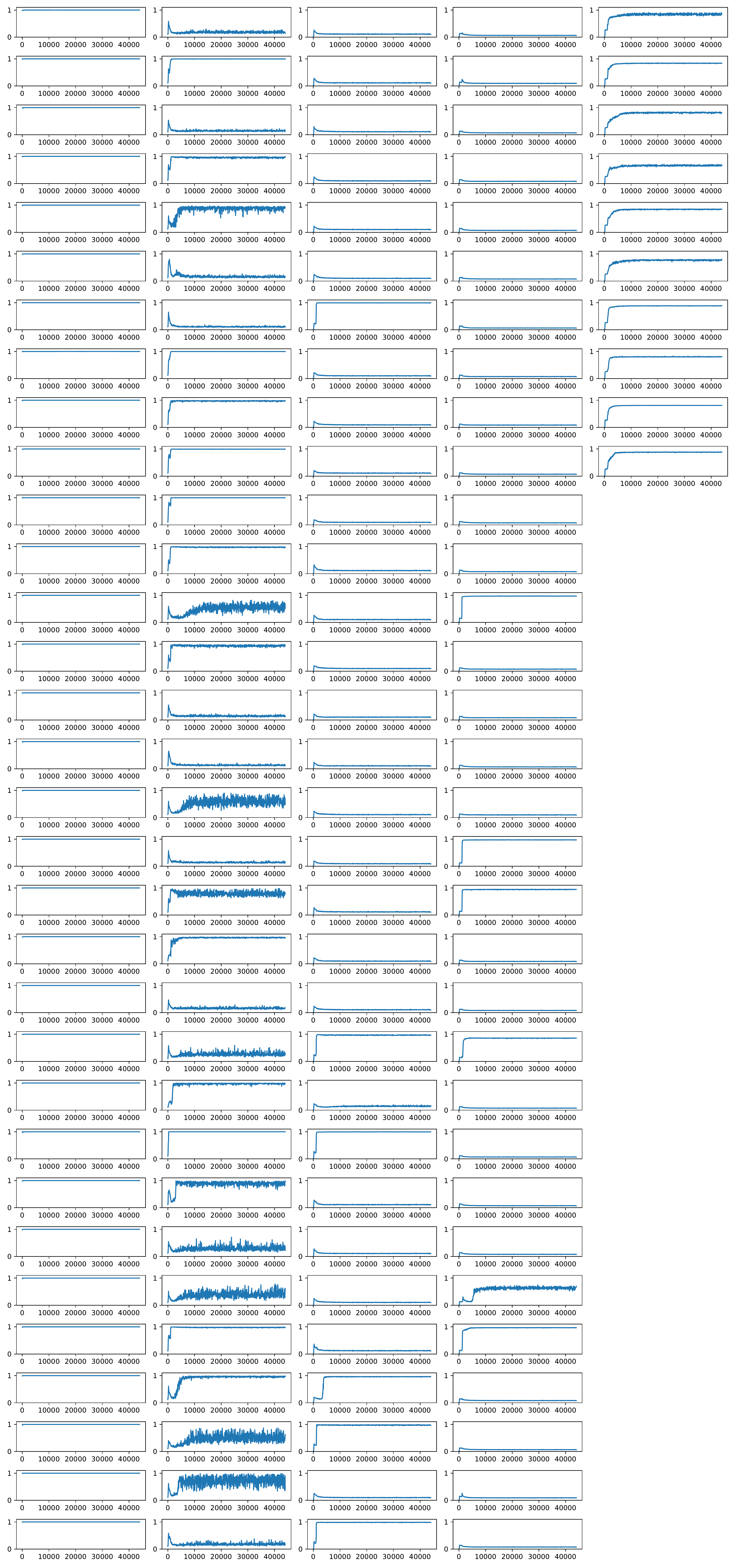}
	\caption{The maximal capsule norm/activation per batch over the training period of the model with the capsule layers shown in columns and the individual capsules per layer shown in rows.}%
	\label{fig:cifar10:main_model:training:caps_norms_max}
\end{figure}

\begin{figure}[!htb]%
	\centering
	\makebox[\textwidth][c]{\includegraphics[width=1.4\textwidth]{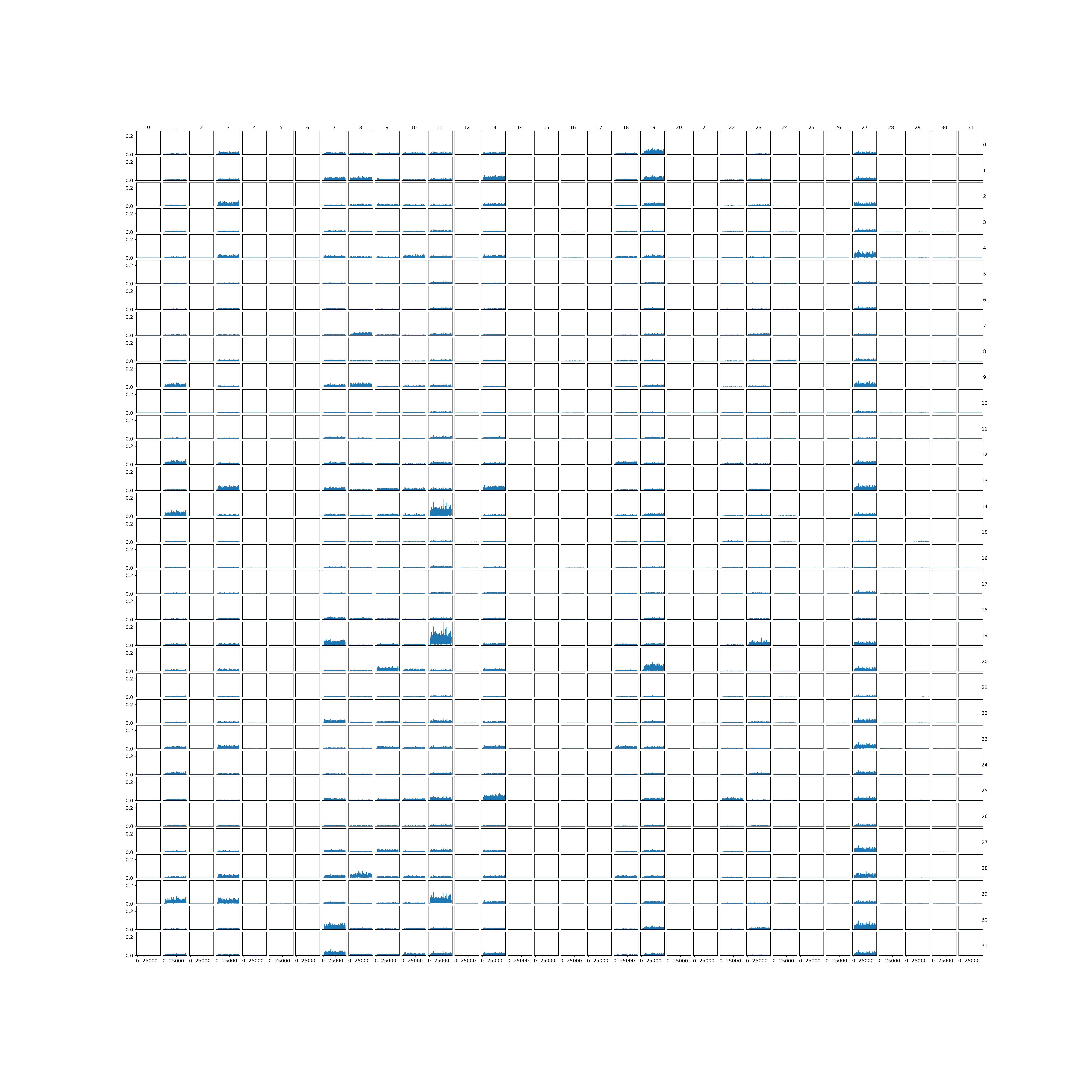}}%
	\vspace{-2cm}
	\caption{\textbf{First routing layer}: The gradient norms for the weight matrices $\frac{\partial L_m}{\partial W^{1}_{(j,i,:,:)}}$ over the training period of the model with lower layer capsules $i$ in rows and upper layer capsules $j$ in columns.}%
	\label{fig:cifar10:main_model:training:grad_norms_0}
\end{figure}

\begin{figure}[!htb]%
	\centering
	\makebox[\textwidth][c]{\includegraphics[width=1.4\textwidth]{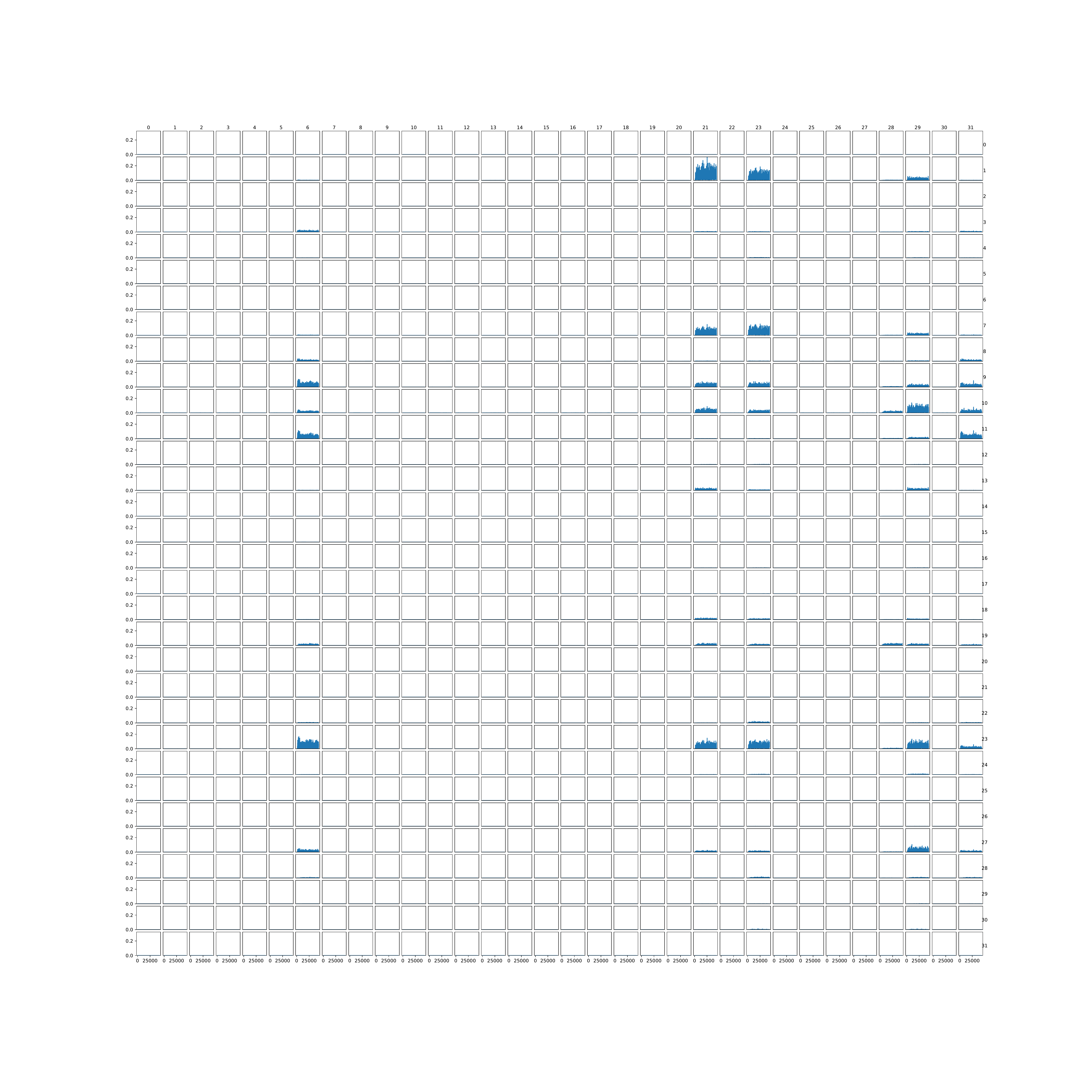}}%
	\vspace{-2cm}
	\caption{\textbf{Second routing layer}: The gradient norms for the weight matrices $\frac{\partial L_m}{\partial W^{2}_{(j,i,:,:)}}$ over the training period of the model with lower layer capsules $i$ in rows and upper layer capsules $j$ in columns.}%
	\label{fig:cifar10:main_model:training:grad_norms_1}
\end{figure}

\begin{figure}[!htb]%
	\centering
	\makebox[\textwidth][c]{\includegraphics[width=1.4\textwidth]{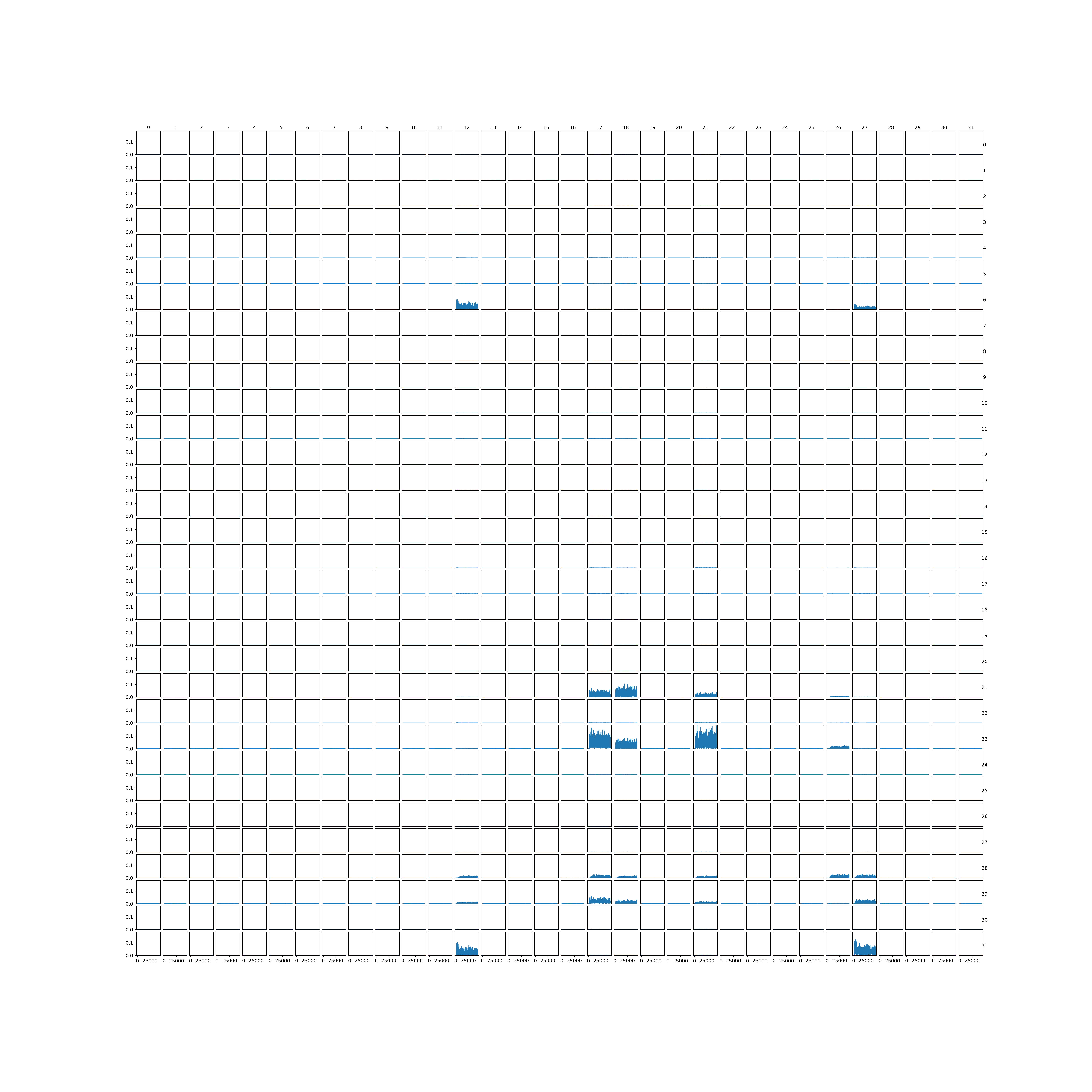}}%
	\vspace{-2cm}
	\caption{\textbf{Third routing layer}: The gradient norms for the weight matrices $\frac{\partial L_m}{\partial W^{3}_{(j,i,:,:)}}$ over the training period of the model with lower layer capsules $i$ in rows and upper layer capsules $j$ in columns.}%
	\label{fig:cifar10:main_model:training:grad_norms_2}
\end{figure}

\begin{figure}[!htb]%
	\centering
	\makebox[\textwidth][c]{\includegraphics[width=0.4\textwidth]{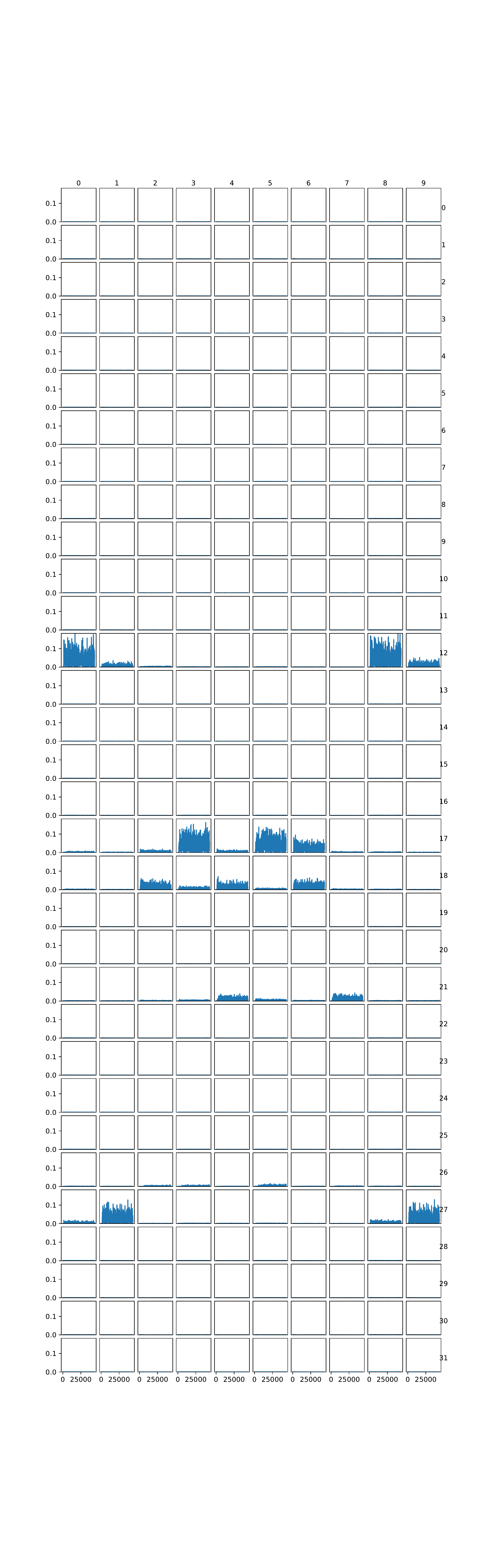}}%
	\vspace{-2cm}
	\caption{\textbf{Final routing layer}: The gradient norms for the weight matrices $\frac{\partial L_m}{\partial W^{4}_{(j,i,:,:)}}$ over the training period of the model with lower layer capsules $i$ in rows and upper layer capsules $j$ in columns.}%
	\label{fig:cifar10:main_model:training:grad_norms_3}
\end{figure}

\clearpage
\section{CIFAR10: Exhaustive Experiments on Model Architectures}\label{app:cifar10_grid}

\begin{figure}[!htb]
	\centering
	\includegraphics[width=1.0\textwidth]{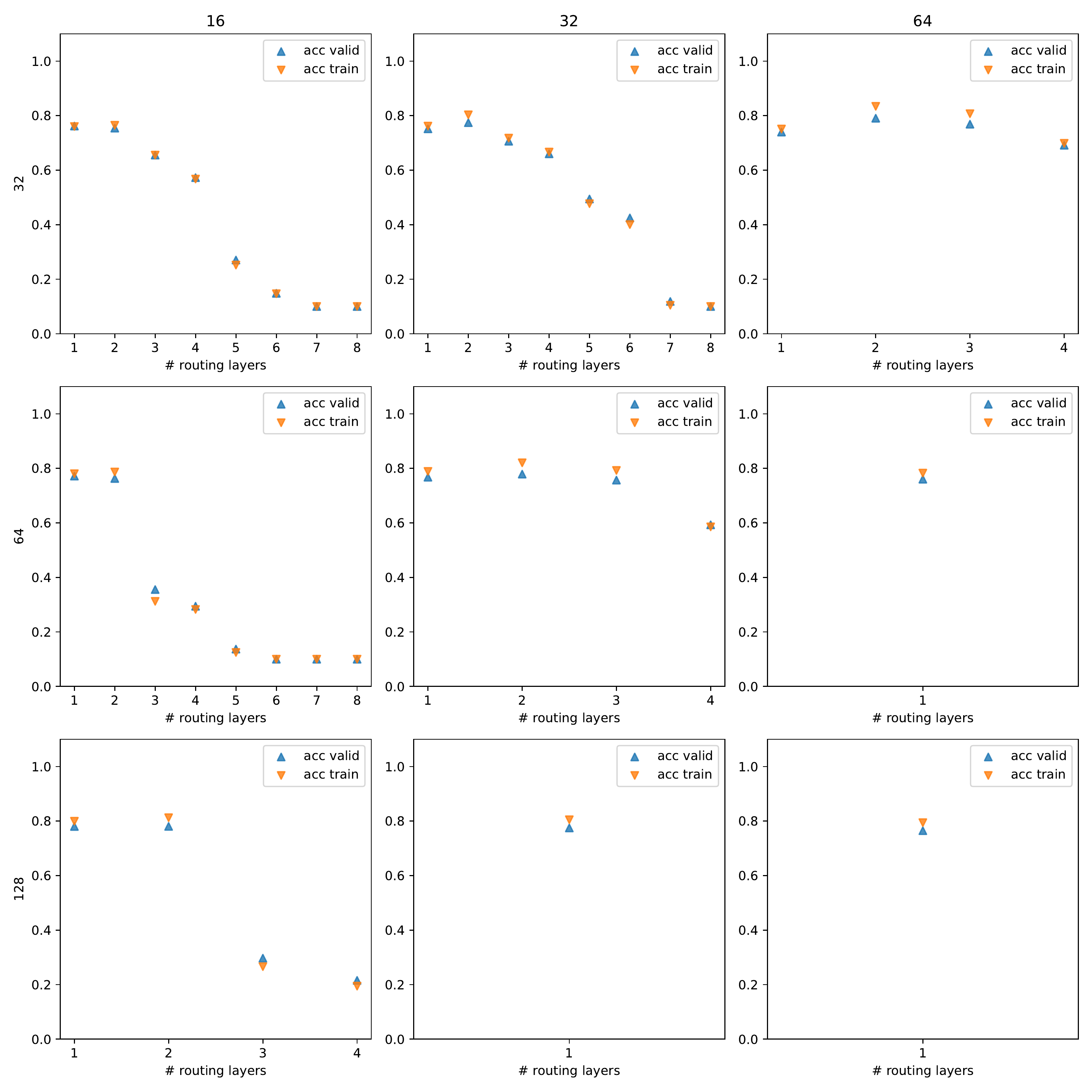}
	\caption{Reported accuracies for the CIFAR10 image classification experiments.}%
	\label{fig:cifar10:grid_run:depth_width_dim_acc}
\end{figure}

In this section, we report the results of our exhaustive experiments on model architectures for the CIFAR10 image classification task. We trained all models following the training procedure as described in Appendix~\ref{app:model_architectures}.
We used a total of 40 different architectures with varying numbers of routing layers $\{1,2,3,4,5,6,7,8\}$, capsules per layer $\{32, 64, 128\}$ as well as capsule dimensions $\{16, 32, 64\}$.
We chose these parameters to cover a broad range of settings, from simple one-layer models to complex models that used all the available GPU RAM.
We report the best achieved accuracies in Figure~\ref{fig:cifar10:grid_run:depth_width_dim_acc}.
Table~\ref{tab:cifar10:grid_run:best_overall_models} lists the best overall models with architecture details, number of parameters and a uniform routing baselines.
The best models per depth are given in Table~\ref{tab:cifar10:grid_run:best_models_depth} and the corresponding metrics and measurements are given in Tables~\ref{tab:cifar10:model_d1:best},~\ref{tab:cifar10:model_d2:best},~\ref{tab:cifar10:model_d3:best},~\ref{tab:cifar10:model_d4:best},~\ref{tab:cifar10:model_d5:best},~\ref{tab:cifar10:model_d6:best}.

\begin{table}[htb!]
	\centering
	\begin{tabular}{ccc|cc|cc}
		\toprule
		\multicolumn{3}{c}{Model Settings} &
		\multicolumn{2}{c}{Parameters} &
		\multicolumn{2}{c}{Valid Acc.} \\
		\#caps & dim & depth  & Routing  & Backbone & RBA  & Uniform  \\
		\midrule
		32 & 64 &  2 & 452  & 704 & 0.79 & 0.79 \\
		128 & 16 &  2 & 454 & 706 & 0.78 & 0.76 \\
		128 & 16 &  1 & 33 & 285 & 0.78 & 0.76 \\
		64 & 32 &  2 & 453 & 704 & 0.78 & 0.77 \\
		128 & 32 &  1 & 66 & 559 & 0.77 & 0.76 \\
		32 & 32 &  2 & 121 & 252 & 0.77 & 0.76 \\
		64 & 16 &  1 & 16 & 147 & 0.77 & 0.75 \\
		32 & 64 &  3 &  872 & 1123 & 0.77 & 0.77 \\
		64 & 32 &  1 &  33 & 284 & 0.77 & 0.74 \\
		128 & 64 &  1 &  131 & 1108 & 0.76 & 0.75 \\
		64 & 16 &  2 & 122 & 252 & 0.76 & 0.73 \\
		32 & 16 &  1 & 8 & 78 & 0.76 & 0.74 \\
		64 & 64 &  1 & 66 & 559 & 0.76 & 0.75 \\
		64 & 32 &  3 & 873 & 1124 & 0.76 & 0.2 \\
		32 & 16 &  2 & 35 & 105 & 0.75 & 0.73 \\
		32 & 32 &  1 & 16 & 147 & 0.75 & 0.73 \\
		32 & 64 &  1 & 33 & 284 & 0.74 & 0.68 \\
		32 & 32 &  3 & 226 & 357 & 0.71 & 0.62 \\
		32 & 64 &  4 & 1291 & 1543 & 0.69 & 0.62 \\
		32 & 32 &  4 & 331 & 462 & 0.66 & 0.13 \\
		32 & 16 &  3 & 61 & 131 & 0.66 & 0.21 \\
		\bottomrule
	\end{tabular}
	\caption{Overview of the best overall models on the CIFAR10 image classification task.
		For a model, we list the number of capsules per layer (\#caps), the dimension of the capsules (dim), and the number of routing layers (depth).
		The number of backbone parameters and the sum of all routing layer parameters are listed separately in \textbf{10k}. We give the validation accuracy for the model when trained with uniform routing and with RBA.}
	\label{tab:cifar10:grid_run:best_overall_models}
\end{table}

\begin{table}[htb!]
	\centering
	\begin{tabular}{ccc|cc|c}
		\toprule
		\multicolumn{3}{c}{Model Settings} &
		\multicolumn{2}{c}{Parameters} &
		\multirow{2}{*}{Valid Acc.} \\
		depth  & \#caps & dim  & Routing  & Backbone &  \\
		\midrule
		1 & 128 & 16 & 33 & 285 & 0.78 \\
		2 & 32 & 64 & 452 & 704 & 0.79 \\
		3 & 32 & 64 & 872 & 1123 & 0.77 \\
		4 & 32 & 64 & 1291 & 1543 & 0.69 \\
		5 & 32 & 32 & 436 & 567 & 0.49 \\
		6 & 32 & 32 & 541 & 672 & 0.43 \\
		\bottomrule
	\end{tabular}
	\caption{Overview of the best models per depth on the CIFAR10 image classification task.
		For a model, we list the number of capsules per layer (\#caps), the dimension of the capsules (dim), and the number of routing layers (depth).
		The number of backbone parameters and the sum of all routing layer parameters are listed separately in \textbf{10k}.}
	\label{tab:cifar10:grid_run:best_models_depth}
\end{table}

\begin{table}%
	\centering
	\subfloat[]{
	\begin{tabular}{ccc|cc|cc}
	\toprule
	\midrule
	\multirow{2}{*}{Capsule Layer} &
	\multicolumn{2}{c}{Capsule Norms} &
	\multicolumn{2}{c}{Capsule Activation} &
	\multicolumn{2}{c}{Capsule Deaths} \\
	& Mean ($\cnm$) & Sum ($\cns$) & Rate ($\car$) & Sum ($\cas$) & Rate ($\cdr$) & Sum ($\cds$) \\
	\midrule
	1 & 0.90  & 115.56 & 1.00  & 128.00 & 0.00  & 0.00\\
	2 & 0.21  & 2.10 & 0.78  & 7.78 & 0.00  & 0.00\\
	\bottomrule
\end{tabular}
	}%
	\vspace{0.5cm}
	\subfloat[]{
	\begin{tabular}{ccc|cc}
	\toprule
	\multirow{2}{*}{Routing Layer} &
	\multicolumn{2}{c}{Capsules Alive} &
	\multicolumn{2}{c}{Routing Dynamics} \\
	& From lower layer & To higher layer & Rate ($\dyr$) & Mean ($\dys$) \\
	\midrule
	1 & 128 & 10 & 0.06 & 0.62 \\
	\midrule
	\bottomrule
\end{tabular}
	}
	\caption{Capsule activation and routing dynamics for the best model with one routing layer.}%
	\label{tab:cifar10:model_d1:best}%
\end{table}

\begin{table}%
	\centering
	\subfloat[]{
	\begin{tabular}{ccc|cc|cc}
	\toprule
	\midrule
	\multirow{2}{*}{Capsule Layer} &
	\multicolumn{2}{c}{Capsule Norms} &
	\multicolumn{2}{c}{Capsule Activation} &
	\multicolumn{2}{c}{Capsule Deaths} \\
	& Mean / Capsule & Mean / Layer & Rate / Layer & Mean / Layer & Rate / Layer & Mean / Layer \\
	\midrule
	1 & 0.99  & 31.76 & 1.00  & 32.00 & 0.00  & 0.00\\
	2 & 0.18  & 5.92 & 0.63  & 20.10 & 0.47  & 15.00\\
	3 & 0.18  & 1.82 & 0.40  & 3.96 & 0.00  & 0.00\\
	\bottomrule
\end{tabular}
	}%
	\vspace{0.5cm}
	\subfloat[]{
	\begin{tabular}{ccc|cc}
	\toprule
	\multirow{2}{*}{Routing Layer} &
	\multicolumn{2}{c}{Capsules Alive} &
	\multicolumn{2}{c}{Routing Dynamics} \\
	& From lower layer & To higher layer & Rate ($\dyr$) & Mean ($\dys$) \\
	\midrule
	1 & 32 & 17 & 0.19 & 3.25 \\
	2 & 17 & 10 & 0.17 & 1.73 \\
	\midrule
	\bottomrule
\end{tabular}
	}
	\caption{Capsule activation and routing dynamics for the best model with two routing layers.}%
	\label{tab:cifar10:model_d2:best}%
\end{table}

\begin{table}%
	\centering
	\subfloat[]{
	\begin{tabular}{ccc|cc|cc}
	\toprule
	\midrule
	\multirow{2}{*}{Capsule Layer} &
	\multicolumn{2}{c}{Capsule Norms} &
	\multicolumn{2}{c}{Capsule Activation} &
	\multicolumn{2}{c}{Capsule Deaths} \\
	& Mean ($\cnm$) & Sum ($\cns$) & Rate ($\car$) & Sum ($\cas$) & Rate ($\cdr$) & Sum ($\cds$) \\
	\midrule
	1 & 0.99  & 31.75 & 1.00  & 32.00 & 0.00  & 0.00\\
	2 & 0.24  & 7.80 & 0.56  & 17.89 & 0.22  & 7.00\\
	3 & 0.10  & 3.12 & 0.18  & 5.81 & 0.72  & 23.00\\
	4 & 0.17  & 1.75 & 0.39  & 3.93 & 0.00  & 0.00\\
	\bottomrule
\end{tabular}
	}%
	\vspace{0.5cm}
	\subfloat[]{
	\begin{tabular}{ccc|cc}
	\toprule
	\multirow{2}{*}{Routing Layer} &
	\multicolumn{2}{c}{Capsules Alive} &
	\multicolumn{2}{c}{Routing Dynamics} \\
	& From lower layer & To higher layer & Rate ($\dyr$) & Mean ($\dys$) \\
	\midrule
	1 & 32 & 25 & 0.26 & 6.49 \\
	2 & 25 & 9 & 0.15 & 1.36 \\
	3 & 9 & 10 & 0.27 & 2.66 \\
	\midrule
	\bottomrule
\end{tabular}
	}
	\caption{Capsule activation and routing dynamics for the best model with three routing layers.}%
	\label{tab:cifar10:model_d3:best}%
\end{table}

\begin{table}%
	\centering
	\subfloat[]{
	\begin{tabular}{ccc|cc|cc}
	\toprule
	\midrule
	\multirow{2}{*}{Capsule Layer} &
	\multicolumn{2}{c}{Capsule Norms} &
	\multicolumn{2}{c}{Capsule Activation} &
	\multicolumn{2}{c}{Capsule Deaths} \\
	& Mean ($\cnm$) & Sum ($\cns$) & Rate ($\car$) & Sum ($\cas$) & Rate ($\cdr$) & Sum ($\cds$) \\
	\midrule
	1 & 0.99  & 31.57 & 1.00  & 32.00 & 0.00  & 0.00\\
	2 & 0.23  & 7.39 & 0.35  & 11.17 & 0.53  & 17.00\\
	3 & 0.14  & 4.43 & 0.27  & 8.79 & 0.66  & 21.00\\
	4 & 0.08  & 2.67 & 0.16  & 5.08 & 0.78  & 25.00\\
	5 & 0.19  & 1.95 & 0.61  & 6.10 & 0.00  & 0.00\\
	\bottomrule
\end{tabular}
	}%
	\vspace{0.5cm}
	\subfloat[]{
	\begin{tabular}{ccc|cc}
	\toprule
	\multirow{2}{*}{Routing Layer} &
	\multicolumn{2}{c}{Capsules Alive} &
	\multicolumn{2}{c}{Routing Dynamics} \\
	& From lower layer & To higher layer & Rate ($\dyr$) & Mean ($\dys$) \\
	\midrule
	1 & 32 & 15 & 0.30 & 4.45 \\
	2 & 15 & 11 & 0.20 & 2.21 \\
	3 & 11 & 7 & 0.13 & 0.93 \\
	4 & 7 & 10 & 0.26 & 2.63 \\
	\midrule
	\bottomrule
\end{tabular}
	}
	\caption{Capsule activation and routing dynamics for the best model with four routing layers.}%
	\label{tab:cifar10:model_d4:best}%
\end{table}

\begin{table}%
	\centering
	\subfloat[]{
	\begin{tabular}{ccc|cc|cc}
	\toprule
	\midrule
	\multirow{2}{*}{Capsule Layer} &
	\multicolumn{2}{c}{Capsule Norms} &
	\multicolumn{2}{c}{Capsule Activation} &
	\multicolumn{2}{c}{Capsule Deaths} \\
	& Mean ($\cnm$) & Sum ($\cns$) & Rate ($\car$) & Sum ($\cas$) & Rate ($\cdr$) & Sum ($\cds$) \\
	\midrule
	1 & 0.98  & 31.41 & 1.00  & 32.00 & 0.00  & 0.00\\
	2 & 0.12  & 3.77 & 0.17  & 5.47 & 0.75  & 24.00\\
	3 & 0.10  & 3.04 & 0.15  & 4.92 & 0.81  & 26.00\\
	4 & 0.08  & 2.71 & 0.11  & 3.62 & 0.84  & 27.00\\
	5 & 0.07  & 2.36 & 0.10  & 3.12 & 0.81  & 26.00\\
	6 & 0.22  & 2.19 & 0.82  & 8.17 & 0.00  & 0.00\\
	\bottomrule
\end{tabular}
	}%
	\vspace{0.5cm}
	\subfloat[]{
	\begin{tabular}{ccc|cc}
	\toprule
	\multirow{2}{*}{Routing Layer} &
	\multicolumn{2}{c}{Capsules Alive} &
	\multicolumn{2}{c}{Routing Dynamics} \\
	& From lower layer & To higher layer & Rate ($\dyr$) & Mean ($\dys$) \\
	\midrule
	1 & 32 & 8 & 0.19 & 1.55 \\
	2 & 8 & 6 & 0.16 & 0.93 \\
	3 & 6 & 5 & 0.21 & 1.06 \\
	4 & 5 & 6 & 0.14 & 0.84 \\
	5 & 6 & 10 & 0.21 & 2.12 \\
	\midrule
	\bottomrule
\end{tabular}
	}
	\caption{Capsule activation and routing dynamics for the best model with five routing layers.}%
	\label{tab:cifar10:model_d5:best}%
\end{table}

\begin{table}%
	\centering
	\subfloat[]{
	\begin{tabular}{ccc|cc|cc}
	\toprule
	\midrule
	\multirow{2}{*}{Capsule Layer} &
	\multicolumn{2}{c}{Capsule Norms} &
	\multicolumn{2}{c}{Capsule Activation} &
	\multicolumn{2}{c}{Capsule Deaths} \\
	& Mean ($\cnm$) & Sum ($\cns$) & Rate ($\car$) & Sum ($\cas$) & Rate ($\cdr$) & Sum ($\cds$) \\
	\midrule
	1 & 0.96  & 30.84 & 1.00  & 32.00 & 0.00  & 0.00\\
	2 & 0.17  & 5.28 & 0.20  & 6.44 & 0.72  & 23.00\\
	3 & 0.11  & 3.60 & 0.18  & 5.75 & 0.72  & 23.00\\
	4 & 0.09  & 2.80 & 0.12  & 3.79 & 0.84  & 27.00\\
	5 & 0.08  & 2.55 & 0.09  & 2.78 & 0.84  & 27.00\\
	6 & 0.07  & 2.20 & 0.09  & 2.97 & 0.84  & 27.00\\
	7 & 0.22  & 2.24 & 0.83  & 8.27 & 0.00  & 0.00\\
	\bottomrule
\end{tabular}
	}%
	\vspace{0.5cm}
	\subfloat[]{
	\begin{tabular}{ccc|cc}
	\toprule
	\multirow{2}{*}{Routing Layer} &
	\multicolumn{2}{c}{Capsules Alive} &
	\multicolumn{2}{c}{Routing Dynamics} \\
	& From lower layer & To higher layer & Rate ($\dyr$) & Mean ($\dys$) \\
	\midrule
	1 & 32 & 9 & 0.25 & 2.24 \\
	2 & 9 & 9 & 0.18 & 1.63 \\
	3 & 9 & 5 & 0.22 & 1.12 \\
	4 & 5 & 5 & 0.25 & 1.25 \\
	5 & 5 & 5 & 0.20 & 0.99 \\
	6 & 5 & 10 & 0.18 & 1.78 \\
	\midrule
	\bottomrule
\end{tabular}
	}
	\caption{Capsule activation and routing dynamics for the best model with six routing layers.}%
	\label{tab:cifar10:model_d6:best}%
\end{table}

\end{document}